\documentclass[]{template}

\usepackage{amsmath}
\usepackage{amssymb}
\usepackage{amsthm}
\usepackage{mathtools}
\usepackage{fontawesome5}
\usepackage{letterspace}  % \textls used by the inlined codec figure (figures/codec_body.tex)
\usetikzlibrary{positioning,arrows.meta,calc,shapes.geometric,fit,backgrounds,shadows,decorations.pathreplacing}

\usepackage{enumitem}
\usepackage{algorithm}
\usepackage{algpseudocode}
\usepackage{booktabs}
\usepackage{float}
\usepackage{subcaption}
\usepackage{pgfplots}
\usetikzlibrary{patterns}
\usetikzlibrary{calc}
\usepgfplotslibrary{groupplots}
\pgfplotsset{compat=1.18}
\usepackage{array}
\usepackage{makecell}
\usepackage{tabularx}
\usepackage{multirow}

\usepackage[most]{tcolorbox}
\usepackage{xcolor}
\usepackage{pifont}
\usepackage{fvextra}

\definecolor{ink}{HTML}{1C2B33}
\definecolor{muted}{HTML}{8A93A6}
\definecolor{line}{HTML}{D7DEE8}
\definecolor{frz}{HTML}{0E8FA1}      % frozen (teal)
\definecolor{frzfl}{HTML}{E9F4F6}
\definecolor{trn}{HTML}{E08A1E}      % trained (orange)
\definecolor{trnfl}{HTML}{FDF2E1}
\definecolor{lossred}{HTML}{D7392A}  % reconstruction / flow-matching loss (red)
\definecolor{screenbg}{HTML}{141E32}
\definecolor{pitchhi}{HTML}{1F6B3A}
\definecolor{pitchlo}{HTML}{0C3B20}
\definecolor{skyhi}{HTML}{1D3157}
\definecolor{skylo}{HTML}{101A33}
\definecolor{ballblue}{HTML}{0091FF}
\newcommand{\ti}[1]{{\fontsize{6.5}{7.6}\selectfont\bfseries #1}}
\newcommand{\sub}[1]{{\fontsize{4.6}{5.4}\selectfont\itshape #1}}
\newcommand{\lab}[1]{{\fontsize{6}{7}\selectfont\bfseries\textls[40]{\MakeUppercase{#1}}}}
\newcommand{\shp}[1]{{\ttfamily\fontsize{4.6}{5.4}\selectfont\color{muted} #1}}
\newcommand{\fac}[1]{{\ttfamily\fontsize{4.6}{5.4}\selectfont #1}}
\newcommand{\pill}[2]{% small badge (frozen/trained)
  \tikz[baseline]{\node[fill=white, draw=#1, line width=0.6pt, rounded corners=2.5pt,
    inner xsep=3pt, inner ysep=1.3pt, text=#1]
    {{\fontsize{4.3}{4.8}\selectfont\bfseries\textls[50]{\MakeUppercase{#2}}}};}}
\newcommand{\bpill}[2]{% larger loss-term pill
  \tikz[baseline]{\node[fill=white, draw=#1, line width=0.7pt, rounded corners=3pt,
    inner xsep=4pt, inner ysep=2pt, text=#1]
    {{\fontsize{5.6}{6.4}\selectfont\bfseries #2}};}}

\definecolor{ink}{HTML}{1C2B33}
\definecolor{muted}{HTML}{8A93A6}
\definecolor{line}{HTML}{D7DEE8}
\definecolor{frz}{HTML}{0E8FA1}
\definecolor{frzfl}{HTML}{E9F4F6}
\definecolor{trn}{HTML}{E08A1E}
\definecolor{trnfl}{HTML}{FDF2E1}
\definecolor{trndk}{HTML}{8A4E0A}
\definecolor{lossred}{HTML}{D7392A}
\definecolor{nz}{HTML}{6A4C93}
\definecolor{nzfl}{HTML}{EFE9F6}
\definecolor{act}{HTML}{2A7DE1}
\definecolor{actfl}{HTML}{E7F0FC}
\definecolor{screenbg}{HTML}{141E32}
\definecolor{slate}{HTML}{667085}
\definecolor{slatefl}{HTML}{F2F4F7}

\providecommand{\wti}[1]{{\fontsize{7}{8}\selectfont\bfseries #1}}
\providecommand{\wsub}[1]{{\fontsize{5}{6}\selectfont\itshape #1}}

\providecommand{\wmono}[1]{{\ttfamily\fontsize{5.2}{6}\selectfont #1}}
\providecommand{\wstack}[3]{%
  \begin{scope}[shift={($(#1)-(0.0845,0.0845)$)}]% shift so the stack's bbox is centered on #1
    \foreach \o in {0.169,0.0845}{
      \draw[fill=white, draw=#3, line width=0.7pt, rounded corners=3pt]
        (-0.26+\o,-0.65+\o) rectangle (0.26+\o,0.65+\o);}
    \draw[fill=white, draw=#3, line width=0.9pt, rounded corners=3pt]
      (-0.26,-0.65) rectangle (0.26,0.65);
    \foreach \gx in {0,1,2,3}{\foreach \gy in {0,...,11}{
      \fill[#2, rounded corners=0.2pt] (-0.1716+0.09295*\gx,-0.56875+0.0975*\gy) rectangle ++(0.065,0.065);}}
  \end{scope}}
\providecommand{\wstackmix}[1]{%
  \begin{scope}[shift={($(#1)-(0.0845,0.0845)$)}]
    \foreach \o in {0.169,0.0845}{
      \draw[fill=white, draw=nz!45, line width=0.7pt, rounded corners=3pt]
        (-0.26+\o,-0.65+\o) rectangle (0.26+\o,0.65+\o);}
    \fill[white, rounded corners=3pt] (-0.26,-0.65) rectangle (0.26,0.65);
    \draw[draw=nz!75, line width=0.9pt, rounded corners=3pt, dash pattern=on 3pt off 3pt]
      (-0.26,-0.65) rectangle (0.26,0.65);
    \draw[draw=muted!85, line width=0.9pt, rounded corners=3pt, dash pattern=on 3pt off 3pt, dash phase=3pt]
      (-0.26,-0.65) rectangle (0.26,0.65);
    \foreach \gx in {0,1,2,3}{\foreach \gy in {0,...,11}{
      \fill[muted!60, rounded corners=0.2pt] (-0.1716+0.09295*\gx,-0.56875+0.0975*\gy) rectangle ++(0.065,0.065);}}
    \foreach \gx/\gy in {0/0,0/2,0/3,0/6,0/7,0/10, 1/1,1/2,1/5,1/8,1/9,1/11,
                         2/0,2/3,2/4,2/7,2/10,2/11, 3/1,3/4,3/5,3/6,3/9,3/10}{
      \fill[nz!70, rounded corners=0.2pt] (-0.1716+0.09295*\gx,-0.56875+0.0975*\gy) rectangle ++(0.065,0.065);}
  \end{scope}}
\providecommand{\wtaustack}[3]{%
  \begin{scope}[shift={($(#1)-(0.055,0.055)$)}]
    \foreach \o in {0.11,0.055}{
      \draw[fill=white, draw=#3, line width=0.6pt, rounded corners=2pt]
        (-0.105+\o,-0.105+\o) rectangle (0.105+\o,0.105+\o);}
    \draw[fill=white, draw=#3, line width=0.8pt, rounded corners=2pt]
      (-0.105,-0.105) rectangle (0.105,0.105);
    \fill[#2, rounded corners=0.3pt] (-0.06,-0.06) rectangle (0.06,0.06);
  \end{scope}}
\providecommand{\wvec}[3]{%
  \begin{scope}[shift={($(#1)-(0.05,0.05)$)}]
    \foreach \o in {0.1,0.05}{
      \draw[fill=white, draw=#3, line width=0.7pt, rounded corners=1pt]
        (-0.075+\o,-0.34+\o) rectangle (0.075+\o,0.34+\o);}
    \draw[fill=white, draw=#3, line width=0.9pt, rounded corners=1pt]
      (-0.075,-0.34) rectangle (0.075,0.34);
    \foreach \gy in {0,...,5}{
      \fill[#2, rounded corners=0.2pt] (-0.0325,-0.2875+0.0975*\gy) rectangle ++(0.065,0.065);}
  \end{scope}}
\graphicspath{{figures/}{./}}
\providecommand{\wmshot}[2]{%
  \begin{scope}[shift={(#1)}]
    \foreach \o in {0.09,0.045}{\fill[screenbg, rounded corners=2.5pt, draw=line, line width=0.5pt]
       (-0.42+\o,-0.236+\o) rectangle (0.42+\o,0.236+\o);}
    \begin{scope}
      \clip[rounded corners=2.5pt] (-0.42,-0.236) rectangle (0.42,0.236);
      \node[inner sep=0pt] at (0,0) {\includegraphics[width=0.9cm]{#2}};
    \end{scope}
    \draw[line, line width=0.6pt, rounded corners=2.5pt] (-0.42,-0.236) rectangle (0.42,0.236);
  \end{scope}}
\providecommand{\wmini}[3]{%
  \begin{scope}[shift={(#1)}]
    \foreach \o in {0.07,0.035}{
      \draw[fill=white, draw=#3, line width=0.5pt, rounded corners=1.6pt]
        (-0.24+\o,-0.17+\o) rectangle (0.24+\o,0.17+\o);}
    \draw[fill=white, draw=#3, line width=0.6pt, rounded corners=1.6pt]
      (-0.24,-0.17) rectangle (0.24,0.17);
    \foreach \gx in {0,1,2,3}{\foreach \gy in {0,1,2}{
      \fill[#2, rounded corners=0.2pt] (-0.18+0.10*\gx,-0.13+0.10*\gy) rectangle ++(0.06,0.06);}}
  \end{scope}}
\providecommand{\pill}[2]{%
  \tikz[baseline]{\node[fill=white, draw=#1, line width=0.6pt, rounded corners=2.5pt,
    inner xsep=3pt, inner ysep=1.3pt, text=#1]
    {{\fontsize{4.3}{4.8}\selectfont\bfseries\textls[50]{\MakeUppercase{#2}}}};}}

\newif\ifshowcomments
\showcommentsfalse   % <- set to \showcommentstrue to show inline comments

\newcommand{\newcommenter}[3]{%
  \expandafter\newcommand\csname #1\endcsname[1]{%
    \ifshowcomments
      \textcolor{#3}{\textbf{\textit{[#2]:}}\ \textit{##1}}%
    \fi
  }%
}

\newcommenter{adrienrr}{AdrienRR}{red}
\newcommenter{chris}{Chris}{blue}
\newcommenter{carol}{Carol}{green!60!black}
\newcommenter{PP}{Pat}{orange}
\newcommenter{ah}{anthony}{purple}
\newcommenter{ea}{eloi}{teal}
\newcommenter{vm}{vince}{green}
\newcommenter{adam}{adam}{magenta}
\newcommenter{michael}{Michael Black}{olive}
\newcommenter{vv}{Václav}{violet}
\newcommenter{viktoriia}{Viktoriia}{cyan}
\newcommenter{lucia}{LucisS}{brown}
\newcommenter{am}{Aditya}{yellow}

\definecolor{cornflowerblue}{HTML}{6495ED}
\definecolor{mediumorchid}{HTML}{BA55D3}
\newcommenter{mle}{amelie}{mediumorchid}

  \definecolor{plotblue}{HTML}{0091FF}
  \definecolor{plotorange}{HTML}{E8662B}
  \definecolor{plotpurple}{HTML}{9B59B6}
  \definecolor{plotgreen}{HTML}{2BB673}
  \definecolor{acForward}{HTML}{0091FF}
  \definecolor{acPowerslide}{HTML}{9B59B6}
  \definecolor{acLeft}{HTML}{2BB673}
  \definecolor{acRight}{HTML}{E8662B}
  \definecolor{acBoost}{HTML}{C79000}
  \definecolor{acJump}{HTML}{0E8FA1}
  \definecolor{acReverse}{HTML}{D7392A}
  \definecolor{acAirL}{HTML}{D63384}
  \definecolor{acAirR}{HTML}{7A5230}
  \definecolor{sizeA}{HTML}{ADD3ED}   % 100M
  \definecolor{sizeB}{HTML}{6FB0DC}   % 300M
  \definecolor{sizeC}{HTML}{3888C6}   % 1B
  \definecolor{sizeD}{HTML}{1B5FA6}   % 2.5B
  \definecolor{sizeE}{HTML}{0A2A5E}   % 5B
  \pgfplotsset{                                                                                                
    rsplot/.style={         
      width=0.76\linewidth, height=0.48\linewidth,
      axis lines=left,
      axis line style={draw=gray!55, line width=0.6pt},
      every tick/.style={draw=gray!55, line width=0.5pt},
      tick align=outside, major tick length=3pt,        
      tick label style={font=\footnotesize, text=black!72},
      label style={font=\small, text=black!85},
      ymajorgrids=true,
      grid style={draw=gray!18, line width=0.5pt},      
      enlarge x limits=0.03,
      legend style={font=\footnotesize, draw=none, fill=white, fill opacity=0.82,
                    text opacity=1, rounded corners=1.5pt, inner sep=3pt, row sep=1.5pt},
      legend cell align=left,
      every axis plot/.append style={line width=1.3pt, mark size=1.7pt,
                    mark options={solid, draw=white, line width=0.5pt}},
    },                      
  }

\usepackage{xurl}
\usepackage{hyperref}
\let\cref\Cref

\usepackage{fontspec}

\newfontfamily\titlefont{texgyreheros}[
  Extension      = .otf,
  UprightFont    = *-regular,
  BoldFont       = *-bold,
  ItalicFont     = *-italic,
  BoldItalicFont = *-bolditalic,
]
\titleformat*{\section}{\Large\titlefont\bfseries}
\titleformat*{\subsection}{\large\titlefont\bfseries}
\titleformat*{\subsubsection}{\normalsize\titlefont\bfseries}
\renewcommand{\title}[1]{\newcommand{\titlelist}{{\huge\titlefont\bfseries #1}}}

\newcolumntype{L}[1]{>{\raggedright\arraybackslash}m{#1}}
\newcolumntype{C}[1]{>{\centering\arraybackslash}m{#1}}

\newtcolorbox{templatebox}[1]{
    unbreakable, enhanced, colback=white, colframe=gray!80!black,
    colbacktitle=gray!80!black, coltitle=white, fonttitle=\bfseries,
    title=#1, arc=3mm, boxrule=1pt, drop fuzzy shadow={gray!50!white},
    left=5mm, right=5mm, top=3mm, bottom=3mm
}

  \newtcolorbox{takeaway}[1]{
      unbreakable, enhanced, colback=titlebg, colframe=accentblue, boxrule=1pt,
      arc=3mm, drop fuzzy shadow={accentblue!25},
      left=8pt, right=8pt, top=13pt, bottom=5pt,
      before skip=10pt, after skip=8pt, fontupper=\small,
      coltitle=accentblue, colbacktitle=white, fonttitle=\small\bfseries,
      attach boxed title to top left={xshift=10pt, yshift=-\tcboxedtitleheight/2},
      boxed title style={colframe=accentblue, boxrule=0.8pt, arc=2pt,
                         left=5pt, right=5pt, top=2pt, bottom=2pt},
      title={#1}
  }

\title{%
  MIRA
  \\[0.1em] {\large\color{black!70} Multiplayer Interactive World Models with Representation Autoencoders}%
}

\renewcommand\authorlist{%
\vskip 0.4cm
{\sffamily\bfseries Core contributors}\\
Anthony Hu\textsuperscript{$\ast$,1}, Václav Volhejn\textsuperscript{$\ast$,2}, Adrien Ramanana Rahary\textsuperscript{$\ast$,2,$\ddagger$}, Chris Mulder\textsuperscript{$\ast$,1}, Aditya Makkar\textsuperscript{1}, Alyx Liao\textsuperscript{2}, Amélie Royer\textsuperscript{2}, Manu Orsini\textsuperscript{2}\\[0.5em]
{\sffamily\bfseries Contributors}\\
Adam Jelley\textsuperscript{1}, Eloi Alonso\textsuperscript{1}, Florian Laurent\textsuperscript{1}, Fredrik Norén\textsuperscript{1}, James Swingos\textsuperscript{1}, Jan Hünermann\textsuperscript{1}, Kent Rollins\textsuperscript{1}, Lucas Hosseini\textsuperscript{1}, Matthieu Le Cauchois\textsuperscript{1}, Maxim Peter\textsuperscript{1}, Pim de Witte\textsuperscript{1}, Tim Brown\textsuperscript{1}, Vincent Micheli\textsuperscript{1}, Moritz Böhle\textsuperscript{2}, Gabriel de Marmiesse\textsuperscript{2}, Viktoriia Sharmanska\textsuperscript{3}, Lucia Specia\textsuperscript{3}, Michael Black\textsuperscript{3}, Patrick Pérez\textsuperscript{2}\\[0.5em]
{\small \textsuperscript{$\ast$}\,\textit{equal contribution}}\\[0.3em]
{\small\bfseries \textsuperscript{1}General Intuition\quad\textsuperscript{2}Kyutai\quad\textsuperscript{3}Epic Games}%
}

\abstract{%
We introduce the first multiplayer world model for highly dynamic environments governed by complex physical interactions. Whereas single-player world models treat the other agents as part of the environment, ours conditions on the action streams of multiple agents, learning to attribute changes in the scene to the correct player and to stay coherent under arbitrary combinations of their actions. We study this problem in the game of Rocket League, where players compete and cooperate under fast, tightly coupled dynamics. Trained on 10,000 hours of gameplay collected with publicly available bots, our 5-billion-parameter latent diffusion model generates four-player matches in real time, producing 20 frames per second on a single Nvidia B200 GPU. Although trained only on short clips, its rollouts stay stable far beyond the training horizon: distributional quality holds steady out to five minutes, the longest horizon we measure, and in practice we observe rollouts continuing for hours with no sign of collapse. We systematically investigate the central design choices: the video codec, the generative objective, and the multiplayer conditioning scheme. In addition, we characterize how behavior changes with model and data scale, including the capabilities that emerge and the failure modes that persist. We further develop targeted evaluations that probe the model's physical understanding rather than visual appearance alone. To support continued research on multiplayer world models, we release our dataset, our full training and inference codebase, and a live demo.
}

\metadata[Live demo]{\url{https://mira-wm.com}}
\metadata[Blog]{\url{https://mira-wm.com/blog-post}}
\metadata[Code]{\url{https://github.com/mira-wm/mira}}
\metadata[Dataset]{\url{https://huggingface.co/datasets/kyutai/rocket-science}}
\begin{document}
\maketitle
{\renewcommand{\thefootnote}{$\ddagger$}\footnotetext{École nationale des ponts et chaussées}}

\begin{figure}[h!]
  \centering
  \setlength{\abovecaptionskip}{4pt}
  \includegraphics[width=0.95\linewidth]{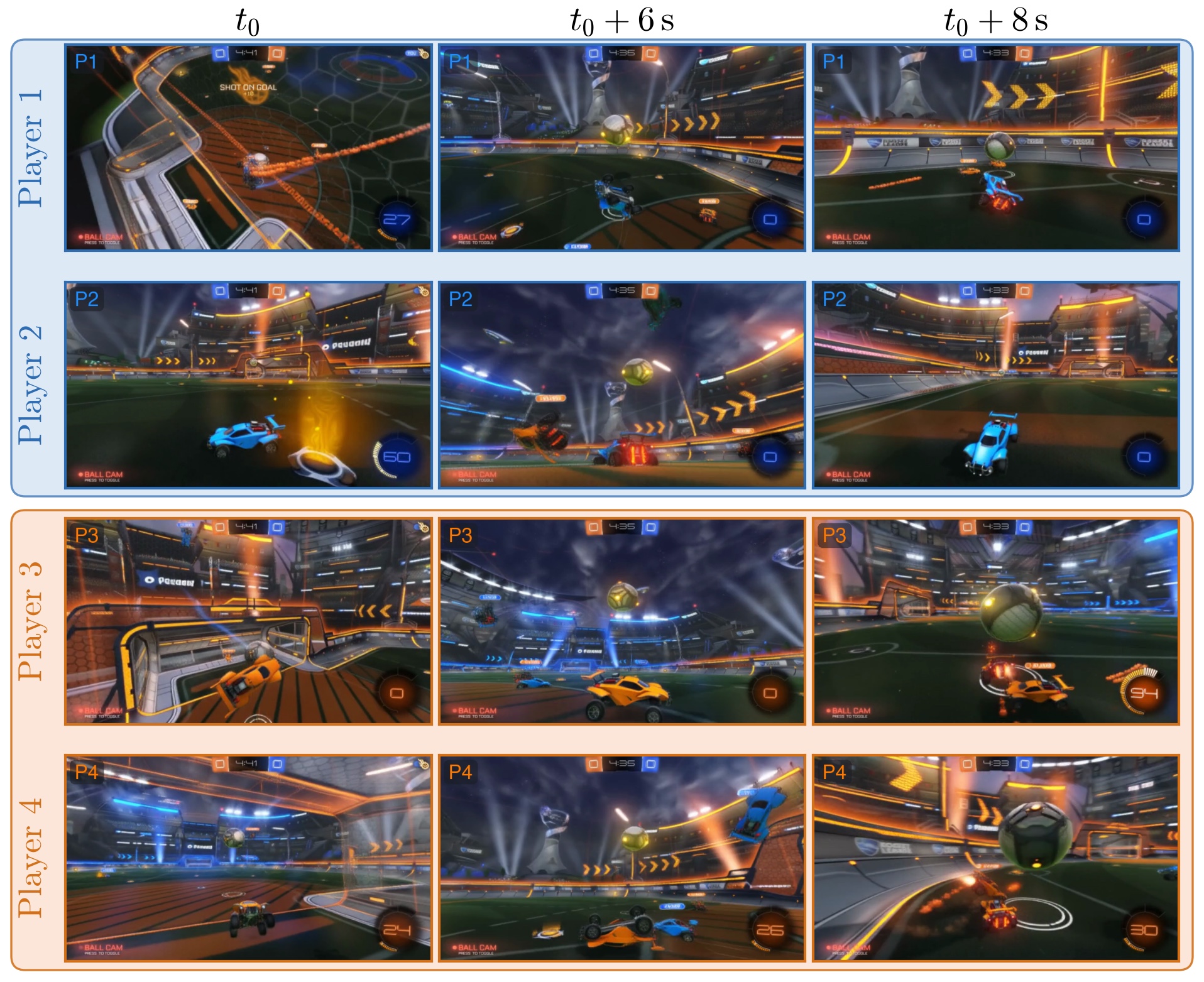}
  \caption{\textbf{World model imagination.} Rows
  are the four players' viewpoints (Players~1--4) and columns show three
  timesteps in the trajectory ($t_0 \to t_0+6\,\text{s} \to t_0+8\,\text{s}$).
  From the initial state at $t_0$ and the players' action
  streams, the model imagines a dynamic game scene that captures the physical interplay
  between the players, the ball, and the scene. The rollout is temporally consistent: the in-game clock counts down with elapsed time
  ($4{:}41 \to 4{:}35 \to 4{:}33$), and the players' views stay mutually
  coherent. For instance, Players~2 and~4 at $t_0+6\,\text{s}$, who are contesting the ball near the goal, each render the
  ball and the cars in a matching spatial configuration. This is best
  experienced interactively: try our live demo at \url{mira-wm.com}.}
  \label{fig:rollout1}
\end{figure}

\section{Introduction}
\label{sec:intro}

Predicting how a scene will evolve in response to actions is a core ability of any embodied agent \citep{lecun2022path, silver2025welcome}. A world model learns this ability directly from experience, typically by building an internal latent representation of the environment and predicting future latent states conditioned on actions. Once learned, the world model can act as a controllable simulator \citep{ha2018world, micheli2023transformers, russell2025gaia2, hafner2025nature}: agents can be trained inside it, evaluated against it, and used to imagine the consequences of candidate actions before committing to them in the real environment.

However, for such imagined rollouts to be useful, the world model must be faithful to the environment it represents \citep{alonso2024diffusion}. As we ultimately aim to build agents that operate in real environments, \mbox{limitations} in world model fidelity can directly hinder downstream agent performance. In particular, the model must capture the causal relationships between actions and their consequences, and preserve the dynamics that determine how the environment evolves. A model that produces visually plausible futures but fails to respond correctly to interventions may be useful as a video predictor, but is insufficient as a substrate for planning and control \citep{quevedo2025worldgymworldmodelenvironment, geminirobotics2025veo}.

This challenge becomes even more important in multi-agent environments. While most real-world environments are inherently multi-agent, the majority of existing world models adopt a single-agent perspective, representing other agents merely as part of the environment. This design limits their applicability to learning paradigms that depend on explicit control over multiple participants, such as self-play reinforcement learning \citep{schrittwieser2020mastering}, multi-agent training \citep{albrecht2024marl}, counterfactual policy evaluation \citep{wayve2025gaia3}, and human-in-the-loop interaction \citep{abramson2022improving}. A multi-agent world model should instead condition on the action streams of multiple agents and predict how their joint behavior shapes the future state of the environment.

\newpage

Extending world models from the single-agent to the multi-agent setting takes more than adding inputs. Conditioning on multiple action streams is strictly more informative than conditioning on a single action stream, but it also imposes a stronger requirement on the model: it must learn to attribute changes in the scene to the correct agent, represent interactions between agents, and remain coherent under arbitrary combinations of actions. This setting introduces potential failure modes beyond inaccurate dynamics: the model can mis-assign agency, ignore the actions of one player, or entangle the effects of multiple players. Faithful multi-agent prediction therefore requires both accurate visual reconstruction and correct action-conditioned dynamics.

In this technical report, we study multi-agent world modeling in Rocket League, a visually rich, physically grounded, and strategically multi-agent environment. Although Rocket League is a video game and thus itself simulated, it serves as a useful proxy for world modeling research by making large-scale data collection and precise action logging feasible in a setting with fast, interactive dynamics that remain difficult to predict from pixels alone. Its appeal as a testbed lies in the combination of discrete high-frequency actions, rich object interactions, partial observability from rendered observations, and tightly coupled coordination and competition between players. We collect 10,000 hours of multiplayer data generated by reinforcement learning policies \citep{nexto} in private matches. Each recording pairs the rendered video stream with the controller actions of every player; we also log privileged game state, such as ball and car physics, for downstream evaluation of the model's ability to learn physically meaningful representations from pixels and actions alone. On this data, we train a multiplayer diffusion world model with 5 billion parameters. The model operates in the latent space of a video representation codec and runs in real time, generating 20 frames per second on a single Nvidia B200 GPU.

\begin{figure*}[t]
    \centering
    \resizebox{\linewidth}{!}{\input{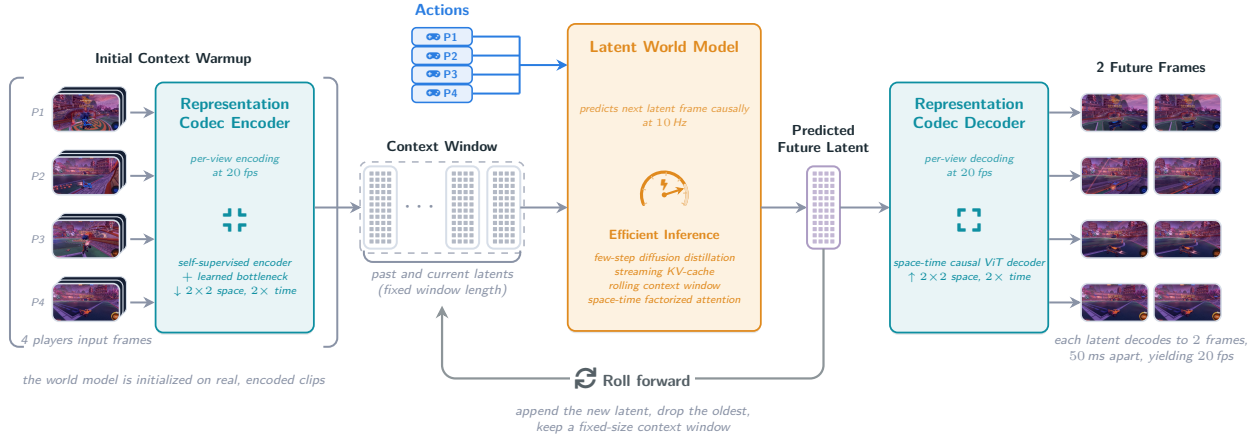}}
    \caption{\textbf{Method overview.} MIRA simulates four-player Rocket League in the latent space of a video representation codec. During an initial context warmup, the codec encoder maps a short clip of each player's view into a stack of latent frames that seed a fixed-length context window (\cref{sec:latent}). Conditioned on the per-player action streams (\cref{sec:multiplayer}) and the latents currently in the context window, the latent world model predicts the next latent frame causally at $10$\,Hz (\cref{sec:objective,sec:architecture}). Few-step diffusion distillation, a streaming KV-cache, and a rolling context window keep inference fast enough to generate $20$ frames per second on a single Nvidia B200 GPU (\cref{sec:streaming_inference}). The codec decoder then renders each predicted latent back into the four player views at $20$\,fps (\cref{sec:latent}). Appending the predicted latent to the context window and dropping the oldest (\emph{roll forward}) autoregressively extends the rollout over long horizons (\cref{sec:streaming_inference}).}
    \label{fig:main}
\end{figure*}

A systematic study of our design choices yields several observations. For the codec, building the latent space on a pretrained feature extractor makes it markedly better suited to world modeling, improving both the quality and the temporal consistency of rollouts, while matching the reconstruction quality of a feature extractor trained from scratch (\cref{sec:exp-latent}). For world modeling, diffusion forcing and few-step distillation together yield stable long-horizon predictions and fast inference (\cref{sec:exp-objective,sec:streaming_inference}). Multiplayer world models substantially outperform their single-player counterparts, indicating that the additional views and action streams reduce uncertainty about the future and yield more accurate predictions; warm-starting a multiplayer model from a single-player model works better than training on multiplayer data directly (\cref{sec:exp-multiplayer}). Across training runs, we observe a consistent progression: the model reaches visually plausible frames early and then spends most of the remaining training refining how actions map to their consequences. We also examine how performance scales as we vary model size from 100M to 5B parameters and training data from 100 to 10,000 hours, uncovering a range of emergent capabilities and characteristic failure modes (\cref{sec:exp-scaling,sec:exp-emergent,sec:exp-failure}).
In summary, our contributions are as follows.
\begin{itemize}
    \item We present the first interactive multiplayer world model for a highly dynamic environment with complex physical interactions. The model is controllable by four concurrent agents, can be rolled out over long horizons, and runs in real time at 20 fps (\cref{sec:method,sec:streaming_inference}).
    \item We conduct a systematic study of the latent space (\cref{sec:exp-latent}), the generative objective (\cref{sec:exp-objective}), and the multiplayer conditioning scheme (\cref{sec:exp-multiplayer}), and we characterize scaling behavior over model and data size, together with the emergent capabilities and failure cases that arise (\cref{sec:exp-scaling,sec:exp-emergent,sec:exp-failure}).
    \item We propose a comprehensive evaluation protocol that measures physical understanding, visual fidelity, and human preference across the models we develop (\cref{sec:metrics}).
    \item To foster future research on multiplayer world models, we release our training data (\cref{sec:data}), together with our full training and inference codebase. A live demo of our model is also available at \\ \url{https://mira-wm.com}.
\end{itemize}

\section{Related Work}
\label{sec:related}

%\subsection{Overview}
% mle: I don't think you necessary need a subtitle since it counds as introduction of the section
We model Rocket League, a fast, physically dynamic multiplayer game, as a real-time
generative simulator that predicts future video frames from several players' actions.
This places our work in the world-modeling literature, which spans two formulations:
an abstract latent dynamics model for control that never renders the world
(\Cref{subsec:rw-control}), and a controllable video generator whose frames are
themselves the objective (\Cref{subsec:rw-games}). Our approach is the latter, in one of its most challenging cases: the multiplayer setting where several players continually
reshape a shared world (\Cref{subsec:rw-multiplayer}). We then review the techniques
these models build on: generative video backbones (\Cref{subsec:rw-video}), the choice
of prediction space (\Cref{subsec:rw-representations}), objectives for stable real-time
rollouts (\Cref{subsec:rw-rollout}), and the evaluation of generative world models
(\Cref{subsec:rw-eval}).

\subsection{World models for control}\label{subsec:rw-control}
A world model predicts how an environment evolves under actions. One line uses this
for control: planning or training an agent inside the model.
\citet{ha2018world} train a recurrent latent model and evolve a controller within
its imagined rollouts. The Dreamer family learns recurrent state-space models and
optimizes behavior through imagination, progressing from latent-space planning
\citep{hafner2019planet, hafner2020dream} to discrete latent representations
\citep{hafner2021mastering}, a single configuration that masters diverse domains
\citep{hafner2025mastering}, and a scalable transformer world model that trains
agents inside it \citep{hafner2025dreamerv4}. Model-based imitation learning trains
a driving policy inside a learned latent world model \citep{hu2022mile}.
Transformer-based world models improve sample efficiency
\citep{micheli2023transformers, micheli2024efficient, robine2023transformer,
zhang2023storm}, and latent model-predictive control operates in decoder-free
latent spaces \citep{hansen2022temporal, hansen2024tdmpc2}. Joint-embedding
predictive architectures (JEPA) predict in representation space rather than pixels
\citep{lecun2022path, assran2023self, bardes2024revisiting, maes2026leworldmodel},
including action-conditioned variants for robotics \citep{assran2025vjepa2} and
approaches built on frozen self-supervised features \citep{zhou2025dino}. These
methods prioritize accurate dynamics in compact state spaces over visual fidelity;
pixels, when reconstructed at all, only probe or learn the representation, and JEPA
omits rendering entirely. Our model also favors a compact prediction space, but is
generative, decoding that latent into the video frames on which it is evaluated.

\subsection{Interactive and playable game world models}\label{subsec:rw-games}
Generative world models instead predict observations directly. At the least
interactive end, they keep the dynamics fixed while synthesising novel views as
the camera moves: navigation world models
\citep{koh2021pathdreamer, bar2025navigation}, monocular novel-view synthesis
\citep{liu2021infinite, li2022infinitenaturezero, liu2023zero1to3,
sargent2024zeronvs, rahary2026oneview}, camera-controlled video diffusion
\citep{he2024cameractrl}, and persistent explorable worlds generated from a single
image \citep{worldlabs2024, worldlabs2025rtfm} or a text prompt \citep{wang2026worldgen}. By
producing 3D scenes that are traversable and support navigation and interaction, these
approaches emphasize perceptual and geometric consistency without modeling scene
dynamics.

Interactive game world models go further, generating frames whose contents change
with each action, requiring both visual realism and dynamical correctness.
\citet{valevski2025diffusion} simulate DOOM in real time from past frames and
player actions, while \citet{bruce2024genie} learn a latent action space from
unlabeled video to generate controllable worlds, later scaled to 3D scenes from a
single prompt image \citep{parkerholder2024genie2, deepmind2025genie3}.
\citet{alonso2024diffusion} train agents inside diffusion world models and deploy
them as standalone engines for Atari and CS:GO. Further systems generate playable
game video in Minecraft
\citep{decart2024oasis, guo2025mineworld, zhang2025matrixgame}, in open-world games
\citep{che2025gamegenx, feng2024matrix}, and in AAA titles such as Grand Theft Auto
\citep{he2025matrixgame2, li2025hunyuangamecraft}. These build on earlier learned
simulators that rendered the next screen directly from controller input
\citep{kim2020learning, menapace2021playable}, which \citet{yu2025position}
frame as a path toward generative game engines. These systems are driven by a
single action stream, that of one player or camera. The same recipe extends to
driving: GAIA generates driving video conditioned on action and text
\citep{hu2023gaia, russell2025gaia2, wayve2025gaia3}, related diffusion models
produce controllable future driving video for prediction and planning
\citep{wang2024drivedreamer, gao2024vista, yang2024genad}, and autoregressive
video pretraining yields driving representations for a downstream action model
\citep{bartoccioni2025vavim}.
Games and driving alike are modeled from a single agent's controls; our model
instead conditions on the simultaneous actions of multiple players.

\subsection{Multiplayer and multi-agent world models}\label{subsec:rw-multiplayer}
The dynamics are hardest to model when multiple agents jointly reshape a shared
environment, the regime our work targets. The World and Human Action
Model \citep{kanervisto2025world}, trained on four-versus-four gameplay, conditions on a single player's
controls and treats all other players as part of the environment, learning multiplayer dynamics through one
control interface. More recent generative approaches condition on several agents
at once: in two-dimensional grid worlds with up to seven agents
\citep{pondaven2026actionparty}; in Minecraft with two players or with per-agent
first-person views generalized beyond two players
\citep{savva2026solaris, liu2026gammaworld}; across multiple camera views of
cooperative play and robot manipulation \citep{wu2026multiworld}; in DOOM levels
anchored to editable two-dimensional maps \citep{po2026multigen}; from precomputed
trajectories in real-world video \citep{hu2026metaworld}; and through per-entity
language prompts over shared combat views \citep{zhu2026incantation}. These works
condition generative world models on multiple agents, but focus on grid worlds,
voxel environments, planar navigation, or combat animation, where rigid-body
interaction is limited, and many treat scaling the agent count as the central
challenge. Our setting, Rocket League, is among the most dynamic and interactive:
a continuous-physics, fully three-dimensional sports game where cars and a ball
collide at speed. We model it in real time, jointly
predicting each of the four players' own view of the shared match and conditioning
on their simultaneous low-level controls.

\subsection{Generative video models and diffusion}\label{subsec:rw-video}
The visual backbone of these world models comes from generative video modeling.
Denoising diffusion made high-fidelity image synthesis practical
\citep{ho2020denoising}, and latent diffusion moved generation into a learned
latent \citep{rombach2022high}; flow matching and rectified flow recast this as a
continuous transport between noise and data, often with a
simpler objective \citep{lipman2023flow, liu2023rectified}. The transformer is the
scalable architecture for both: the diffusion transformer
\citep{peebles2023scalable} and its interpolant counterpart \citep{ma2024sit}.
Extending these processes across time yields video diffusion
\citep{ho2022video, ho2022imagen, singer2023makeavideo}, and large latent video
models now reach strong fidelity over long durations \citep{blattmann2023stable,
yang2025cogvideox, bartal2024lumiere, gupta2024photorealistic, kong2024hunyuanvideo,
hacohen2025ltxvideo, hacohen2026ltx2}; at scale such
models have been described as general simulators of the visual world
\citep{brooks2024video}, while token-autoregressive models offer an alternative
\citep{kondratyuk2024videopoet}. These models produce realistic video,
but are neither causal nor action-controllable by default, the two properties
a world model needs. We therefore adopt a latent video diffusion backbone and make
it causal and action-conditioned.

\subsection{Visual representations for generation}\label{subsec:rw-representations}
Beyond the choice of backbone, two design questions cut across these world models.
The first concerns the representation space in which prediction occurs. Predicting in a
compact learned latent rather than in pixels makes the problem lower-dimensional,
removes the burden of synthesizing unpredictable high-frequency texture, and, when
the latent is well-structured, keeps predictions stable over long rollouts. Latent diffusion established
this paradigm by compressing images into a latent through an autoencoder and
training the diffusion process in that latent \citep{rombach2022high}. Generation
quality then depends on the structure of the latent, motivating work to improve it. One direction distills a pretrained self-supervised encoder such
as DINO \citep{caron2021emerging, oquab2024dinov2, simeoni2025dinov3} into the
latent space of an autoencoder: VA-VAE and the GAIA driving models both align that
latent to DINO features, in an image autoencoder and a video tokenizer
respectively \citep{yao2025reconstruction, hu2023gaia, russell2025gaia2}, while
earlier BEiT v2 distills a pretrained teacher into a vector-quantized tokenizer
\citep{peng2022beit2}. This principle carries over to other modalities:
in audio, the Mimi codec~\citep{defossez2024moshi} distills a pretrained WavLM encoder~\citep{chen2022wavlm}
into its latent, and SpeechTokenizer~\citep{zhang2023speechtokenizer} distills a self-supervised speech
teacher into its tokens. A related
direction regularizes the latent through self-supervised objectives that encourage
semantic structure, denoising, or equivariance
\citep{chen2025maetok, chen2025dcae15, yang2025ldetok, kouzelis2025eqvae}. At the
extreme, some approaches remove the learned encoder entirely and generate in the
feature space of frozen self-supervised encoders such as DINO \citep{caron2021emerging}
or SigLIP \citep{zhai2023sigmoid, tschannen2025siglip2}, training only the decoder
\citep{zheng2025rae, bi2025vfmvae, gui2025reptok}. This
representation-autoencoder paradigm has since been extended to text-to-image
generation \citep{tong2026scalerae}, simplified \citep{singh2026raev2}, and reduced
to lightweight adaptation layers over frozen encoders \citep{gao2025fae}.
Our codec follows the principle these works share, that representation quality
matters for prediction, and builds its prediction space on DINOv3, a strong
self-supervised encoder \citep{caron2021emerging, oquab2024dinov2, he2022masked,
simeoni2025dinov3}.

Latent prediction is also long established for world models. The recurrent world
model of \citet{ha2018world}, the Dreamer family
\citep{hafner2020dream, hafner2025mastering}, and transformer world models
\citep{micheli2023transformers} all predict within the latent of a learned
variational or discrete autoencoder, while \citet{zhou2025dino} predict future
frozen DINO features for planning. Prediction in pixel space remains competitive
in some domains, for image generation with pixel-space transformers
\citep{li2025jit} and in some world models used to train policies, where fine
visual detail matters \citep{alonso2024diffusion}, so the best prediction space remains an open question. We
therefore treat the prediction space as a design choice and study it empirically in
Section~\ref{sec:exp-latent}.

\subsection{Long-horizon, real-time rollout and the training objective}\label{subsec:rw-rollout}
The second design question concerns keeping a rollout coherent over a long horizon
while producing it in real time. Rolling out autoregressively exposes a mismatch
between training, where the model conditions on clean ground-truth frames, and
inference, where it conditions on its own predictions, so small errors compound
into drift. This exposure bias has long been studied in sequence modeling
\citep{bengio2015scheduled, lamb2016professor}. Several schemes reduce it for
video: diffusion forcing assigns each frame an independent noise level so the
model trains under partially corrupted context \citep{chen2024diffusion}, history
guidance conditions on a variable number of past frames \citep{song2025history},
self-forcing unrolls the model on its own samples during training
\citep{huang2025self}, and sliding-window noise schedules limit accumulation over
long videos \citep{ruhe2024rolling, kim2024fifo, liu2025rolling}.
Few-step training objectives such as consistency and shortcut models
\citep{song2023consistency, song2024improved, luo2023latent, boffi2025consistency, frans2025one}
reduce
the number of sampling steps a model needs, which matters because an interactive
simulator must render each frame before the next action arrives; our model
predicts at $10$\,Hz in latent space and renders video at $20$\,fps. We compare
diffusion forcing and teacher forcing within our
world model and measure their long-horizon stability in
Section~\ref{sec:exp-objective}.

\subsection{Evaluating generative world models}\label{subsec:rw-eval}
Evaluating generative world models requires more than visual fidelity. The
Fr\'echet Inception Distance and Inception Score capture per-frame realism
\citep{heusel2017gans, salimans2016improved}, and the Fr\'echet Video Distance
extends it to short clips \citep{unterthiner2018towards}, all distributional
metrics that need no aligned reference. Reference-based metrics such as PSNR, SSIM
\citep{wang2004image}, and LPIPS \citep{zhang2018unreasonable} instead score each
frame against a ground-truth target, which suits short rollouts with a known
reference but not the many plausible futures of a stochastic world model. Recent
benchmarks score video generation along human-aligned axes such as temporal
consistency and motion quality \citep{huang2024vbench}, probe physical plausibility
\citep{motamed2025physicsiq, bansal2024videophy, meng2025phygenbench}, and judge
video models directly as world models
\citep{duan2025worldscore, li2025worldmodelbench}, though they target text-to-video
generation rather than action-conditioned simulation.
A standard alternative probes frozen features with lightweight readout models to
recover interpretable state variables
\citep{alain2017understanding, li2023emergent, maes2026leworldmodel}; we adapt this
to recover game-state quantities and measure physical consistency. The same probing
idea extends to actions: playable video generation predicts the action from generated
motion \citep{menapace2021playable}, and we build on this to introduce the
Action Recoverability Ratio (ARR), a metric that measures how faithfully the rollout
obeys the commanded actions (\Cref{sec:metrics}).
None of these metrics captures slow temporal drift over long autoregressive
rollouts, a failure mode highlighted by recent work on video generation
\citep{chen2024diffusion, yin2025causvid, liu2025rolling}. We therefore complement
visual fidelity with metrics for physical consistency and long-horizon stability,
validated against human judgment (\Cref{sec:metrics}).

\section{Data}
\label{sec:data}

We train on $10{,}000$ match-hours of recorded 2v2 Rocket League matches,
generated entirely by self-play between game-playing bots. We did not use any human data to build this dataset. Each match yields four
synchronized first-person recordings, one per player. Every recording pairs
the gameplay video with the player's action stream and the underlying physics state,
all aligned to the image frame boundaries on a shared timeline. We first describe how the
data is produced and recorded (\Cref{sec:data-collection}) and then the physics
game state recorded alongside each frame (\Cref{sec:data-physics}); the
processing that turns the raw per-player recordings into the frame-aligned chunks
the model trains on is detailed in \Cref{sec:data-processing}.

\subsection{Data collection}
\label{sec:data-collection}

\paragraph{Game environment.} We restrict the game to its 2v2 mode and to
three arenas chosen for visual and layout diversity: Champions Field, Forbidden
Temple, and Deadeye Canyon. Every car in
every match is driven by Nexto \citep{nexto}, the highest-skilled publicly
available Rocket League bot. Nexto is a state-based policy trained by self-play
reinforcement learning in RLGym \citep{rlgym}. It observes the privileged
game state, meaning all of the game's internal parameters rather than the pixels.
Fifteen times per second, once every $8$ game ticks, it emits three-valued
axis commands $\{-1, 0, 1\}$ for steer, throttle, yaw, pitch, and roll, together
with binary jump, boost, and handbrake actions. Using a single bot
to drive all four cars makes the matches reproducible and removes any dependence
on human players, at the cost of behavioral diversity, which we discuss
later in the section.

\paragraph{Match generation.} Each game instance runs a custom
BakkesMod \citep{bakkesmod} plugin that both records the session and bridges the
game to the bot: it extracts the game state, passes it to the Nexto process, and
applies the actions the bot returns. A centralized orchestrator pools virtual
machines (VMs) into matches and assigns teams. One machine is designated host and
starts a LAN match; the next three machines in the queue are told to join it, and
once all four players are present the match begins and runs to completion. A
single match is the atomic unit of data: matches in which any VM had problems
(network lag, recording hiccups, or a failure to play to completion) are
discarded before they reach the dataset.

\paragraph{Recorded signals.} For each player we record three time-synchronized
streams, summarized in \Cref{tab:data-streams}: the gameplay video, the game
state (ball and per-car physics, scores), and the bot's
actions. Actions are captured at three layers, so downstream consumers can choose
the level they need: (a)~the policy's intended action (the raw Nexto output);
(b)~the mapped keypresses we actually issue to the game; and (c)~the action the
engine ultimately consumed. The world model trains on layer (b); the other two
serve as sanity checks. Nexto's outputs are translated to a canonical nine-key keyboard
vocabulary: its three-valued ($\{-1, 0, 1\}$) axis commands drive the six directional
keys (\texttt{W},\,\texttt{A},\,\texttt{S},\,\texttt{D},\,\texttt{Q},\,\texttt{E}), and its
binary outputs drive \texttt{SPACE},\,\texttt{LSHIFT},\,and\,\texttt{LCTRL},
whose in-game meaning depends on whether the car is grounded or airborne; the
keybindings and the full policy-to-keyboard mapping are given in
\Cref{app:data}. All four per-player recordings from a match share a common
match identifier and are aligned on the shared kickoff-countdown anchor, so they
stitch together cleanly. Processing resamples these raw streams into the $20$\,fps video and per-frame action labels the model trains on (\Cref{sec:data-processing}).

\begin{table}[htbp]
\centering\small
\setlength{\tabcolsep}{8pt}\renewcommand{\arraystretch}{1.2}
\caption{\textbf{Recorded streams (per player).} Each match produces four of these streams, one per player.}
\label{tab:data-streams}
\begin{tabular}{l l l}
\toprule
Stream & Contents & Sample rate \\
\midrule
Video     & Per-player gameplay ($720$p)                  & $30$\,fps \\
Game state & Ball $+$ per-car physics, scores, events      & $120$\,Hz (physics tick) \\
Actions   & Nexto output: $3$-valued axes $+$ buttons     & $15$\,Hz (every $8$ ticks) \\
\bottomrule
\end{tabular}
\end{table}

\paragraph{Scale.} After processing, the training set comprises $\approx 10{,}000$
match-hours of clean gameplay ($82{,}983$ matches, or $331{,}932$ per-view
recordings), evenly split across the three arenas. The full per-map breakdown is given in \Cref{app:data} (\Cref{tab:data-scale}).

\paragraph{Limitations.}
\label{sec:data-limitations}
Every car is driven by its own independent instance of the same bot policy: each
VM runs its own game instance with its own Nexto process controlling a single car,
so the four cars in a match are four separate instances of the same policy rather
than one multi-agent controller. This limits behavioral diversity, as the dataset
reflects one policy's style of play rather than the full range of human
strategies. Environment diversity is likewise limited: all matches are played on
three fixed maps, whereas games with procedurally generated worlds such as Minecraft
expose a model to far more varied scenes. The setting also carries little stochasticity,
since the game is nearly deterministic once all four players' actions are known, leaving
less predictive uncertainty than partially observed environments.

\subsection{Physics game state}
\label{sec:data-physics}

Alongside video and actions, we log the game's privileged physics state: the
exact internal configuration of the simulation rather than what is visible on
screen. This signal is not consumed by the world model, which trains on pixels
and actions alone; instead it serves as ground truth for evaluation, letting us
probe whether the learned representations encode physically meaningful quantities
such as ball and car dynamics (\Cref{sec:metrics}).

\paragraph{Extraction.} The same BakkesMod plugin that bridges the game to the
bot (\Cref{sec:data-collection}) reads the physics state directly from the game's
internal memory. Rocket League advances its physics at $120$ steps per second,
and we hook the physics step so that, immediately after every tick, we record the
state the engine just computed. Collection is therefore exactly in lockstep with
the simulation: one physics game state per tick at $120$\,Hz, with no
interpolation or resampling at capture time. This is the same extraction that
supplies the game state to Nexto.

\paragraph{Contents.} Each state comprises match-level information (score, time
remaining, overtime flag), the ball's physics (position, velocity, orientation,
and spin), and one entry per car for all four cars on the field (pose, velocity,
boost, and gameplay flags such as ground/wall contact, jump and dodge status).
Positions are in Unreal units in the game's $Z$-up world frame. The full schema,
units, and physical value ranges are given in \Cref{app:physics}.
\section{Method}
\label{sec:method}

We model 2v2 Rocket League with a world model that operates in the latent space of
a representation codec (\cref{fig:main}). After formalizing the prediction problem
(\cref{sec:setup}), we describe the codec and then the world model that runs in its
latent. The codec is a video representation autoencoder built on a frozen self-supervised feature extractor
(\cref{sec:latent}): it compresses frames spatially and temporally into a compact
latent, and decodes that latent back to video. Within this latent space, the world model, a
flow-matching transformer (\cref{sec:architecture}) predicts the future
autoregressively, one latent frame at a time, and the codec decodes the predicted
latents into the video the players see and act on. Training this transformer for stable
long-horizon rollouts (\cref{sec:objective}), modeling all four players jointly
(\cref{sec:multiplayer}), and running it in real time (\cref{sec:streaming_inference})
each raise a distinct problem.

\subsection{Problem setup}
\label{sec:setup}

We consider a game with $P$ players; the two-versus-two matches we study have
$P=4$. Each player $p$ observes their own video stream of the shared match and issues
a stream of actions encoding their keyboard inputs, meaning one video stream and one
action stream per player. A codec compresses each player's video, both spatially and
temporally, into a sequence of latent frames at a lower rate than the video
(\cref{sec:latent}); we index these latent frames by $t$, writing $z_t^{p}$ for
player $p$'s latent at step $t$ and $a_t^{p}$ for that player's actions over the same
step. The world model defines a distribution over all players' next latents given the
past latents and every player's actions,
\[
p_\theta\!\left(z_{t+1}^{1:P} \mid z_{\le t}^{1:P},\, a_{\le t}^{1:P}\right),
\]
which is rolled out autoregressively and decoded back to frames.
We learn $p_\theta$ from recorded play alone, without access to the game engine or
to any privileged state.

\subsection{Codec}
\label{sec:latent}

\begin{figure}[htbp]
  \centering
  \resizebox{\linewidth}{!}{\input{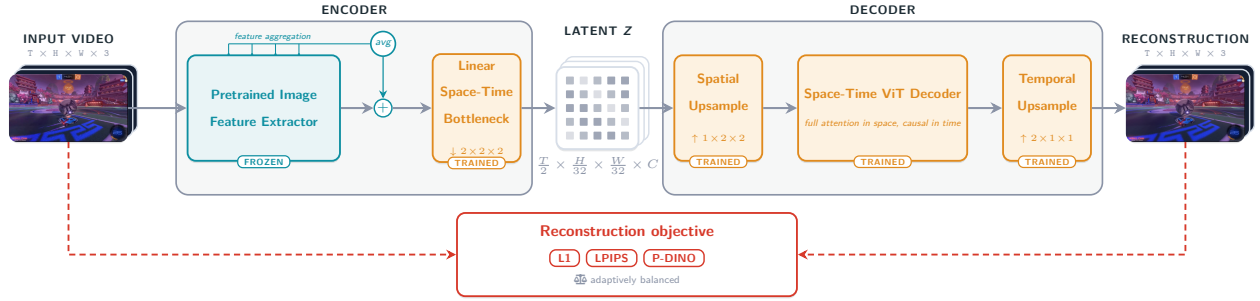}}
  \caption{\textbf{Codec.} A frozen pretrained feature extractor (DINOv3-L) extracts per-frame features,
  mixed across several intermediate layers, and a learned linear bottleneck downsamples them
  by $2\times2$ in space and $2\times$ in time and projects to a latent $z$ with $C$ channels.
  The decoder reverses these steps: a spatial upsampling restores the spatial resolution, a
  causal spatio-temporal vision transformer reconstructs the video, and a temporal upsampling
  restores the frame rate.}
  \label{fig:codec}
\end{figure}

Predicting in pixel space is expensive, spends capacity on high-frequency detail
unrelated to dynamics, and is too slow to roll out in real time. We instead predict
in the latent space of a \emph{representation codec}: an autoencoder whose latent is built on a frozen
self-supervised feature extractor, so the prediction space inherits strong, semantically
meaningful features instead of being learned from scratch, following the
representation-autoencoder idea \citep{zheng2025rae}.
The codec must preserve the
feature extractor's representation, stay compact enough for real-time prediction, and decode
back to sharp video.

\paragraph{Pretrained feature extractor.} The feature extractor is a frozen DINOv3-L/16
\citep{simeoni2025dinov3}, applied per frame. Following RAEv2 \citep{singh2026raev2}, the codec reads several of
DINOv3's intermediate blocks, not its final layer alone. Writing $f_\ell$ for the patch
features at block $\ell$ and $\mathcal{S}$ for the chosen blocks, we aggregate
\[
\bar{f} = \frac{1}{|\mathcal{S}|}\sum_{\ell \in \mathcal{S}} f_\ell + f_{\ell_{\max}},
\]
the mean over the selected blocks plus the deepest of them
($\ell_{\max} = \max\mathcal{S}$): the earlier blocks supply spatial and low-level
detail that the deepest layer has abstracted away, while $f_{\ell_{\max}}$ retains its
semantics. The feature extractor stays frozen; we train only the bottleneck and the decoder.

\paragraph{Bottleneck.} A learned \emph{linear} bottleneck maps $\bar{f}$ to the latent
with a single $2\times2\times2$ patchifier and reduces the channel dimension from 1024 to 32.
Building on DINOv3's $16\times16$ patches, the bottleneck coarsens the spatial grid by a
further $2\times$ per side, so each latent token covers a $32\times32$ pixel region; its
temporal patchifier halves the rate, so the bottleneck emits latents at $10$\,Hz from the
$20$\,fps video, about a $192\times$ reduction in the number of values
relative to the raw RGB frames. This departs from a standard representation autoencoder, which
keeps the feature extractor's grid, frame rate, and dimension: our single learned linear
bottleneck instead compresses along two axes at once. It downsamples in space and time,
so the shorter latent sequence makes real-time rollout feasible, and it reduces the latent
dimension, staying linear so it compresses the features without distorting them.

\paragraph{Decoder.} A vision transformer decoder with factorized spatial
(bidirectional attention within a frame) and temporal (causal: each token attends only to the same spatial position in earlier frames) attention reconstructs the video. Its first layer
is an unpatchify layer that upsamples the latent by $2\times2$ spatially, from
the coarse latent grid (one token per $32\times32$ pixels) back to the feature extractor's
finer patch grid (one token per $16\times16$ pixels), so the attention blocks
decode at that finer grid rather than at the compressed latent; this single step is important for
reconstruction quality, and we ablate it in \cref{sec:exp-latent}. Time is upsampled
only at the end, in the unpatchification that expands each latent frame into two video
frames, so the attention runs at the latent's $10$\,Hz rate.

\paragraph{Loss.} We train the codec to reconstruct frames with an L1 loss, an
LPIPS perceptual loss \citep{zhang2018unreasonable}, and a feature-consistency loss
that re-encodes the decoded frames and matches their DINOv3 features, read at
several intermediate layers, to those of the originals,
\[
\mathcal{L} = \lVert x - \hat{x}\rVert_1
  + \lambda_{p}\,\mathrm{LPIPS}(x, \hat{x})
  + \lambda_{d}\,\lVert \phi(x) - \phi(\hat{x})\rVert_2^2,
\]
with $\phi$ the frozen DINOv3-L features stacked over those layers and $\hat{x}$ the
reconstruction. This mirrors LPIPS, which compares reweighted intermediate VGG
activations; using DINOv3 features in this role improves reconstruction and
generation quality in recent autoencoders and pixel-space generators
\citep{hansen2026vitokv2, ma2026pixelgen, rahary2026oneview}. We set the perceptual weights
$\lambda_{p}$ and $\lambda_{d}$ adaptively, reusing the gradient-norm balancing that VQ-GAN
\citep{esser2021taming} applies to its single adversarial term. Our codec has no discriminator; we
apply the same rule to each of our two perceptual losses, giving LPIPS ($\lambda_{p}$) and the
feature-consistency loss ($\lambda_{d}$) their own weights and rescaling each every step so its
gradient magnitude matches that of the reconstruction loss at the decoder's last layer $\theta$,
\[
\lambda = \frac{\lVert \nabla_\theta \mathcal{L}_{\mathrm{rec}}\rVert}
               {\lVert \nabla_\theta \mathcal{L}_{\mathrm{perc}}\rVert + \epsilon},
\]
with $\mathcal{L}_{\mathrm{rec}} = \lVert x - \hat{x}\rVert_1$ the reconstruction loss,
$\mathcal{L}_{\mathrm{perc}}$ either perceptual term, and $\epsilon$ a small constant, so neither
perceptual term dominates training regardless of its raw scale.
We ablate this rule against fixed weights in \cref{sec:exp-latent}. Two
ingredients common in latent autoencoders are deliberately absent. We use \emph{no
adversarial loss}: \citet{hansen2026vitokv2} show that a feature-space objective
can replace the discriminator and train stably at scale, and our codec follows
suit. We also inject \emph{no noise} into the latent and impose no KL penalty: where
representation autoencoders and variational autoencoders regularize the latent with
added noise or a Gaussian prior \citep{zheng2025rae, yang2025ldetok}, our latent is
deterministic, kept faithful by the linear bottleneck and recoverable through the
consistency loss.

We evaluate the codec's design choices, from the feature extractor it builds on to the scale
of the decoder, in \cref{sec:exp-latent}.

\subsection{Generative objective}
\label{sec:objective}

We train the world model with flow matching \citep{lipman2023flow, liu2023rectified}.
For a target latent $z_1$ and a Gaussian noise sample $z_0$, we interpolate
$z_\tau = \tau z_1 + (1-\tau)\,z_0$ and regress the velocity that carries noise to
data,
\[
\mathcal{L} = \mathbb{E}_{\tau, z_0, z_1}
  \bigl\lVert v_\theta(z_\tau, \tau) - (z_1 - z_0) \bigr\rVert_2^2 ,
\]
with flow time $\tau \in [0,1]$. A rollout integrates $v_\theta$ from noise to a latent
over a few steps, conditioned on a bounded window of past latents and the players' actions; \cref{sec:streaming_inference} details how the rollout is run in practice.

\paragraph{Diffusion forcing.} Each frame receives its own flow time $\tau$, drawn independently
\citep{chen2024diffusion}, so a training sequence mixes clean, lightly noised, and heavily noised
frames. This emulates what the model meets at rollout, where it conditions on its own imperfect past
predictions, and keeps small per-step errors from compounding over a long horizon. We ablate it
against teacher forcing, which conditions on fully clean context, in \cref{sec:exp-objective}.

\paragraph{Few-step distillation.} Sampling a frame with standard flow matching takes many sequential
integration steps, which caps the rollout speed, so we distill the pretrained model for few-step
sampling with a progressive self-distillation objective (PSD), following shortcut and consistency
models \citep{frans2025one, song2023consistency, boffi2025consistency}. We condition the velocity
network on a step size $\Delta$ alongside the noise level $s$, so that $v_\theta(z_s, s, \Delta)$
predicts the \emph{average} velocity of the probability-flow ODE over the interval $[s, s{+}\Delta]$,
and a single evaluation advances the sample across the whole interval; setting $\Delta{=}0$ recovers
the instantaneous velocity of the standard flow-matching loss. To keep this average-velocity field
self-consistent, the model maps one large step to two smaller ones: for a sampled pair $s<t$ with
midpoint $u=(s{+}t)/2$, the student $v_\theta(z_s,s,t{-}s)$ regresses onto a stop-gradient two-hop
target,
\[
\tfrac12\big[\,v_\theta(z_s,s,u{-}s) + v_\theta(\hat z_u,u,t{-}u)\,\big], \qquad
\hat z_u = z_s + (u{-}s)\,v_\theta(z_s,s,u{-}s),
\]
which replaces one step with two half-size steps. We mix this term into training either
stochastically or deterministically with a fixed weight, and the distilled model then matches
many-step sampling quality in one or two function evaluations, making real-time rollouts feasible.

\subsection{World model architecture}
\label{sec:architecture}

\begin{figure}[tbp]
  \centering
  \resizebox{\linewidth}{!}{\input{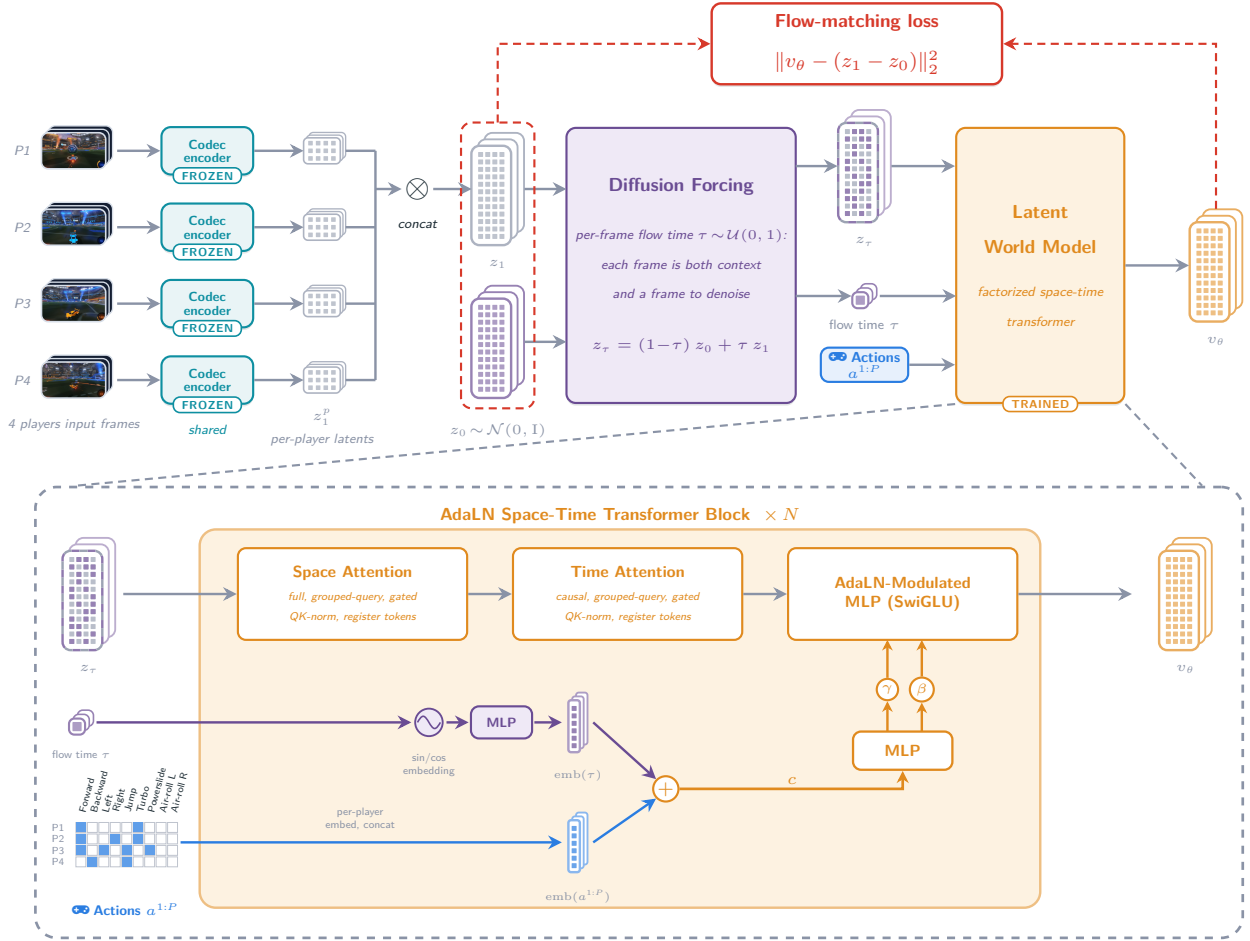}}
  \caption{\textbf{World model training and architecture.}
  \emph{Top:} the training pipeline, read left to right. Each of the $P$ players'
  frames is mapped by a shared, frozen codec encoder
  (\cref{sec:latent}) to a per-player latent $z_1^{p}$, and the $P$ latents are
  concatenated into a single clean joint latent $z_1$. Together with Gaussian noise
  $z_0\!\sim\!\mathcal{N}(0,\mathrm{I})$ of the same shape, \emph{diffusion forcing}
  (\cref{sec:objective}) forms the noised latent $z_\tau=(1-\tau)\,z_0+\tau\,z_1$,
  drawing an independent flow time $\tau\in[0,1]$ for each frame so that every frame
  is at once clean context and a target to denoise. The trained
  latent world model, conditioned on all players' actions $a^{1:P}$ and the flow time
  $\tau$, predicts the flow velocity $v_\theta$, supervised against $z_1-z_0$ by the
  flow-matching loss $\lVert v_\theta-(z_1-z_0)\rVert_2^2$ (\cref{sec:objective}).
  \emph{Bottom:} one of the $N$ stacked AdaLN space-time transformer blocks inside
  the world model. The noised latent $z_\tau$ passes through \emph{space attention}
  (bidirectional, within a frame), \emph{time attention} (causal, across frames), and
  an \emph{AdaLN-modulated} SwiGLU feed-forward layer. Conditioning enters once per
  block: the action embedding $\mathrm{emb}(a^{1:P})$ (a per-player embedding of the
  keyboard controls, concatenated over players) and the flow-time embedding
  $\mathrm{emb}(\tau)$ (a sinusoidal encoding followed by an MLP) are summed into a
  single vector $c$, from which an MLP predicts the scale $\gamma$ and shift $\beta$
  that modulate the feed-forward layer normalization, $(1+\gamma)\,\mathrm{LN}(x)+\beta$;
  the same $c$ is broadcast over every spatial position. Here $p$ indexes the
  $P$ players ($P{=}4$ in our 2v2 setting).}
  \label{fig:worldmodel}
\end{figure}

This subsection specifies the velocity network $v_\theta$ trained by the objective of
\cref{sec:objective}. We describe it for a single player and defer the tiling that
handles all four players to \cref{sec:multiplayer}.

The world model is a diffusion transformer \citep{peebles2023scalable, ma2024sit}
adapted to video with factorized space-time attention
\citep{bertasius2021space, arnab2021vivit} (\cref{fig:worldmodel}). During training it
denoises a whole chunk of latent frames jointly, each frame at its own flow time
(\cref{sec:objective}); at rollout it generates one latent frame at a time,
autoregressively (\cref{sec:streaming_inference}).

\paragraph{Tokenization.} It operates on the grid of continuous latent vectors:
a linear layer embeds each latent vector to the model width; with a patch size of one,
every latent vector is one token. In the multiplayer setting the per-player views
are tiled into a single grid before this embedding (\cref{sec:multiplayer}).

\paragraph{Positional encoding.} We encode position with factorized rotary embeddings (RoPE)
\citep{su2024roformer}: a two-dimensional axial encoding
over the spatial grid and a separate temporal encoding indexed in seconds. Because
RoPE encodes position through relative rotations rather than a learned table of
absolute positions, the transformer carries no positional parameters tied to a fixed
grid size; the same weights run whether a single view or several tiled views fill the
grid, which we use when warm-starting the multiplayer model from the single-player
one (\cref{sec:multiplayer}).

\paragraph{Space-time attention.} Within each block, attention runs in two stages. A spatial layer attends
bidirectionally within a frame, over all token positions, so every part of the scene
informs every other; a temporal layer then attends causally across frames at the same
spatial position, so a token attends only to that same location in earlier frames, seeing
the past but not the future. The blocks otherwise
use components that help large transformers train stably at scale: grouped-query
attention \citep{ainslie2023gqa}, query-key normalization
\citep{henry2020query, dehghani2023scaling}, a sigmoid gate on the attention output
\citep{qiu2025gated}, SwiGLU feed-forward layers \citep{shazeer2020glu}, and learnable
spatial and temporal register tokens \citep{darcet2024vision}. The model is a stack of
these blocks.

\paragraph{Action conditioning.} The model is action-conditioned through adaptive layer
normalization \citep{peebles2023scalable}. At each step we form a
conditioning vector $c$ (dimension of the transformer block) by summing an embedding of the flow time $\tau$
(\cref{sec:objective}) and an embedding of all players' actions
(\cref{sec:multiplayer}), and we broadcast $c$ over every spatial position. Because
actions are recorded once per video frame, at twice the latent rate, a learned linear
layer pools the two action frames spanned by each latent frame into one action
embedding per latent frame. The
conditioning sets the scale and shift of the layer normalization before each
feed-forward layer,
\[
\mathrm{AdaLN}(x, c) = \bigl(1 + \gamma(c)\bigr)\,\mathrm{LN}(x) + \beta(c),
\]
with $\gamma$ and $\beta$ linear in $c$, so the actions and the noise level modulate
the network in every block. A final linear layer maps each token to the flow velocity
the model predicts (\cref{sec:objective}), which we split back into the per-player
views. \cref{fig:worldmodel} illustrates this conditioning path: the action and flow-time
embeddings are summed into $c$, which produces the AdaLN scale and shift $\gamma,\beta$.
The past latents enter the model through the causal temporal attention (the space-time
attention above), while AdaLN carries only the actions and the flow time.

\subsection{Multiplayer conditioning}
\label{sec:multiplayer}

\paragraph{Tiled views.} We stack the four players' latent views into one grid along
its height and predict them together. Spatial attention then spans all the views, so
the model renders the shared world, the ball, the field, and the other cars,
consistently across the four perspectives in a single pass. We find this consistency reflected in
the model's cross-view attention (\cref{fig:mp-attention}).

\paragraph{Per-player actions.} Each player's key presses are embedded and passed
through a per-player network; the four embeddings are concatenated height-wise in a fixed player
order and mixed into the conditioning vector that drives the model through AdaLN
(\cref{sec:architecture}). This order, aligned with the tiling order of the views,
binds each action stream to its player. The model conditions on key presses only.
Because this conditioning vector is broadcast to every latent position, each player's
view is conditioned on all four players' actions, and the model learns which actions
affect a given view and which leave it unchanged, rather than being told the mapping.

\paragraph{Action dropout.} During training we randomly replace a player's action
embedding with a learned absent token, so the model must predict that player from the
scene alone and learns to drive the cars whose actions are withheld. At inference this
lets the world model itself act as the policy for any player the user does not
control, a limited form of agent modeling that needs no separately trained agent.

\paragraph{Two-stage training.} We train the world model in two stages: we first pretrain on
single-player views, then continue on all four players' views jointly, initialized
from the single-player checkpoint, so the model keeps the dynamics it has learned and
only adapts to the joint layout and the per-player conditioning.
\cref{sec:exp-multiplayer} isolates the ingredients of this recipe: a single player
with a single action stream, a single player with the full four-action
representation, four players each with their own action stream, and the warm start
from the single-player model.
The codec is trained in one stage, purely on single-player clips; for the multi-player world model, each player stream is encoded and decoded independently.
\section{Streaming inference}
\label{sec:streaming_inference}

We serve the world model as an interactive online demo: a user plays the game
from a browser with the keyboard, and the model generates the resulting
video on the fly. This imposes a fixed time constraint on the generation of each frame, with a streaming loop that conditions each new frame on the player's most recent actions and on the previously generated frames.
Inference
runs at the rate of play: the model produces a $10$\,Hz latent stream that the codec
temporally upsamples and decodes to $20$\,fps video, and a single Nvidia B200 GPU keeps up with
this rate. This section condenses the design choices that make real-time,
interactive generation possible.

\paragraph{Autoregressive rollout with a streaming cache.} We generate the latent
stream autoregressively with a key-value cache \citep{shazeer2019fast}: each new
latent frame attends to the stored keys and values of the past instead of
recomputing them, so the per-frame cost remains constant as the game progresses. Concretely, the model holds a fixed-size window of $T=20$ latents, rolling the window by 1 each step to append a fresh noise latent which is denoised into the next frame. Similarly the codec's
space-time decoder is conditioned on the three most recent latent frames, enabling constant-cost frame decoding.

\paragraph{Priming the rollout.} A session starts from a short clip of real
gameplay: we encode its frames to latents and prefill the cache, allowing the model to roll out from a coherent, in-distribution context. During prefill, the demo displays the decoded
ground-truth latents; afterwards, during the decode stage, every frame is generated by the model.

\paragraph{Few-step sampling.} The model denoises each new frame in a few steps on a linear-quadratic schedule \citep{polyak2024moviegen}, using the distilled
few-step sampler from \cref{sec:objective}.

\paragraph{Decoupled production and display.} Producing frames and displaying them
run as a producer/consumer pair. The producer advances the model on the GPU worker
thread, copies the decoded frames to the host, JPEG-encodes them off the GPU thread, and pushes them onto a small bounded queue; a separate sender drains that queue at a steady per-frame cadence. Because each world-model step emits two video frames at once, this decoupling turns bursts of two frames per step into evenly spaced delivery. The bounded queue also provides backpressure: when it is full the producer blocks, which paces generation to the display rate and keeps the model from running ahead of what the client can show.

\paragraph{System optimizations.}

We further incorporate system optimizations for real-time serving. We use the TorchInductor backend in PyTorch \citep{10.1145/3620665.3640366} for operator fusion and CUDA graphs to reduce host overhead. We also use different attention backends depending on the stage and type of factorized attention. CuDNN \citep{DBLP:journals/corr/ChetlurWVCTCS14} is used for spatial attention and during prefill for temporal attention, while custom Triton \citep{10.1145/3315508.3329973} kernels are used during the decode stage.

\paragraph{Real-time rollout.} We aim to serve at $20$\, fps on a single Nvidia B200 GPU for the $5$B model with a latent window size of 20. Combining all the optimizations presented above, one full step, from the world-model update through decoding, takes roughly $70$\,ms end to end and produces two video frames (about $35$\,ms per frame), comfortably inside the $50$\,ms budget of a $20$\,fps stream. Running faster than real time leaves headroom to absorb scheduling jitter and to pace delivery smoothly (below).

 \begin{figure}[h!]
  \centering
  \includegraphics[width=\linewidth]{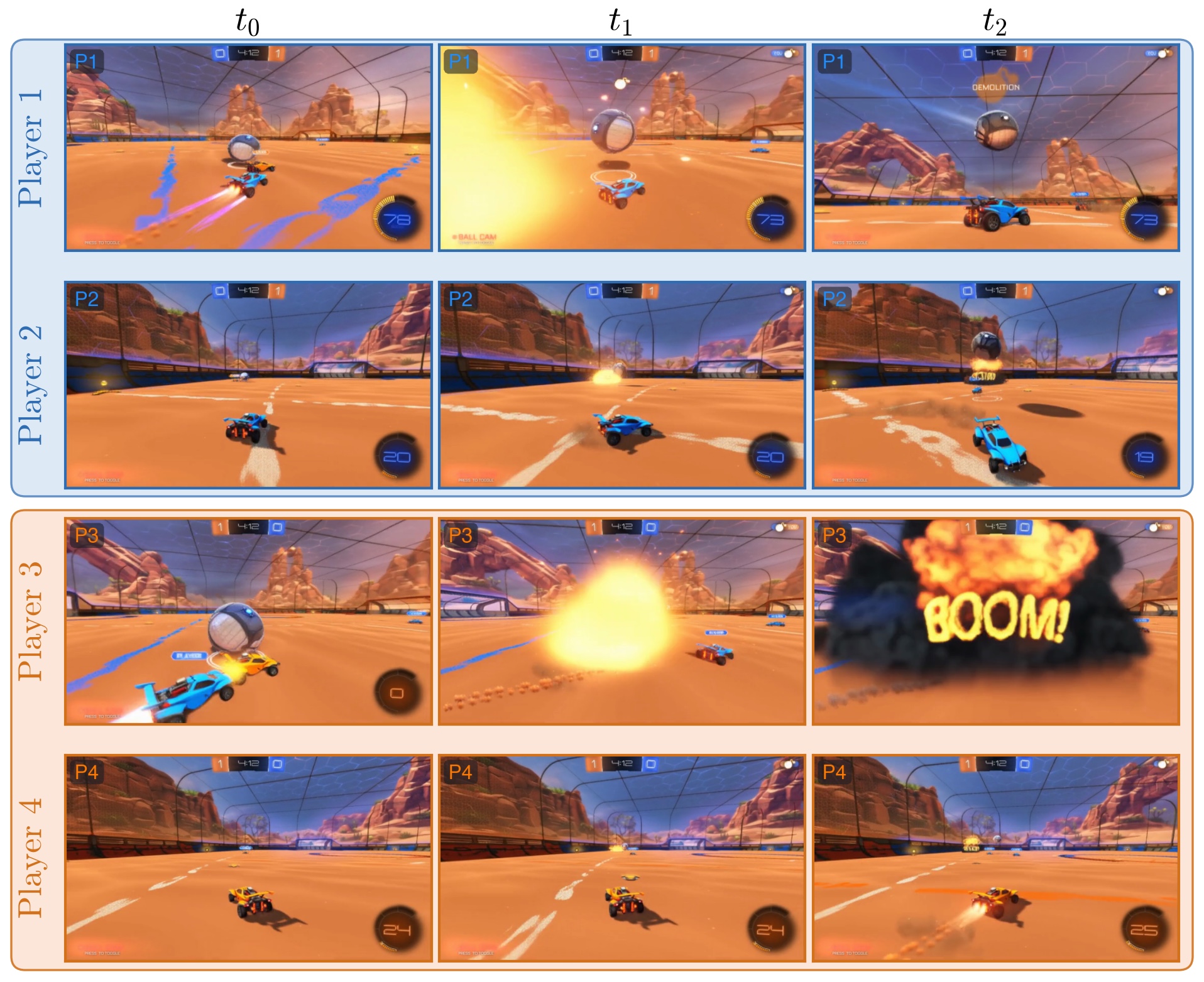}
  \caption{\textbf{World model imagination of a demolition event.} Rows
  are the four players' viewpoints (Players~1--4) and columns show three
  timesteps ($t_0 \to t_1 \to t_2$) of an imagined rollout in the desert arena.
  From the initial state at $t_0$ and the players' action streams, the model
  imagines a fast, sub-second collision: a blue car boosts into an opponent and
  demolishes it. The event stays mutually coherent across viewpoints. At $t_2$,
  Player~1 renders the ``DEMOLITION'' callout for its successful hit, while
  Player~3 (the demolished car) sees the ``BOOM!'' explosion fill its own view,
  and Player~2 renders the very same blast as a distant burst of smoke on the
  pitch. The in-game clock ($4{:}12$) is consistent across all players and
  barely advances, reflecting how brief the event is.}
  \label{fig:rollout_destroy}
\end{figure}

\begin{figure}[h!]
  \centering
  \includegraphics[width=\linewidth]{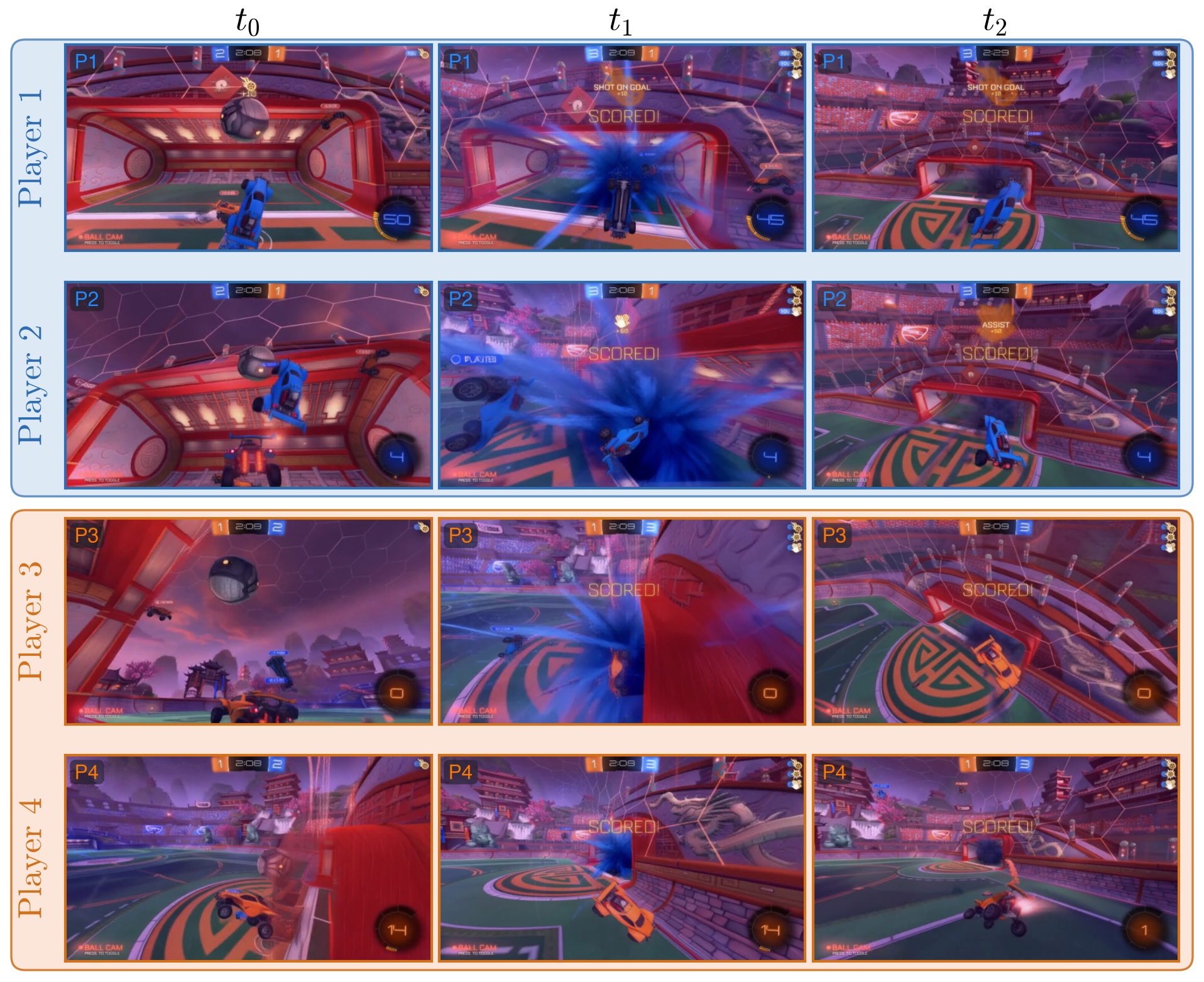}
  \caption{\textbf{World model imagination of a goal.} Rows are the four
  players' viewpoints (Players~1--4) and columns show three timesteps
  ($t_0 \to t_1 \to t_2$) of an imagined rollout in the temple arena. From the
  initial state at $t_0$ and the players' action streams, the model imagines the
  ball being driven into the net for a goal. The event stays mutually coherent
  across viewpoints: at $t_1$ the ``SCORED!'' callout and the blue goal-explosion
  appear simultaneously in all four players' views, each rendering the same blast
  from its own camera, and by $t_2$ every player sees the aftermath on the pitch.
  The rollout is also temporally consistent: the scoreline updates from
  $2$--$1$ to $3$--$1$ for the blue team as the goal registers.}
  \label{fig:goal}
\end{figure}

\section{Experiments}
\label{sec:experiments}

Our experiments cover three design axes. The first is the prediction space. We predict in
a learned latent; \cref{sec:exp-latent} tests this against pixels, then studies each part
of the codec that produces the latent: the feature extractor, the bottleneck, and the decoder. The
second is the world model's training objective; \cref{sec:exp-objective}
compares diffusion forcing \citep{chen2024diffusion} against teacher forcing for
long-rollout stability, and \cref{sec:exp-controllability} checks whether the trained model
actually executes the actions it is conditioned on. The third is conditioning on four simultaneous players
(\cref{sec:exp-multiplayer}). We then measure how performance scales with model and data
size (\cref{sec:exp-scaling}), and close with the properties that emerge in the trained
model (\cref{sec:exp-emergent}) and the failure modes that remain (\cref{sec:exp-failure}).

\subsection{The flagship model}
\label{sec:flagship}

The model behind our live demo is a $5$-billion-parameter multiplayer world model, the largest
configuration in this paper and the culmination of the design choices the rest of this section
validates: a $600$M-parameter video representation codec built on a DINOv3-L feature extractor with a trained bottleneck and decoder (\cref{sec:latent}), flow matching with diffusion
forcing (\cref{sec:objective}), and tiled four-player conditioning (\cref{sec:multiplayer}). It is
warm-started from a single-player model and then trained on multiplayer data over a dataset comprising
${\sim}10{,}000$ hours of gameplay; its full configuration is given in \cref{tab:app-demo-hparams}.

It simulates two-versus-two Rocket League interactively and in real time, rendering all four
players' views at $20$ frames per second on a single Nvidia B200 GPU, and stays controllable under four
simultaneous action streams, including under live human play. Its rollouts stay coherent far beyond
the training window, running without collapse for far longer than they are trained on
(\cref{sec:exp-emergent}). Most importantly, it reproduces the game's dynamics rather than only its
appearance: cars are knocked off course by contact and bump one another, demolitions are rendered
consistently from every camera (\cref{fig:rollout_destroy}), and a shot into the net sets off the
goal explosion and the on-screen \textsc{scored!} callout across all four views (\cref{fig:goal}). A
game-state probe further confirms that the rolled-out car and ball positions track the true physics
(\cref{fig:rollout2}).

\subsection{Evaluation metrics}
\label{sec:metrics}

Because the world model generates seconds of video whose exact future is unknowable, we judge a
rollout by its realism rather than by an exact match to the reference. We evaluate the codec and the
world model along five complementary axes: distributional distances that compare generated or
reconstructed video against real play, tracked against the rollout horizon to reveal drift;
pixel-level and perceptual distances that measure frame-by-frame fidelity; a probe of physical
state; a controllability metric that tests whether the model executes the actions it is given; and
human evaluation. Distributional distances compare statistics over a set of clips and capture global
realism, while pixel-level metrics are computed per frame and capture local fidelity. We prefix each
distributional metric with \emph{r} or \emph{g} to indicate whether it is applied to codec
\emph{reconstructions} (rFID, rFVD, rFDD) or to world-model \emph{generations} (gFID, gFVD, gFDD).

\paragraph{Distributional distances.}
The Fr\'echet Inception Distance (FID)~\citep{heusel2017gans} measures the distance between
the feature distributions of real and generated frames, extracted by an Inception-V3 network
trained on ImageNet. The Fr\'echet Video Distance (FVD)~\citep{unterthiner2018towards}
extends this to short video clips using a 3D feature extractor, making it sensitive to
temporal artefacts that per-frame FID misses. Because both backbones are trained on natural
images, they may transfer imperfectly to Rocket League's synthetic domain, so we complement them with the
Fr\'echet DINO Distance (FDD), a Fr\'echet distance computed in the feature space of the
DINOv3-B encoder~\citep{simeoni2025dinov3}, which better supports out-of-domain generalization and has richer features. All three distances are lower for better
quality. For reconstruction we apply them to $40$-frame codec outputs ($2$\,s windows) against real clips; for
generation we apply them to world-model rollouts conditioned on ground-truth actions, scored
in $1$\,s windows. Unless stated otherwise, we sample each world model with $10$ flow-matching
steps and re-noise the past context to level $0.2$ (its effect on drift is studied
in \cref{fig:exp-noise}), and generation distances are
reported at the $4$\,s horizon of the rollout.

\paragraph{Pixel and perceptual distances.}
Reconstruction quality is measured per frame. PSNR and SSIM~\citep{wang2004image} compare
pixel-level fidelity and structural similarity against the ground truth. LPIPS~\citep{zhang2018unreasonable}
measures perceptual distance in AlexNet feature space, and P-DINO does the same in DINOv3-B
feature space. These metrics apply only to the codec,
not to the world model, whose exact future is unknowable. PSNR and SSIM are higher for
better reconstruction; LPIPS and P-DINO are lower.

\paragraph{Game-state probing.}
To test whether generated rollouts respect the physics and rules of the game, we train a
probe that reads out interpretable game-state quantities from the world model's internal
representations. The probe hooks the last layer of world-model activations, spatially pools
them per player, concatenates the four player vectors, and passes the result through a
three-layer MLP with hidden dimension $1024$. The prediction layer maps to the game state:
position, quaternion, and linear velocity for each of the four players and the ball, $50$
dimensions in total. We train the probe with an L2 loss against the ground-truth states, reading the
world model's activations as it processes the real, encoded latent sequence rather than its own
generated rollout. At evaluation time, we instead apply the probe to a generated rollout: we encode
a video, roll out the world model conditioned on the ground-truth actions, run the probe on the
rollout activations, and compare the predicted states against the corresponding ground-truth
physical states. Learning the readout on real latents and testing it on the model's own rollout
isolates whether the generated trajectory preserves the same physical information. This metric targets correctness of
dynamics rather than appearance.

\paragraph{Action Recoverability Ratio.} Distributional and pixel metrics judge how a rollout looks, not
whether it obeys the actions it is given. To measure controllability, we test whether each commanded
action is recoverable from the generated video. We train an action probe, a frozen DINOv3-B encoder
\citep{simeoni2025dinov3} with a small attention-pooling head, to detect which of the nine controls
are pressed within an $8$-frame ($0.4$\,s) window, supervised by the logged key presses with a
per-action binary cross-entropy. Only the head is trained, so the probe reads domain-general features
and transfers across real, reconstructed, and generated frames. We score its detections with
\emph{average precision} (AP), the area under a per-action precision--recall curve; on real held-out
video the probe reaches $0.84$ mean AP (mAP, averaged over the nine actions), which makes it a
reliable instrument for this calibration (\cref{app:action-probe}). For a rollout conditioned on the
real actions, we slide the probe over both the generated video and the codec reconstruction of the
same clip and compute each action's AP against the commanded actions, $\mathrm{AP}_{\mathrm{gen}}$ and
$\mathrm{AP}_{\mathrm{recon}}$. The action recoverability ratio for action $a$ is
\[
\mathrm{ARR}(a) = \frac{\mathrm{AP}_{\mathrm{gen}}(a)}{\mathrm{AP}_{\mathrm{recon}}(a)},
\]
averaged over the nine actions and reported with bootstrap confidence intervals over clips. The
reconstruction is the achievable ceiling, since a perfect world model given the real actions would
reproduce the real clip's latents; dividing by it cancels the probe's own imperfection and the codec's
visual domain, leaving only how faithfully the world model renders the action. An $\mathrm{ARR}$ of $1$
means actions are as recoverable from the generation as from a faithful reconstruction, below $1$ means
the model under-renders them.

\paragraph{Human evaluation.} Automatic metrics are proxies, so we corroborate them with human
preference studies scored by a Bayesian Elo. Two protocols mirror our two main axes. For
\emph{quality}, raters compare a pair of models rollout-for-rollout from the same initial frame and
action sequence and pick the video that better matches the ground-truth reference, both in appearance
and in the end state of the play. For \emph{action adherence}, we build a hybrid clip by applying one
reference's actions to another's initial frame, holding appearance and starting point fixed while the
actions differ; raters then judge which clip follows the reference actions more faithfully, the human
analog of ARR (Action Recoverability Ratio, defined above). Sample and rater counts vary by comparison. We report the full studies in
\cref{sec:human-eval}; the action-adherence study validates ARR in \cref{sec:exp-controllability}.

\subsection{Designing the codec's latent space}
\label{sec:exp-latent}

The codec learns to reconstruct video, but the world model must generate in its latent space,
and good reconstruction does not guarantee an easy-to-generate latent
\citep{yao2025reconstruction, xu2026ifid}. We therefore judge each codec by the world
model trained on it, and treat reconstruction quality as secondary evidence.

All codec ablations run in a single-player setting, on one player's $288\times512$,
$20$\,fps view; the codec operates per view, so one view is enough to compare codecs. The baseline is a representation codec: a frozen
DINOv3-L feature extractor \citep{simeoni2025dinov3} extracts per-frame features, a learned linear
bottleneck compresses them from $1024$ to $32$ channels and downsamples them to one latent
cell per $32\times32$ pixels at $10$\,Hz, half the input rate. A spatio-temporal
ViT-XL decoder, causal in time, reconstructs the full-rate video. The feature extractor and bottleneck
shape the latent; the decoder governs only how faithfully it is rendered back to pixels.
The ablation runs use a shorter schedule than the flagship: each codec is taken at its $125$k-step
checkpoint on $40$-frame clips (\cref{tab:codec-hparams} gives the baseline codec's full
configuration), and a $1$B-parameter world model then
trains for $100$k steps at global batch $16$ on $80$-frame clips on that codec. Both stages
process $1280$ frames per step, so every ablated codec sees $160$M frames and every world
model $128$M frames.

For each codec we report reconstruction quality, with pixel and perceptual metrics (PSNR,
SSIM, LPIPS, and P-DINO, a perceptual distance in DINOv3 feature space) and reconstruction
Fr\'echet distances (rFID, rFVD, rFDD). For the world model trained on that codec, we roll
out $12$\,s conditioned on the ground-truth actions and measure distributional distances to
real play (gFID, gFVD, gFDD) in $1$\,s windows, and summarize each rollout by its distance
at the $4$\,s horizon: $4$\,s is the model's training window, long enough to exercise
multi-step generation yet short enough to keep models discriminable. We score
reconstruction on $40$-frame clips and every distance over a
frozen set of $2048$ clips (\cref{sec:metrics}). In every table below, the best value in
each column is bold and the second best underlined, generation columns are taken at the
$4$\,s horizon, and dashes mark runs not yet scored.

\begin{figure}[htbp]
  \centering
  \begin{subfigure}[b]{0.58\linewidth}
  \centering
  \begin{tikzpicture}
  \begin{axis}[rsplot, width=\linewidth, height=4.2cm, ymin=0, ymax=32,
    xmode=log, log ticks with fixed point, xmin=1, xmax=300,
    xtick={1,4,12,30,60,120,300}, x tick label style={font=\scriptsize},
    xlabel={rollout time (s)}, ylabel={gFID\,$\downarrow$},
    legend style={at={(0.5,-0.42)}, anchor=north, legend columns=2, column sep=6pt, font=\scriptsize, draw=none}]
    \addplot[gray!55, line width=0.8pt, forget plot] coordinates {(1,6.8)(2,8.7)(3,10.2)(4,10.9)(5,11.4)(6,11.4)(7,11.7)(8,12.4)(9,12.6)(10,12.5)(11,12.7)(12,12.6)(30,14.8)(60,15.2)(120,15.7)(300,15.7)};
    \addplot[gray!55, line width=0.8pt, forget plot] coordinates {(1,8.1)(2,10.2)(3,11.2)(4,12.3)(5,12.8)(6,12.5)(7,12.8)(8,13.1)(9,14.0)(10,14.3)(11,14.6)(12,14.3)(30,16.0)(60,16.9)(120,16.9)(300,17.2)};
    \addplot[gray!55, line width=0.8pt, forget plot] coordinates {(1,7.3)(2,9.3)(3,10.3)(4,11.4)(5,12.1)(6,11.9)(7,11.8)(8,12.4)(9,13.0)(10,13.0)(11,13.2)(12,13.1)(30,15.8)(60,16.3)(120,16.7)(300,16.5)};
    \addplot[plotblue, line width=1.4pt, forget plot] coordinates {(1,6.6)(2,8.5)(3,9.6)(4,10.7)(5,11.4)(6,11.2)(7,11.5)(8,11.8)(9,12.3)(10,12.4)(11,12.5)(12,12.3)(30,15.0)(60,15.5)(120,15.8)(300,15.9)};
    \addplot[plotgreen, line width=1.4pt, forget plot] coordinates {(1,10.5)(2,13.3)(3,14.6)(4,15.5)(5,15.9)(6,15.8)(7,15.8)(8,15.9)(9,16.2)(10,16.7)(11,16.8)(12,16.9)(30,21.5)(60,22.1)(120,21.5)(300,22.3)};
    \addplot[plotorange, line width=1.4pt, forget plot] coordinates {(1,13.7)(2,17.9)(3,19.9)(4,22.5)(5,23.2)(6,23.0)(7,23.5)(8,24.5)(9,25.6)(10,26.0)(11,25.8)(12,26.0)(30,27.6)(60,28.3)(120,28.8)(300,29.3)};
    \addlegendimage{plotblue, line width=1.4pt}\addlegendentry{ours}
    \addlegendimage{gray!55, line width=0.8pt}\addlegendentry{other pretrained enc.}
    \addlegendimage{plotgreen, line width=1.4pt}\addlegendentry{distill}
    \addlegendimage{plotorange, line width=1.4pt}\addlegendentry{from scratch}
    \draw[dashed, gray!55!black, semithick] (axis cs:4,0) -- (axis cs:4,32);
    \node[anchor=north west, font=\scriptsize, fill=white, inner sep=1pt, text=gray!50!black]
      at (axis cs:4,31.5) {Train.\ window};
  \end{axis}
  \end{tikzpicture}
  \caption{gFID over a five-minute rollout (log time) for each codec.}
  \label{fig:rollout-drift}
  \end{subfigure}\hfill
  \begin{subfigure}[b]{0.40\linewidth}
  \centering
  \begin{tikzpicture}
  \begin{axis}[rsplot, ybar, bar width=12pt, width=\linewidth, height=4.2cm,
    ymin=0, ymax=1.85, ytick={0,0.5,1,1.5},
    ylabel={gFID drift rel.\ to DINOv3-L\,$\downarrow$},
    symbolic x coords={DINOv3-L, Distill, From scratch},
    xtick={DINOv3-L, Distill, From scratch},
    x tick label style={rotate=25, anchor=north east, font=\scriptsize},
    enlarge x limits=0.28,
    extra y ticks={1}, extra y tick labels={},
    extra y tick style={grid=major, major grid style={dashed, gray!55!black, line width=0.5pt}}]
    \addplot[draw=plotblue, fill=plotblue!80, bar shift=0pt] coordinates {(DINOv3-L,1.00)};
    \addplot[draw=plotgreen, fill=plotgreen!80, bar shift=0pt] coordinates {(Distill,1.27)};
    \addplot[draw=plotorange, fill=plotorange!85, bar shift=0pt] coordinates {(From scratch,1.67)};
  \end{axis}
  \end{tikzpicture}
  \caption{gFID drift ($0\!\to\!300$\,s) relative to DINOv3-L; the dashed line marks the baseline.}
  \label{fig:drift-rate}
  \end{subfigure}
  \caption{\textbf{Rollout drift.} Each variant uses the same codec architecture and
  training, differing only in the feature extractor: ours uses a frozen DINOv3-L, ``distill'' trains
  the feature extractor from scratch with DINO distillation, and ``from scratch'' trains it from
  scratch without distillation. (\subref{fig:rollout-drift})~gFID over a five-minute rollout: the codecs with a
  pretrained feature extractor stay low and nearly flat on long horizons, the distilled codec sits
  higher, and the from-scratch codec is highest. (\subref{fig:drift-rate})~gFID drift over the
  five-minute rollout relative to DINOv3-L: the distilled and from-scratch codecs drift $1.3$ and
  $1.7\times$ as much as the pretrained baseline.}
  \label{fig:drift}
\end{figure}
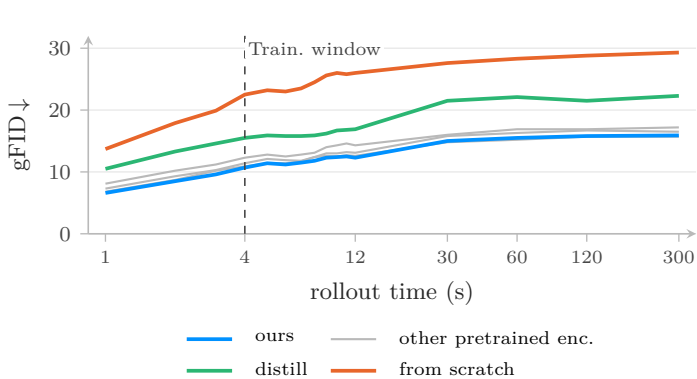
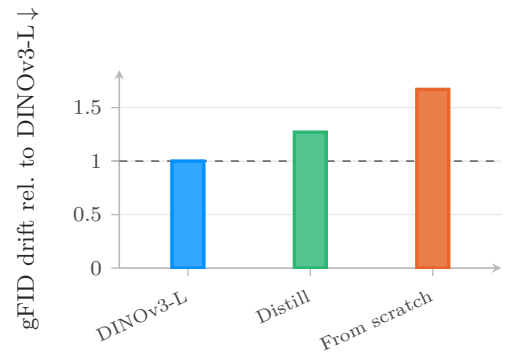

\begin{takeaway}{Designing the latent space, in brief.} We judge every codec by the world model trained on it and select for an easy-to-generate latent, treating reconstruction quality as secondary.

\smallskip
\emph{Presented in the main text:}
\begin{itemize}[nosep, leftmargin=1.3em, topsep=2pt]
\item \textbf{A latent is essential.} Pixel-space world models trail ours by an order of magnitude on generation and controllability (\cref{tab:exp-pixels}).
\item \textbf{The feature extractor must be pretrained.} A frozen DINOv3 extractor makes the latent easier to generate and far more stable over long rollouts, whereas a from-scratch extractor reconstructs more sharply yet drifts (\cref{tab:exp-enc-adapt}, \cref{fig:drift}).
\item \textbf{The bottleneck is best learned, but need not be.} A learned linear bottleneck adds a small, consistent margin; a random or PCA projection of the pretrained features is already predictable (\cref{tab:exp-bottleneck}).
\item \textbf{Temporal downsampling is free.} A $10$\,Hz latent leaves generation unchanged while halving the world model's sequence length, buying the real-time $20$\,fps rollout at no cost in quality (\cref{tab:exp-temporal}).
\item \textbf{Perceptual losses are essential and sufficient.} LPIPS and P-DINO are complementary and removing both collapses generation; together they reach enough fidelity that no adversarial (GAN) loss is needed (\cref{tab:exp-perceptual}).
\item \textbf{Decoder quality saturates with scale.} Reconstruction improves as the decoder grows but with diminishing returns beyond the Large ($0.3$B) decoder (\cref{tab:exp-dec-size}).
\end{itemize}

\smallskip
\emph{Detailed in \cref{sec:codec-ablations}:}
\begin{itemize}[nosep, leftmargin=1.3em, topsep=2pt]
\item \textbf{Any reasonable pretrained extractor works.} Smaller DINOv3 variants and EUPE-B trail DINOv3-L only slightly, and all resist drift (\cref{tab:exp-enc-family}).
\item \textbf{Read several feature-extractor layers, not just the last.} Averaging a spread of the feature extractor's intermediate blocks keeps spatial detail that the deepest block discards (\cref{tab:exp-enc-layers}).
\item \textbf{Compression can go in the codec or the world model.} The $2\times$ reduction works in either, so we do all of it in the codec (\cref{tab:exp-where}).
\item \textbf{Adaptive loss balancing helps.} Gradient-matched weights beat fixed weights on nearly all reconstruction and generation metrics (\cref{tab:exp-balance}).
\item \textbf{Upsample before decoding, not after.} Expanding the compact latent to the feature-extractor grid at the decoder's input, so the decoder runs on more tokens, far outperforms upsampling only at the output (\cref{tab:exp-upconv}).
\end{itemize}
\end{takeaway}

\paragraph{Is pixel-space world modeling enough?} Before designing the latent, we ask whether a codec is needed at all. Pixel-space generation has drawn renewed interest, with transformers that flow-match directly on pixels \citep{li2025jit, ma2026pixelgen}. We test two pixel-space world models against our latent-space baseline. The first changes only the world model's input and output, replacing the codec latent with patchified pixels while keeping the architecture, flow-matching velocity-based objective, and training unchanged; the second also adopts the JiT recipe \citep{li2025jit}, which feeds the pixel patches through a linear bottleneck before the transformer and predicts the clean signal instead of the velocity. Both pixel-space models train for $225$k steps, matching the total budget of the codec-plus-world-model pipeline ($125$k\,$+$\,$100$k). The pixel-space models have no codec and thus no reconstruction to score, so we compare on generation quality and controllability (\cref{tab:exp-pixels}).

\begin{table}[htbp]
\centering\scriptsize
\setlength{\tabcolsep}{6pt}\renewcommand{\arraystretch}{1.15}
\caption{\textbf{Latent vs pixel space.} Our latent world model against two pixel-space
world models. All runs match the total training budget ($225$k steps); the latent splits this
into codec ($125$k) plus world model ($100$k). The pixel models have no autoencoder, so their
controllability (ARR) is calibrated against real frames rather than a reconstruction.}
\label{tab:exp-pixels}
\begin{tabular}{l c cccc}
\toprule
& Steps & gFID\,$\downarrow$ & gFVD\,$\downarrow$ & gFDD\,$\downarrow$ & ARR\,$\uparrow$ \\
\midrule
\textbf{Latent (ours)} & $125$k\,$+$\,$100$k & \textbf{10.7} & \textbf{163.1} & \textbf{0.55} & \textbf{0.91} \\
Pixels, plain      & $225$k & 104.9 & 1456.3 & \underline{16.19} & \underline{0.61} \\
Pixels, JiT recipe & $225$k & \underline{81.0} & \underline{961.2} & 17.05 & 0.49 \\
\bottomrule
\end{tabular}
\end{table}

\begin{takeaway}{Is pixel-space world modeling enough?} No. Both pixel models trail the latent baseline by an order of magnitude on generation quality, even with the JiT bottleneck and a matched total training budget, and their rollouts are markedly less controllable, with commanded actions far harder to recover; the pixel-space rollout also drifts into warped, unstructured texture within a second (\cref{fig:pixeldrift}). A learned latent is essential.\end{takeaway}

\paragraph{Must the feature extractor be pretrained?} The baseline feature extractor is a frozen, pretrained DINOv3-L. To test whether the pretrained features are needed, we replace it with a feature extractor of the same architecture trained from scratch (randomly initialized, learned end-to-end with the rest of the codec), in two forms: on its own, and with DINO distillation, where a learned linear map $W$ projects the bottleneck latent $z$ to match a frozen DINOv3-L teacher's features $f$, minimizing the squared error $\lVert Wz - f\rVert^2$, so the latent inherits DINO's structure even though the feature extractor is learned (\cref{tab:exp-enc-adapt}). Beyond generation quality at the $4$\,s horizon, we also roll the world model out to five minutes to measure drift (\cref{fig:drift}).

\begin{table}[htbp]
\centering\scriptsize
\setlength{\tabcolsep}{3pt}\renewcommand{\arraystretch}{1.2}
\caption{\textbf{Feature-extractor adaptation.} Frozen pretrained DINOv3-L against a feature extractor trained from scratch, with and without DINO distillation.}
\label{tab:exp-enc-adapt}
\begin{tabular}{l ccccccc | ccc}
\toprule
& \multicolumn{7}{c}{Codec (reconstruction)} & \multicolumn{3}{c}{World model (generation)} \\
\cmidrule(lr){2-8}\cmidrule(lr){9-11}
Feature extractor & PSNR\,$\uparrow$ & SSIM\,$\uparrow$ & LPIPS\,$\downarrow$ & P-DINO\,$\downarrow$ & rFID\,$\downarrow$ & rFVD\,$\downarrow$ & rFDD\,$\downarrow$ & gFID\,$\downarrow$ & gFVD\,$\downarrow$ & gFDD\,$\downarrow$ \\
\midrule
\textbf{Frozen DINOv3-L (ours)} & 29.7 & 0.891 & \textbf{0.051} & \textbf{0.021} & \textbf{5.4} & 38.6 & \textbf{0.17} & \textbf{10.7} & \textbf{163.1} & \textbf{0.55} \\
From scratch & \textbf{32.2} & \textbf{0.923} & 0.068 & 0.046 & 13.1 & \textbf{23.2} & 1.39 & 22.5 & 375.0 & 1.93 \\
From scratch $+$ DINO distillation & \underline{31.4} & \underline{0.916} & \underline{0.056} & \underline{0.037} & \underline{8.8} & \underline{26.2} & \underline{0.98} & \underline{15.7} & \underline{271.5} & \underline{1.35} \\
\bottomrule
\end{tabular}
\end{table}

We show qualitative examples of drift elsewhere: the pixel-space rollout in
\cref{fig:pixeldrift} and the out-of-distribution excursion in \cref{fig:recovery}.

\begin{takeaway}{Must the feature extractor be pretrained?} Yes, and pretraining helps generation twice over.
\begin{itemize}[nosep, leftmargin=1.3em, topsep=2pt]
\item \textbf{Quality:} a from-scratch feature extractor reconstructs more sharply, yet its latent is far harder to generate, inflating every generation distance at the $4$\,s horizon.
\item \textbf{Stability:} over a sustained rollout the from-scratch and distilled codecs drift substantially more than the pretrained baseline (\cref{fig:drift}); ours stays visibly stable far longer.
\end{itemize}
We attribute this to the smoothness of the pretrained features: nearby states map to nearby latents, so a wrong prediction stays valid and is absorbed.\end{takeaway}

\begin{figure}[htbp]
  \centering
  \begin{tikzpicture}
  \begin{axis}[rsplot, width=0.6\linewidth, height=0.4\linewidth, ymin=0, ymax=95,
    ytick={0,20,40,60,80},
    xmin=22, xmax=103, xtick={25,50,75,100},
    xlabel={training step (k)}, ylabel={gFID at $4$\,s\,$\downarrow$},
    legend style={at={(0.97,0.97)}, anchor=north east}]
    \addplot[plotblue, mark=*] coordinates {(25,30.2)(50,12.2)(75,10.8)(100,10.7)};
    \addlegendentry{DINOv3-L}
    \addplot[plotorange, mark=*] coordinates {(25,88.8)(50,31.0)(75,23.6)(100,22.5)};
    \addlegendentry{from scratch}
    \addplot[plotgreen, mark=*] coordinates {(25,43.5)(50,19.8)(75,17.3)(100,15.7)};
    \addlegendentry{from scratch $+$ distillation}
  \end{axis}
  \end{tikzpicture}
  \caption{\textbf{A pretrained feature extractor accelerates learning.} gFID at the $4$\,s horizon against
  training step, from $25$k steps. The latent on a pretrained
  DINOv3-L feature extractor reaches low gFID far sooner than a feature extractor trained from scratch.
  Distilling DINO into a from-scratch feature extractor recovers much of the gap.}
  \label{fig:gfid-train}
\end{figure}
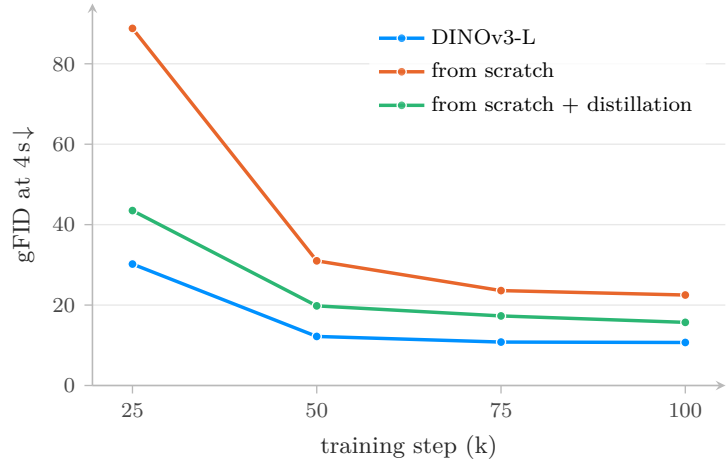

\paragraph{Does the bottleneck need to be learned?} 
To compress the aggregated DINOv3-L features to the $32$-channel latent (one cell per $32\times32$ pixels, $10$\,Hz; \cref{sec:latent}), our baseline uses a learned linear $2\times2\times2$ strided convolution.
% The baseline bottleneck is a learned linear $2\times2\times2$ strided convolution that compresses the aggregated DINOv3-L features to the $32$-channel latent (one cell per $32\times32$ pixels, $10$\,Hz; \cref{sec:latent}).
We replace it with progressively simpler maps, holding the feature extractor and decoder fixed: a random frozen projection, a fixed PCA projection with pooling, and pooling alone with no channel reduction (keeping all $1024$ channels). The pooling-only latent is high-dimensional, so for that variant we also shift the world model's flow-matching noise schedule toward higher noise, following the RAE recipe \citep{zheng2025rae} (\cref{tab:exp-bottleneck}).

\begin{table}[htbp]
\centering\scriptsize
\setlength{\tabcolsep}{3pt}\renewcommand{\arraystretch}{1.2}
\caption{\textbf{Bottleneck.} Replacing the learned channel compression with simpler, parameter-free maps. The shifted-noise row reuses the pooling-only codec, so only its world model differs.}
\label{tab:exp-bottleneck}
\begin{tabular}{l ccccccc | ccc}
\toprule
& \multicolumn{7}{c}{Codec (reconstruction)} & \multicolumn{3}{c}{World model (generation)} \\
\cmidrule(lr){2-8}\cmidrule(lr){9-11}
Bottleneck & PSNR\,$\uparrow$ & SSIM\,$\uparrow$ & LPIPS\,$\downarrow$ & P-DINO\,$\downarrow$ & rFID\,$\downarrow$ & rFVD\,$\downarrow$ & rFDD\,$\downarrow$ & gFID\,$\downarrow$ & gFVD\,$\downarrow$ & gFDD\,$\downarrow$ \\
\midrule
\textbf{Learned convolution (ours)} & \textbf{29.7} & \textbf{0.891} & \textbf{0.051} & \textbf{0.021} & \textbf{5.4} & \textbf{38.6} & \textbf{0.17} & \textbf{10.7} & \textbf{163.1} & \textbf{0.55} \\
Random frozen projection & 28.3 & 0.869 & 0.067 & \underline{0.025} & 6.6 & 65.9 & 0.20 & 11.9 & 219.9 & 0.67 \\
PCA $+$ pooling & 28.4 & 0.869 & 0.068 & \underline{0.025} & 6.8 & 67.0 & 0.21 & \underline{11.7} & \underline{182.4} & \textbf{0.55} \\
Pooling only (no channel reduction) & \underline{29.2} & \underline{0.882} & \underline{0.055} & \textbf{0.021} & \underline{6.0} & \underline{46.8} & \underline{0.18} & 13.1 & 243.4 & \underline{0.60} \\
\quad $+$ shifted noise schedule & -- & -- & -- & -- & -- & -- & -- & 12.7 & 238.1 & 0.62 \\
\bottomrule
\end{tabular}
\end{table}

\begin{takeaway}{Does the bottleneck need to be learned?} Not strictly, but learning helps. A learned linear bottleneck adds a small, consistent margin over a random or PCA projection of the DINO features, yet even those parameter-free maps already give a predictable latent. The pretrained features do most of the work.\end{takeaway}

\paragraph{Does temporal downsampling hurt quality?} The baseline halves the latent's frame rate in the codec ($20$\,fps video to a $10$\,Hz latent), which shortens the world model's sequence but discards temporal detail. We compare against a codec that keeps the full $20$\,Hz latent (\cref{tab:exp-temporal}); the two are matched in wall-clock context, so the full-rate variant carries twice the world model's latent sequence length.

\begin{table}[htbp]
\centering\scriptsize
\setlength{\tabcolsep}{3pt}\renewcommand{\arraystretch}{1.2}
\caption{\textbf{Temporal downsampling.} A $10$\,Hz latent (the codec halves the frame rate) against a $20$\,Hz latent (no temporal downsampling).}
\label{tab:exp-temporal}
\begin{tabular}{l ccccccc | ccc}
\toprule
& \multicolumn{7}{c}{Codec (reconstruction)} & \multicolumn{3}{c}{World model (generation)} \\
\cmidrule(lr){2-8}\cmidrule(lr){9-11}
Latent rate & PSNR\,$\uparrow$ & SSIM\,$\uparrow$ & LPIPS\,$\downarrow$ & P-DINO\,$\downarrow$ & rFID\,$\downarrow$ & rFVD\,$\downarrow$ & rFDD\,$\downarrow$ & gFID\,$\downarrow$ & gFVD\,$\downarrow$ & gFDD\,$\downarrow$ \\
\midrule
\textbf{10\,Hz (ours)} & 29.7 & 0.891 & 0.051 & 0.021 & 5.4 & 38.6 & 0.17 & 10.7 & \textbf{163.1} & \textbf{0.55} \\
20\,Hz (no temporal ds.) & \textbf{30.6} & \textbf{0.909} & \textbf{0.039} & \textbf{0.016} & \textbf{4.3} & \textbf{23.0} & \textbf{0.13} & \textbf{10.5} & 166.3 & 0.65 \\
\bottomrule
\end{tabular}
\end{table}

\begin{takeaway}{Does latent temporal downsampling hurt quality?} No, it is essentially free. Halving the latent frame rate to $10$\,Hz only softens reconstruction and leaves generation unchanged, while halving the world model's sequence length, which buys the real-time $20$\,fps rollout at no cost in generation quality.\end{takeaway}

\paragraph{What perceptual losses does the codec need?} Beyond the L1 reconstruction loss, the codec uses two perceptual terms, an LPIPS term \citep{zhang2018unreasonable} and a P-DINO term (the perceptual DINO distance used as a metric above). We toggle which are active: both, LPIPS alone, P-DINO alone, or neither (\cref{tab:exp-perceptual}).

\begin{table}[htbp]
\centering\scriptsize
\setlength{\tabcolsep}{3pt}\renewcommand{\arraystretch}{1.2}
\caption{\textbf{Perceptual terms.} Toggling the LPIPS and P-DINO perceptual losses.}
\label{tab:exp-perceptual}
\begin{tabular}{l ccccccc | ccc}
\toprule
& \multicolumn{7}{c}{Codec (reconstruction)} & \multicolumn{3}{c}{World model (generation)} \\
\cmidrule(lr){2-8}\cmidrule(lr){9-11}
Perceptual loss & PSNR\,$\uparrow$ & SSIM\,$\uparrow$ & LPIPS\,$\downarrow$ & P-DINO\,$\downarrow$ & rFID\,$\downarrow$ & rFVD\,$\downarrow$ & rFDD\,$\downarrow$ & gFID\,$\downarrow$ & gFVD\,$\downarrow$ & gFDD\,$\downarrow$ \\
\midrule
\textbf{LPIPS $+$ P-DINO (ours)} & 29.7 & 0.891 & \textbf{0.051} & \textbf{0.021} & \textbf{5.4} & \textbf{38.6} & \textbf{0.17} & \textbf{10.7} & \textbf{163.1} & \textbf{0.55} \\
LPIPS only & \underline{30.1} & \underline{0.899} & \underline{0.067} & 0.041 & \underline{10.6} & \underline{49.9} & 1.16 & \underline{14.4} & \underline{175.6} & 1.36 \\
P-DINO only & 29.8 & 0.890 & 0.080 & \underline{0.028} & 11.9 & 61.4 & \underline{0.43} & 16.6 & 196.6 & \underline{0.83} \\
No perceptual loss & \textbf{30.3} & \textbf{0.903} & 0.104 & 0.069 & 25.0 & 72.3 & 2.57 & 26.8 & 194.1 & 2.58 \\
\bottomrule
\end{tabular}
\end{table}

\begin{takeaway}{What perceptual losses does the codec need?} Both, and only these. LPIPS and P-DINO are complementary, LPIPS helping gFID and gFVD more and P-DINO helping gFDD more; removing either hurts generation and removing both collapses it. Together they reach enough fidelity that we add no adversarial (GAN) loss.\end{takeaway}

\paragraph{How large must the decoder be?} We scale the ViT decoder \citep{dosovitskiy2021image} across three sizes, Base ($12$ layers, width $768$; ${\approx}85$M parameters), Large ($24$ layers, width $1024$; ${\approx}0.3$B), and XL (ours; $28$ layers, width $1152$; ${\approx}0.45$B), holding the rest of the codec fixed (\cref{tab:exp-dec-size}).

\begin{table}[htbp]
\centering\scriptsize
\setlength{\tabcolsep}{3pt}\renewcommand{\arraystretch}{1.2}
\caption{\textbf{Decoder size.} Base, Large, and XL (ours) decoders, all upsampling before the ViT.}
\label{tab:exp-dec-size}
\begin{tabular}{l ccccccc | ccc}
\toprule
& \multicolumn{7}{c}{Codec (reconstruction)} & \multicolumn{3}{c}{World model (generation)} \\
\cmidrule(lr){2-8}\cmidrule(lr){9-11}
Decoder size & PSNR\,$\uparrow$ & SSIM\,$\uparrow$ & LPIPS\,$\downarrow$ & P-DINO\,$\downarrow$ & rFID\,$\downarrow$ & rFVD\,$\downarrow$ & rFDD\,$\downarrow$ & gFID\,$\downarrow$ & gFVD\,$\downarrow$ & gFDD\,$\downarrow$ \\
\midrule
\textbf{XL (ours)} & \textbf{29.7} & \textbf{0.891} & \textbf{0.051} & \textbf{0.021} & \textbf{5.4} & \textbf{38.6} & \textbf{0.17} & \underline{10.7} & \textbf{163.1} & \textbf{0.55} \\
Base & 27.6 & 0.842 & 0.082 & 0.029 & 8.2 & 89.8 & \underline{0.27} & 12.5 & 201.3 & \underline{0.66} \\
Large & \underline{29.3} & \underline{0.882} & \underline{0.055} & \underline{0.022} & \underline{5.6} & \underline{43.1} & \textbf{0.17} & \textbf{10.5} & \underline{169.9} & \textbf{0.55} \\
\bottomrule
\end{tabular}
\end{table}

\begin{takeaway}{How large must the decoder be?} Large is enough. Reconstruction improves with capacity but saturates with diminishing returns from the Large decoder onward, which already matches XL, while a Base decoder clearly degrades reconstruction.\end{takeaway}

\subsection{World model objectives}
\label{sec:exp-objective}

We train the world model with diffusion forcing (\cref{sec:objective}), and test whether
it improves on teacher forcing, a long-established sequence-modeling scheme that earlier
diffusion world models adopt. We study its effect on sample quality and on robustness
to drift over long rollouts, ask whether the context must be noised at inference, and whether
sampling can be accelerated by distilling the model into a few-step sampler. All three studies
use $1$B-parameter world models on the baseline codec, with the same batch size and schedule
as the codec ablations (\cref{sec:exp-latent}).

\paragraph{How do teacher and diffusion forcing compare?} Earlier
diffusion world models~\citep{alonso2024diffusion, valevski2025diffusion} are trained by
teacher forcing: at each step the model denoises the next frame from clean, ground-truth
context, so it is supervised on a single frame and never trains on the imperfect context it
must build on at rollout time. Diffusion forcing~\citep{chen2024diffusion}, introduced to
narrow this train/inference gap, instead noises every frame of the clip independently,
supervising all frames at once and exposing the model to noised context that stands in for the
imperfect predictions it conditions on at rollout. We
compare the two on the baseline codec, both on sample quality at the $4$\,s horizon
(\cref{tab:exp-objective}) and on drift across a five-minute rollout (\cref{fig:tf-drift}).

\begin{figure}[htbp]
\centering
\begin{minipage}[c]{0.46\linewidth}
  \centering\footnotesize
  \setlength{\tabcolsep}{4pt}\renewcommand{\arraystretch}{1.2}
  \captionof{table}{\textbf{Training objective.} Teacher forcing against diffusion
  forcing on the baseline codec; generation quality at the $4$\,s horizon.}
  \label{tab:exp-objective}
  \begin{tabular}{l ccc}
  \toprule
  Objective & gFID\,$\downarrow$ & gFVD\,$\downarrow$ & gFDD\,$\downarrow$ \\
  \midrule
  Teacher forcing                   & 32.5 & 944.1 & 2.21 \\
  \textbf{Diffusion forcing (ours)} & \textbf{10.7} & \textbf{163.1} & \textbf{0.55} \\
  \bottomrule
  \end{tabular}
\end{minipage}\hfill
\begin{minipage}[c]{0.51\linewidth}
  \centering
  \begin{tikzpicture}
  \begin{axis}[rsplot, width=\linewidth, height=0.66\linewidth,
    log ticks with fixed point, ymin=0, ymax=170, ytick={0,50,100,150},
    xmode=log, xmin=1, xmax=300, xtick={1,4,12,30,60,120,300}, x tick label style={font=\scriptsize},
    xlabel={rollout time (s)}, ylabel={gFID\,$\downarrow$}]
    \addplot[plotorange, line width=1.4pt, mark=*, mark size=1.1pt] coordinates {(1,10.4)(2,16.9)(3,28.2)(4,32.5)(5,79.8)(6,116.8)(7,135.6)(8,141.0)(9,140.1)(10,142.1)(11,143.7)(12,143.8)(30,143.3)(60,147.0)(120,149.1)(300,154.9)};
    \addplot[plotblue, line width=1.4pt, mark=*, mark size=1.1pt] coordinates {(1,6.6)(2,8.5)(3,9.6)(4,10.7)(5,11.4)(6,11.2)(7,11.5)(8,11.8)(9,12.3)(10,12.4)(11,12.5)(12,12.3)(30,15.0)(60,15.5)(120,15.8)(300,15.9)};
    \draw[dashed, gray!55!black, semithick] (axis cs:4,0) -- (axis cs:4,170);
  \end{axis}
  \end{tikzpicture}
  \par\smallskip
  \begin{tikzpicture}[font=\footnotesize]
    \draw[plotorange, line width=1.4pt] (0,0) -- (0.4,0);
    \node[anchor=west, inner sep=2pt] at (0.4,0) {Teacher forcing};
    \draw[plotblue, line width=1.4pt] (2.6,0) -- (3.0,0);
    \node[anchor=west, inner sep=2pt] at (3.0,0) {Diffusion forcing (ours)};
  \end{tikzpicture}
  \caption{\textbf{Objective and drift.} gFID across a $5$-minute rollout (log time) on the
  baseline codec. Past the $4$\,s training window (dashed), teacher forcing degrades sharply
  and stays about $10\times$ higher, while diffusion forcing stays low and flat.}
  \label{fig:tf-drift}
\end{minipage}
\end{figure}

\begin{takeaway}{How do teacher and diffusion forcing compare?} Diffusion forcing
wins clearly. It is markedly better across the generation metrics at the $4$\,s horizon, and over a
long rollout it limits drift far better: teacher forcing collapses to a much higher gFID while
diffusion forcing stays flat (\cref{fig:tf-drift}).\end{takeaway}

\begin{figure}[htbp]
  \centering
  \resizebox{\linewidth}{!}{\input{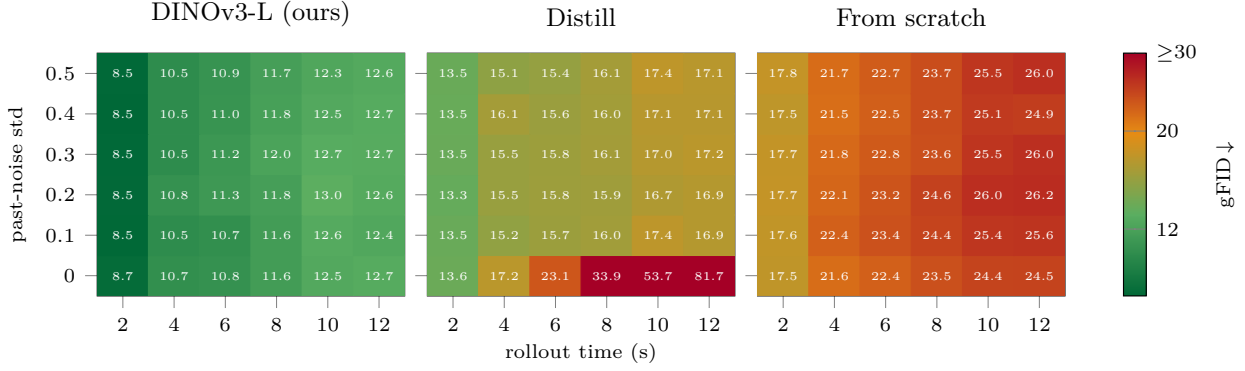}}
  \caption{\textbf{Context noise at inference.} gFID as a function of the past-noise standard
  deviation (vertical) and rollout time (horizontal), for three representative codecs; cell values
  are gFID and color runs green (low) to red (high, clipped at $30$). The distilled codec's quality
  collapses over the rollout when the past is fed clean (noise $0$) and is rescued by a small amount
  of noise, while the pretrained and from-scratch codecs are flat in noise. The other pretrained
  feature extractors and the gFVD/gFDD metrics show the same pattern (\cref{app:noise-heatmaps}).}
  \label{fig:exp-noise}
\end{figure}

\paragraph{Should the context be noised at inference?} Diffusion forcing trains on partially
noised context, so at inference the past latents can be fed clean or re-noised to a chosen
level; adding noise to the context is a common way to curb drift in autoregressive diffusion
models. We roll out while injecting Gaussian noise into the past latents at standard
deviations from $0$ to $0.5$, and track gFID across the rollout for every codec, so any
instability from rolling out on clean latents becomes visible as it compounds over time
(\cref{fig:exp-noise}).

\begin{takeaway}{Should the context be noised at inference?} For most codecs, no, but it
rescues the fragile one. The pretrained and from-scratch codecs are flat across noise levels
even at the $12$\,s horizon (\cref{fig:exp-noise}), so they roll out on clean past frames just as
well. The distilled codec is the exception: fed a clean past its quality collapses at the long
horizon, and a little noise restores it. A light default therefore costs the stable codecs nothing
and keeps the fragile one from diverging.\end{takeaway}

\begin{figure}[htbp]
  \centering
  \begin{subfigure}{0.32\linewidth}\centering
    \begin{tikzpicture}
    \begin{axis}[rsplot, width=\linewidth, height=0.92\linewidth,
      xmin=1, xmax=10, xtick={1,2,4,6,8,10},
      ymin=0, ymax=20, xlabel={flow matching steps}, ylabel={gFID\,$\downarrow$},
      legend style={at={(1.0,0.03)}, anchor=south east, font=\scriptsize}]
      \addlegendimage{plotblue, line width=1.3pt}\addlegendentry{baseline}
      \addlegendimage{plotorange, line width=1.3pt}\addlegendentry{PSD}
      \addplot[plotblue, line width=1.3pt, mark=*, forget plot] coordinates {(1,17.76)(2,12.44)(4,10.82)(6,10.61)(8,10.51)(10,9.93)};
      \addplot[plotorange, line width=1.3pt, mark=*, forget plot] coordinates {(1,10.25)(2,10.02)(4,9.68)(6,9.69)(8,9.46)(10,9.44)};
    \end{axis}
    \end{tikzpicture}
    \caption{gFID}
  \end{subfigure}\hfill
  \begin{subfigure}{0.32\linewidth}\centering
    \begin{tikzpicture}
    \begin{axis}[rsplot, width=\linewidth, height=0.92\linewidth,
      xmin=1, xmax=10, xtick={1,2,4,6,8,10},
      ymin=0, ymax=320, xlabel={flow matching steps}, ylabel={gFVD\,$\downarrow$}]
      \addplot[plotblue, line width=1.3pt, mark=*] coordinates {(1,307.1)(2,235.1)(4,189.3)(6,170.3)(8,163.4)(10,150.8)};
      \addplot[plotorange, line width=1.3pt, mark=*] coordinates {(1,147.4)(2,136.4)(4,128.4)(6,126.6)(8,127.7)(10,125.0)};
    \end{axis}
    \end{tikzpicture}
    \caption{gFVD}
  \end{subfigure}\hfill
  \begin{subfigure}{0.32\linewidth}\centering
    \begin{tikzpicture}
    \begin{axis}[rsplot, width=\linewidth, height=0.92\linewidth,
      xmin=1, xmax=10, xtick={1,2,4,6,8,10},
      ymin=0, ymax=1.3, xlabel={flow matching steps}, ylabel={gFDD\,$\downarrow$}]
      \addplot[plotblue, line width=1.3pt, mark=*] coordinates {(1,1.181)(2,0.752)(4,0.595)(6,0.566)(8,0.538)(10,0.486)};
      \addplot[plotorange, line width=1.3pt, mark=*] coordinates {(1,0.466)(2,0.427)(4,0.402)(6,0.398)(8,0.384)(10,0.392)};
    \end{axis}
    \end{tikzpicture}
    \caption{gFDD}
  \end{subfigure}
  \caption{\textbf{Few-step sampling.} Generation quality against the number of flow-matching
  steps, for the baseline and the PSD self-distilled model (trained from scratch).}
  \label{fig:fewstep}
\end{figure}
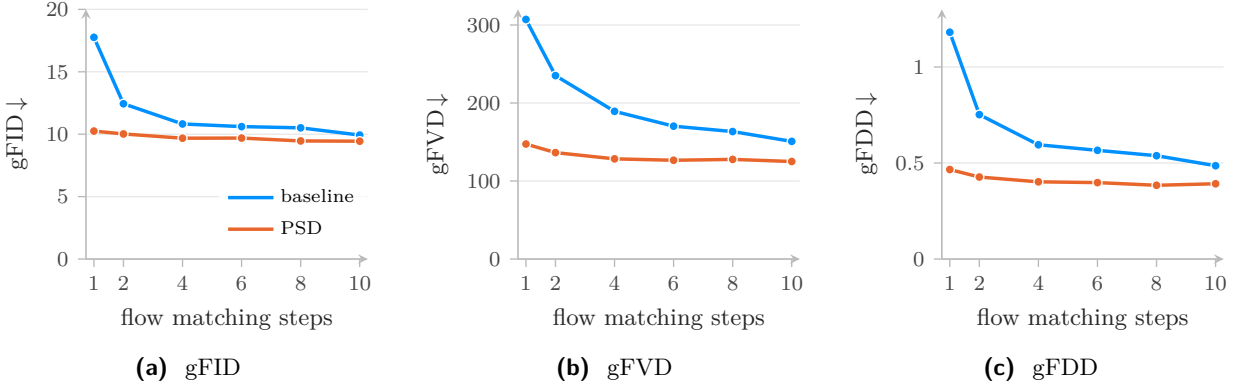

\paragraph{How does few-step distillation affect quality and inference cost?} Sampling integrates
the flow-matching field over many steps, and this step count sets the inference cost.
Self-distillation can shorten it: the model is distilled so that one large step replaces two
smaller ones, an idea introduced concurrently as progressive self-distillation~\citep{boffi2025consistency} and shortcut models~\citep{frans2025one}. We compare two settings on the baseline codec: the standard world model, and a model
trained from scratch with the self-distillation objective (PSD), applied stochastically on a
random $10\%$ of the training updates. \cref{fig:fewstep} traces each generation metric
against the number of flow-matching steps.

\begin{takeaway}{How does few-step distillation affect quality and inference cost?} Yes. PSD
outperforms the baseline at every step count, by a wide margin in the few-step regime, while staying stable even when using far fewer flow-matching steps.\end{takeaway}

\subsection{Controllability}
\label{sec:exp-controllability}

Conditioning the world model on actions does not guarantee that it obeys them. We measure how
faithfully the model renders the commanded actions with the action recoverability ratio (ARR,
\cref{sec:metrics}), evaluated on the single-player model.

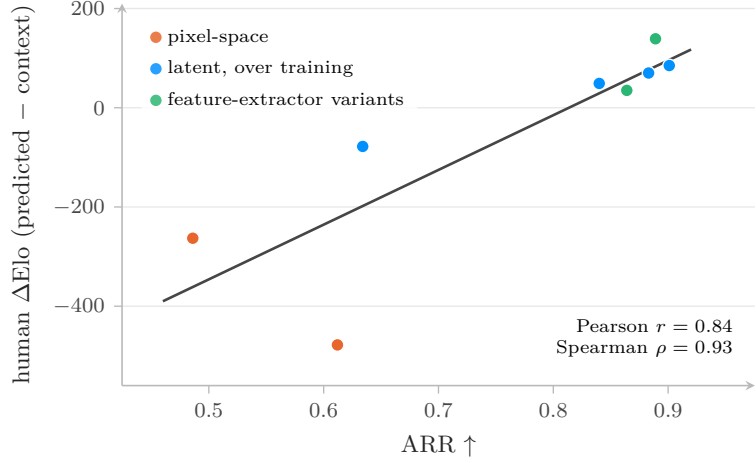
\begin{figure}[htbp]
  \centering
  \begin{tikzpicture}
  \begin{axis}[rsplot, width=0.6\linewidth, height=0.4\linewidth,
    xlabel={ARR $\uparrow$}, ylabel={human $\Delta$Elo (predicted $-$ context)},
    xmin=0.44, xmax=0.96, ymin=-560, ymax=210,
    xtick={0.5,0.6,0.7,0.8,0.9}, ytick={-400,-200,0,200},
    legend style={at={(0.03,0.97)}, anchor=north west, font=\scriptsize, draw=none,
                  fill=white, fill opacity=0.85, text opacity=1},
    legend cell align=left]
    \addplot[gray!55!black, line width=1pt, forget plot, domain=0.46:0.92, samples=2]{1103*x - 897.4};
    \addplot[plotorange, only marks, mark=*, mark size=2.4pt] coordinates {(0.612,-478)(0.486,-263)};
    \addlegendentry{pixel-space}
    \addplot[plotblue, only marks, mark=*, mark size=2.4pt] coordinates {(0.634,-78)(0.840,49)(0.883,70)(0.901,85)};
    \addlegendentry{latent, over training}
    \addplot[plotgreen, only marks, mark=*, mark size=2.4pt] coordinates {(0.864,35)(0.889,139)};
    \addlegendentry{feature-extractor variants}
    \node[anchor=south east, font=\scriptsize, align=right] at (rel axis cs:0.98,0.05)
      {Pearson $r=0.84$\\Spearman $\rho=0.93$};
  \end{axis}
  \end{tikzpicture}
  \caption{\textbf{ARR tracks human judgment of controllability.} Automatic ARR against the human
  action-adherence preference (an Elo delta, predicted vs.\ context) for the models rated in both,
  spanning pixel-space, feature-extractor, and training-checkpoint comparisons. The two are strongly
  correlated (Pearson $r = 0.84$) and rank the models almost identically (Spearman $\rho = 0.93$); the
  line is a linear least-squares fit.}
  \label{fig:arr-human}
\end{figure}

\paragraph{Does the Action Recoverability Ratio (ARR) agree with human judgment?} Before using ARR to study the model, we check that it
reflects how people perceive controllability: we run a human action-adherence study
(\cref{sec:human-eval}) and plot ARR against the human preference for the models rated in both
(\cref{fig:arr-human}).

\begin{takeaway}{Does the Action Recoverability Ratio (ARR) agree with human judgment?} Closely. ARR and human action-adherence
preference rank the models almost identically, so ARR captures controllability as human raters
perceive it.\end{takeaway}

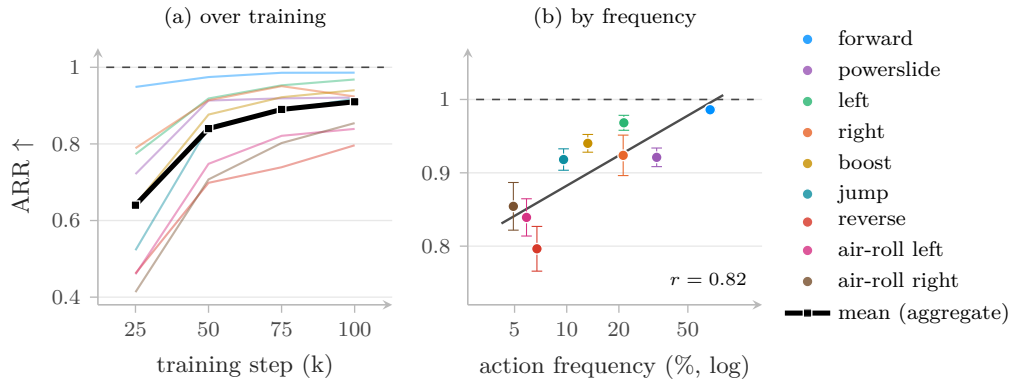
\begin{figure}[htbp]
  \centering
  \begin{tikzpicture}
  % ---- (a) ARR over training: faint per-action curves, bold black aggregate ----
  \begin{axis}[name=ax1, rsplot, width=0.33\linewidth, height=0.30\linewidth,
    ymin=0.38, ymax=1.05, xmin=15, xmax=110, xtick={25,50,75,100},
    xlabel={training step (k)}, ylabel={ARR $\uparrow$},
    title={\footnotesize (a) over training}, title style={yshift=-2pt},
    every axis plot/.append style={line width=0.8pt, mark=none, opacity=0.45}]
    \draw[dashed, gray!55!black, semithick] (axis cs:15,1) -- (axis cs:110,1);
    \addplot[acForward]    coordinates {(25,0.9487)(50,0.9745)(75,0.9857)(100,0.986)};
    \addplot[acPowerslide] coordinates {(25,0.7212)(50,0.9132)(75,0.9189)(100,0.9211)};
    \addplot[acLeft]       coordinates {(25,0.7732)(50,0.9185)(75,0.9529)(100,0.9682)};
    \addplot[acRight]      coordinates {(25,0.7888)(50,0.9139)(75,0.9508)(100,0.9238)};
    \addplot[acBoost]      coordinates {(25,0.6419)(50,0.8766)(75,0.9219)(100,0.9402)};
    \addplot[acJump]       coordinates {(25,0.5224)(50,0.8447)(75,0.8873)(100,0.9181)};
    \addplot[acReverse]    coordinates {(25,0.4619)(50,0.6984)(75,0.7391)(100,0.7964)};
    \addplot[acAirL]       coordinates {(25,0.4597)(50,0.748)(75,0.8211)(100,0.8392)};
    \addplot[acAirR]       coordinates {(25,0.413)(50,0.7073)(75,0.8023)(100,0.8543)};
    \addplot[black, mark=square*, mark size=1.6pt, line width=1.9pt, opacity=1]
      coordinates {(25,0.64)(50,0.84)(75,0.89)(100,0.91)};
  \end{axis}
  % ---- (b) ARR by frequency: per-action points, regression, shared legend ----
  \begin{axis}[name=ax2, rsplot, width=0.33\linewidth, height=0.30\linewidth, xmode=log,
    at={($(ax1.east)+(1.0cm,0)$)}, anchor=west,
    ymin=0.72, ymax=1.07, xmin=3.0, xmax=120, xtick={5,10,20,50}, xticklabels={5,10,20,50},
    xlabel={action frequency (\%, log)}, ylabel={},
    title={\footnotesize (b) by frequency}, title style={yshift=-2pt},
    legend style={at={(1.04,0.5)}, anchor=west, font=\footnotesize, draw=none, fill=none,
                  cells={anchor=west}, row sep=0.6pt},
    legend cell align=left]
    \draw[dashed, gray!55!black, semithick] (axis cs:3.0,1) -- (axis cs:120,1);
    \addplot[gray!60!black, line width=0.9pt, forget plot] coordinates {(4.2,0.8309)(80,1.006)};
    \addplot[acForward, only marks, mark=*, mark size=2pt, error bars/.cd, y dir=both, y explicit]    coordinates {(67.23,0.986)  +- (0,0.0027)};  \addlegendentry{forward}
    \addplot[acPowerslide, only marks, mark=*, mark size=2pt, error bars/.cd, y dir=both, y explicit] coordinates {(33.04,0.9211) +- (0,0.0127)}; \addlegendentry{powerslide}
    \addplot[acLeft, only marks, mark=*, mark size=2pt, error bars/.cd, y dir=both, y explicit]       coordinates {(21.36,0.9682) +- (0,0.0102)}; \addlegendentry{left}
    \addplot[acRight, only marks, mark=*, mark size=2pt, error bars/.cd, y dir=both, y explicit]      coordinates {(21.17,0.9238) +- (0,0.0276)}; \addlegendentry{right}
    \addplot[acBoost, only marks, mark=*, mark size=2pt, error bars/.cd, y dir=both, y explicit]      coordinates {(13.23,0.9402) +- (0,0.0121)}; \addlegendentry{boost}
    \addplot[acJump, only marks, mark=*, mark size=2pt, error bars/.cd, y dir=both, y explicit]       coordinates {(9.59,0.9181)  +- (0,0.0146)}; \addlegendentry{jump}
    \addplot[acReverse, only marks, mark=*, mark size=2pt, error bars/.cd, y dir=both, y explicit]    coordinates {(6.73,0.7964)  +- (0,0.0305)}; \addlegendentry{reverse}
    \addplot[acAirL, only marks, mark=*, mark size=2pt, error bars/.cd, y dir=both, y explicit]       coordinates {(5.86,0.8392)  +- (0,0.0254)}; \addlegendentry{air-roll left}
    \addplot[acAirR, only marks, mark=*, mark size=2pt, error bars/.cd, y dir=both, y explicit]       coordinates {(4.92,0.8543)  +- (0,0.0325)}; \addlegendentry{air-roll right}
    \addlegendimage{black, line width=1.9pt, mark=square*, mark size=1.6pt}\addlegendentry{mean (aggregate)}
    \node[anchor=south east, font=\scriptsize] at (rel axis cs:0.98,0.04) {$r=0.82$};
  \end{axis}
  \end{tikzpicture}
  \caption{\textbf{Controllability.} \textbf{(a)} Per-action ARR (faint) and their aggregate (bold
  black) against training step for the single-player model. \textbf{(b)} Per-action ARR at the final
  checkpoint against training frequency, with bootstrap $95\%$ confidence intervals and a log-linear
  fit ($r=0.82$). Colors identify the nine controls across both panels (shared legend); the dashed
  line marks a faithful reconstruction ($\mathrm{ARR}=1$).}
  \label{fig:controllability}
\end{figure}

\paragraph{How does controllability improve over training?} We measure controllability over the
course of training, computing aggregate ARR at successive checkpoints of the single-player model
together with the nine per-action curves (\cref{fig:controllability}a), and read it against how
generation quality converges over the same run (\cref{fig:gfid-train}).

\begin{takeaway}{How does controllability improve over training?} Gradually, and it keeps
improving after image quality has saturated. Aggregate ARR rises steadily and every action becomes
more recoverable, approaching the reconstruction ceiling. Generation quality converges far earlier
(\cref{fig:gfid-train}): the model first learns to render realistic frames, then continues to refine
how faithfully it obeys the commanded actions.\end{takeaway}

\paragraph{Are rarer actions harder to control?} We test whether infrequent actions are modeled
worse, plotting each control's ARR at the final checkpoint against how often it appears in the
training data, across the nine actions (\cref{fig:controllability}b).

\begin{takeaway}{Are rarer actions harder to control?} Yes. ARR rises with an action's
training frequency: common controls are recovered near-perfectly while the rarest air-roll and
reverse inputs lag behind, so the model under-renders actions it rarely saw.\end{takeaway}

\subsection{Going from single-player to multiplayer}
\label{sec:exp-multiplayer}

\paragraph{How should a fixed budget be split between single- and multiplayer training?}
The world model must ultimately drive four interacting players (Section~\ref{sec:multiplayer});
\cref{fig:rollout2} shows a rollout from the multiplayer model, with the game-state probe's
predicted ball and car positions overlaid. A
multiplayer step tiles the four player views and ingests about $4\times$ as many views as a
single-player step (baseline configurations in \cref{tab:wm-hparams}), so at a fixed budget, matched to the single-player model's compute, multiplayer
training runs proportionally fewer steps. We ask how best to spend that budget to reach a multiplayer
model: on multiplayer alone, on single-player alone, or on a mix that pretrains on single-player and
then trains into multiplayer.

\begin{figure}[htbp]
  \centering
  \begin{tikzpicture}
  \begin{axis}[rsplot, width=0.66\linewidth, height=0.42\linewidth,
    xmin=-8, xmax=108, xtick={0,25,50,75,100},
    xticklabels={%
      {0\%\\SP\,0k\\MP\,25k},
      {25\%\\SP\,25k\\MP\,18.75k},
      {50\%\\SP\,50k\\MP\,12.5k},
      {75\%\\SP\,75k\\MP\,6.25k},
      {100\%\\SP\,100k\\MP\,0k}},
    x tick label style={font=\scriptsize, align=center},
    xlabel={single-player share of the budget},
    ymin=10, ymax=23, ytick={10,13,16,19,22},
    ylabel={gFID at $4$\,s\,$\downarrow$},
    legend style={at={(0.97,0.97)}, anchor=north east, font=\scriptsize, draw=none,
                  fill=white, fill opacity=0.85, text opacity=1, cells={anchor=west}}]
    \addplot[plotblue, mark=*, mark size=1.6pt, line width=1.3pt]
      coordinates {(0,22.361)(12.5,12.96)(25,11.578)(37.5,11.17)(50,11.482)(62.5,11.79)(75,12.861)(87.5,12.75)};
    \addlegendentry{multiplayer}
    \addplot[plotorange, only marks, mark=*, mark size=2.4pt] coordinates {(100,10.554)};
    \addlegendentry{single-player (ref.)}
  \end{axis}
  \end{tikzpicture}
  \caption{\textbf{Allocating a fixed budget between single- and multiplayer training.} gFID of one
  player's view at the $4$\,s horizon at the single-player model's budget ($1.6$M player-views). The
  single-player share sets how long we pretrain on single-player before continuing on multiplayer;
  each tick gives the resulting single-player and multiplayer optimization-step counts. Blue marks the
  multiplayer runs ($0\%$ is multiplayer from scratch); orange marks the pure single-player reference.
  Multiplayer from scratch collapses; pretraining on single-player first rescues it, and a modest
  single-player share is best, though at this budget pure single-player still edges ahead.}
  \label{fig:mp-data}
\end{figure}
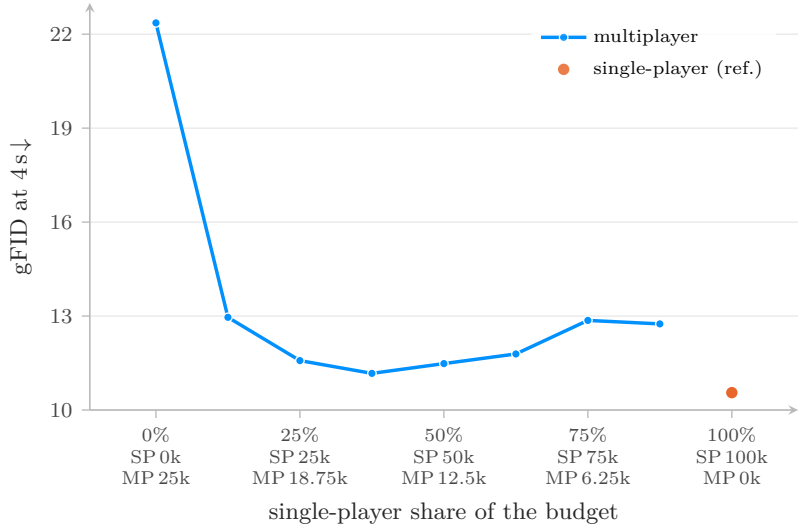

We fix the budget at the single-player model's own ($1.6$M player-views, $100$k single-player steps)
and sweep the single-player share from $0\%$ (multiplayer from scratch) to $100\%$ (pure
single-player); each intermediate split pretrains on single-player for that share of the budget, then
continues on multiplayer with the remainder, warm-starting from the single-player checkpoint. The
$x$-axis reports the single-player and multiplayer optimization-step counts at each split. The $100\%$
point is the single-player model itself, a per-view reference evaluated with the single-player
pipeline. All runs share the same codec, world model, and data, and we score one player's view at the
$4$\,s horizon (\cref{fig:mp-data}).

\begin{takeaway}{How should a fixed budget be split between single- and multiplayer training?} 
Pretrain on single-player, then continue on multiplayer. Training multiplayer from scratch collapses
at this budget; pretraining on single-player first and then continuing on multiplayer recovers most
of the gap, and a modest single-player share works best. Pure single-player is stronger on per-view
quality here, but it never models the four-player interaction, so it serves only as a reference.\end{takeaway}

At a larger budget the collapse disappears: with twice the clips ($3.2$M player-views), multiplayer
trains adequately even from scratch (gFID $9.9$ at $4$\,s), and a mostly-multiplayer mix ($75\%$
multiplayer, $25\%$ single-player) does better still (gFID $9.4$; \cref{fig:mp-data-3p2m} in
\cref{app:mp-larger}).

\subsection{Scaling behavior}
\label{sec:exp-scaling}
We study how the single-player world model scales along two axes: the amount of training data
and the size of the model.

\begin{figure}[htbp]
  \centering
  \begin{tikzpicture}
  \begin{axis}[name=ax1, rsplot, width=0.42\linewidth, height=0.34\linewidth, xmode=log,
    xmin=8, xmax=2600, xtick={10,100,1000}, xticklabels={10,100,1000},
    x tick label style={font=\scriptsize}, xlabel={unique data (hours, log)},
    ymin=8, ymax=17, ytick={8,10,12,14,16}, ylabel={gFID at $4$\,s\,$\downarrow$},
    title={\footnotesize (a) image fidelity}, title style={yshift=-2pt}]
    \addplot[plotblue, mark=*, mark size=1.8pt, line width=1.4pt]
      coordinates {(10,16.397)(50,10.351)(100,10.328)(250,10.017)(500,10.208)(1000,10.119)(2000,10.554)};
  \end{axis}
  \begin{axis}[name=ax2, rsplot, width=0.42\linewidth, height=0.34\linewidth, xmode=log,
    at={($(ax1.east)+(1.1cm,0)$)}, anchor=west,
    xmin=8, xmax=2600, xtick={10,100,1000}, xticklabels={10,100,1000},
    x tick label style={font=\scriptsize}, xlabel={unique data (hours, log)},
    ymin=0.77, ymax=0.93, ytick={0.80,0.85,0.90}, ylabel={ARR\,$\uparrow$},
    title={\footnotesize (b) controllability}, title style={yshift=-2pt}]
    \addplot[plotorange, mark=square*, mark size=1.8pt, line width=1.4pt]
      coordinates {(10,0.788)(50,0.870)(100,0.881)(250,0.885)(500,0.885)(1000,0.888)(2000,0.910)};
  \end{axis}
  \end{tikzpicture}
  \caption{\textbf{Data scaling at fixed compute.} \textbf{(a)} gFID and \textbf{(b)} controllability
  ARR of the single-player model against the amount of unique training data seen, at a fixed
  $100$k-step budget (smaller subsets are revisited more, so unique data and repetition vary together).
  Below about $50$ hours both metrics collapse as distinct data runs short. Above it gFID saturates
  while ARR keeps climbing, so more unique data buys action fidelity that gFID no longer registers.}
  \label{fig:data-scaling}
\end{figure}
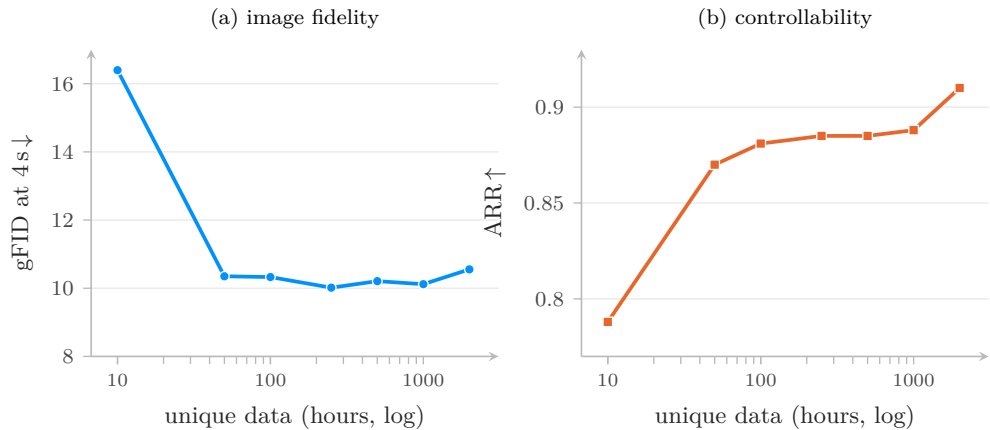

\paragraph{How much does more data help?} We isolate the value of unique data at a fixed compute
budget. Holding the model, batch size, optimizer, and total number of steps fixed at $100$k updates,
we train the $1$B world model on nested random subsets of the single-player data. At this budget a
run sees about $1.6$M clips, so the unique data it covers is capped at roughly $2{,}000$ hours;
smaller subsets are therefore revisited more often, up to about $18\times$ at $100$ hours and down to
a single pass for the full data. Compute is identical across points; only the amount of unique data,
and with it the degree of repetition, changes. We report gFID and controllability (ARR) at
the $4$\,s horizon (\cref{fig:data-scaling}).

\begin{takeaway}{How much does more data help?} It depends on what is measured, and on whether
the model has enough data to begin with. When unique data is very scarce, both per-frame fidelity
(gFID) and controllability (ARR) collapse. Beyond the data-starved regime, more data leaves per-frame
appearance unchanged at this budget while steadily improving how faithfully the model follows the
commanded actions.\end{takeaway}

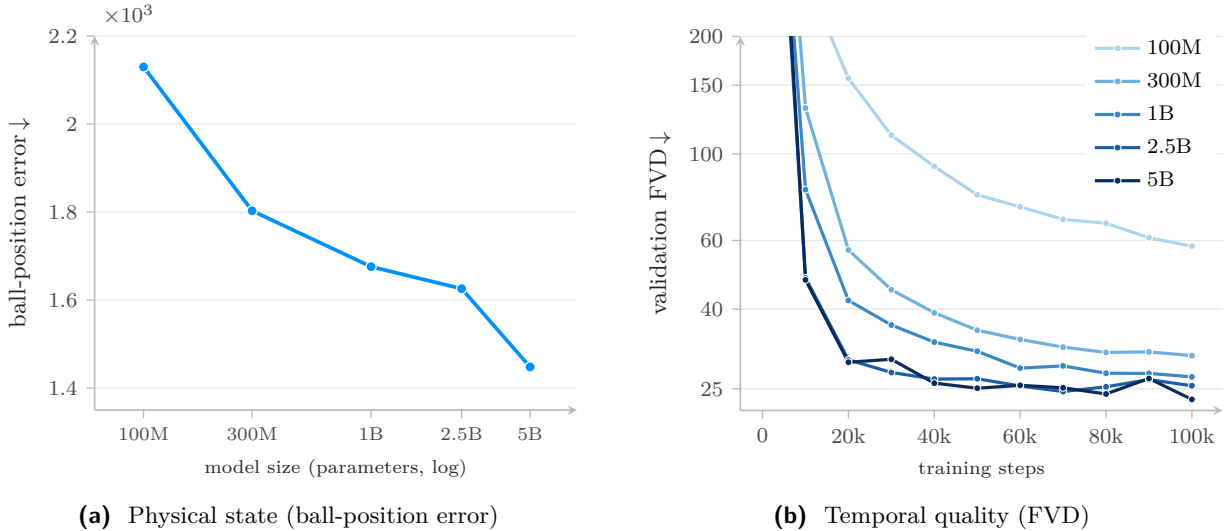
\begin{figure}[htbp]
  \centering
  % Model-size scaling training curves: validation FID / FVD vs training step,
% one curve per model size (100M -> 5B), colored light -> dark. Data parsed from
% fid_scaling.csv and fvd_scaling.csv (repo root), truncated to <=100k steps;
% coordinates are inlined here (in thousands of steps on the x-axis) so the
% figure compiles standalone.
\pgfplotsset{
  modelcurves/.style={
    rsplot,
    width=\linewidth, height=0.82\linewidth,
    xmin=-2, xmax=104,
    xtick={0,20,40,60,80,100},
    xticklabels={0,20k,40k,60k,80k,100k},
    xlabel={training steps}, xlabel style={font=\scriptsize},
    log ticks with fixed point,
    legend style={at={(0.985,1.03)}, anchor=north east, fill=white, fill opacity=1, text opacity=1},
    every axis plot/.append style={line width=1.2pt, mark=*, mark size=1.3pt,
      mark options={solid, draw=white, line width=0.4pt}},
  },
}
\begin{subfigure}{0.48\linewidth}\centering
  \begin{tikzpicture}
  \begin{axis}[rsplot, width=\linewidth, height=0.82\linewidth, xmode=log,
      xmin=0.07, xmax=7, xtick={0.1,0.3,1,2.5,5}, xticklabels={100M,300M,1B,2.5B,5B},
      log ticks with fixed point, x tick label style={font=\scriptsize},
      xlabel={model size (parameters, log)}, xlabel style={font=\scriptsize},
      ymin=1350, ymax=2200, ytick={1400,1600,1800,2000,2200},
      scaled y ticks={base 10:-3}, ytick scale label code/.code={$\times 10^{3}$},
      ylabel={ball-position error\,$\downarrow$}]
    \addplot[plotblue, mark=*, mark size=1.9pt, line width=1.4pt]
      coordinates {(0.1,2130)(0.3,1803)(1,1676)(2.5,1626)(5,1448)};
  \end{axis}
  \end{tikzpicture}\caption{Physical state (ball-position error)}
\end{subfigure}\hfill
\begin{subfigure}{0.48\linewidth}\centering
  \begin{tikzpicture}
  \begin{axis}[ymode=log, modelcurves, ymin=22, ymax=200, ylabel={validation FVD\,$\downarrow$},
      ytick={25,40,60,100,150,200}, minor ytick={}]
    \addplot[color=sizeA] coordinates {(0.001,2707.53)(1,2678.93)(10,284.05)(20,156.23)(30,111.63)(40,92.87)(50,78.55)(60,73.17)(70,67.97)(80,66.39)(90,60.97)(100,58.00)};
    \addplot[color=sizeB] coordinates {(0.001,2784.00)(1,2386.68)(10,131.10)(20,56.71)(30,44.86)(40,39.16)(50,35.33)(60,33.46)(70,31.98)(80,30.99)(90,31.09)(100,30.41)};
    \addplot[color=sizeC] coordinates {(0.001,2450.91)(1,1641.17)(10,80.97)(20,42.12)(30,36.43)(40,32.93)(50,31.20)(60,28.25)(70,28.62)(80,27.41)(90,27.38)(100,26.83)};
    \addplot[color=sizeD] coordinates {(0.001,2590.48)(1,1889.37)(10,48.11)(20,29.66)(30,27.53)(40,26.48)(50,26.53)(60,25.42)(70,24.60)(80,25.30)(90,26.39)(100,25.46)};
    \addplot[color=sizeE] coordinates {(0.001,3278.13)(1,2320.63)(10,47.54)(20,29.25)(30,29.76)(40,25.85)(50,25.09)(60,25.52)(70,25.15)(80,24.25)(90,26.57)(100,23.48)};
    \legend{100M,300M,1B,2.5B,5B}
  \end{axis}
  \end{tikzpicture}\caption{Temporal quality (FVD)}
\end{subfigure}
  \caption{\textbf{Larger models improve generation quality and encode physical state more faithfully.}
  \textbf{(a)} Ball-position readout error of the linear game-state probe (L2, in Unreal units)
  against model size, and \textbf{(b)} validation FVD throughout training, for five multiplayer world
  models spanning $100$M to $5$B parameters (light to dark), with the data and training recipe held
  fixed. Lower is better on both axes; the FVD $y$-axis is logarithmic, cropped to the converged
  regime. Larger models converge faster and to lower error, their ordering by size holds for almost
  the entire run, and the probe error decreases monotonically with model size. Returns diminish beyond
  $2.5$B, where the two largest models reach comparable quality.}
  \label{fig:model-scaling-curves}
\end{figure}

\paragraph{How does quality scale with model size?} Following standard scaling practice, we
grow the multiplayer world model across five sizes from $100$M to $5$B parameters ($100$M, $300$M, $1$B,
$2.5$B, $5$B), holding the data and training recipe fixed so that any difference in quality is
attributable to capacity alone; the architecture dimensions and training schedules for each size are
given in \cref{app:impl}. Throughout training we track validation FID (image quality) and
FVD (temporal coherence), and observe a clear and consistent scaling trend: larger models train to
lower error and converge faster, and their ordering by size is preserved for essentially the entire
run (\cref{fig:model-scaling-curves}). The gains are steepest from $100$M to $1$B (the $100$M model
plateaus at a markedly higher error), after which returns diminish, with the $2.5$B and $5$B
models converging to comparable quality at this fixed data budget. As a complementary,
capacity-sensitive probe of physical understanding, we also measure how well a linear classifier
recovers the game's physical state (ball and car positions and velocities) from the model's frozen
features.

The probe tells the same story from the side of physical understanding: the ball-position readout
error falls monotonically with capacity, from $2130$ at $100$M to $1448$ at $5$B, with the largest drop
between the $100$M and $300$M models (\cref{fig:model-scaling-curves}). Larger models therefore not only look
better but encode the game's physical state more faithfully in their features.

\begin{takeaway}{How does quality scale with model size?} Generation quality improves
monotonically with model size: larger models reach lower FID and FVD and converge faster, with the
steepest gains at smaller scales and diminishing returns for the largest models at this fixed data
budget. The game-state probe corroborates this, with ball-position readout error decreasing as
capacity grows.
\end{takeaway}

\subsection{Emergent properties}
\label{sec:exp-emergent}

The model is trained only to predict future frames from past frames and actions, yet its
rollouts show behavior it was never explicitly supervised on. We highlight four such
properties.

\paragraph{The model emulates a theory of mind for players whose actions are withheld.}
During training we randomly drop some players' action streams, so the model must predict
those players' behavior without actions to condition on. At rollout time it fills the gap
itself: an uncommanded car keeps playing plausibly, moving and contesting the ball as a real
player would, with no action supervision. By predicting how each unconditioned player would
act, the model learns a policy for these agents on top of the environment's dynamics. As discussed in the data generation pipeline (\cref{sec:data-physics}), policies used to generate the action-annotated clips have access to the full game state, far more than the
world model receives, yet the model recovers their complex decisions from pixels alone.

\begin{figure}[htbp]
  \centering
  \newlength{\attnimw}\setlength{\attnimw}{0.2\linewidth}
  \newcommand{\plab}[2]{{\footnotesize\bfseries\textcolor{#1}{Player #2}}}
  \newcommand{\im}[1]{\includegraphics[width=\attnimw]{figures/mp_attention/#1.png}}
  \setlength{\tabcolsep}{2.5pt}\renewcommand{\arraystretch}{1.1}
  \begin{tabular}{@{}>{\centering\arraybackslash}m{2.05cm}@{\hspace{4pt}}*{4}{>{\centering\arraybackslash}m{\attnimw}}@{}}
     & \plab{plotblue}{1} & \plab{plotblue}{2} & \plab{plotorange}{3} & \plab{plotorange}{4} \\[1pt]
    {\scriptsize generated views} & \im{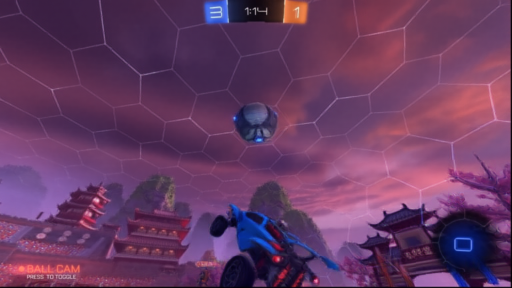} & \im{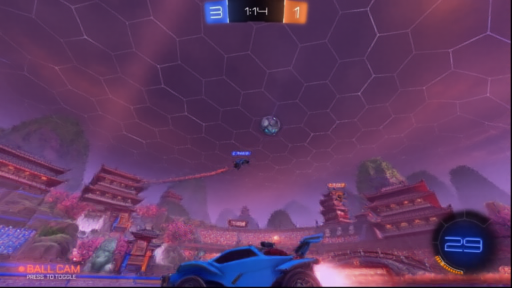} & \im{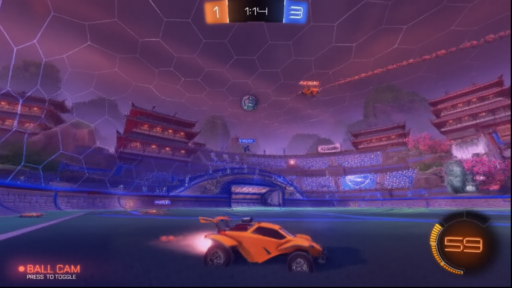} & \im{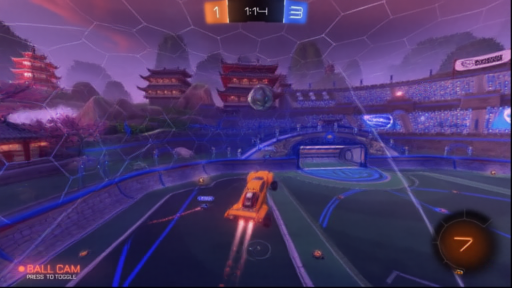} \\
    {\scriptsize Player 1's car, through Player 2's view} & \im{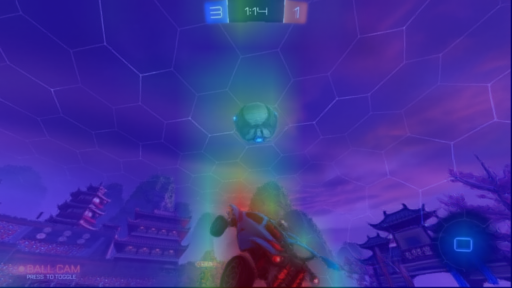} & \im{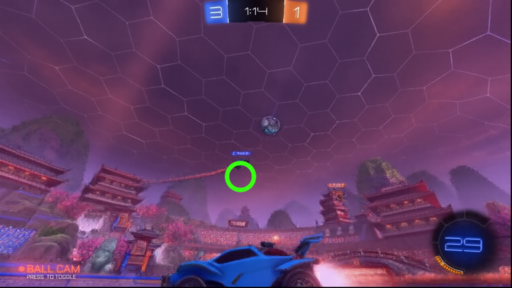} & \im{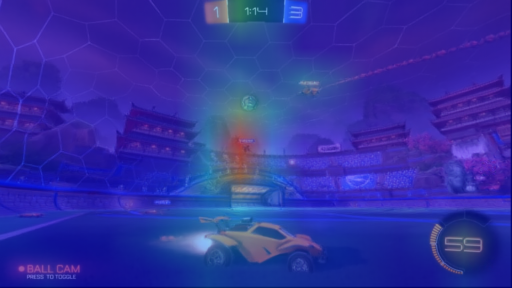} & \im{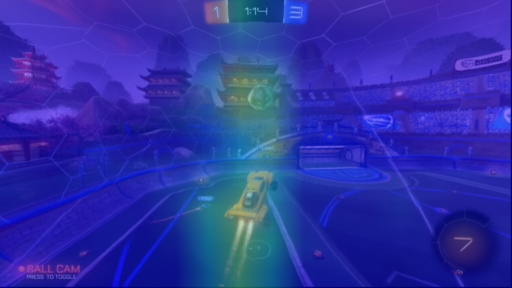} \\
    {\scriptsize the ball, through Player 1's view} & \im{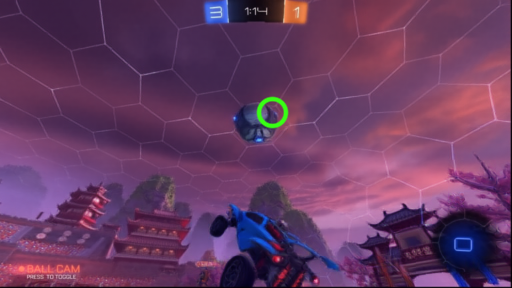} & \im{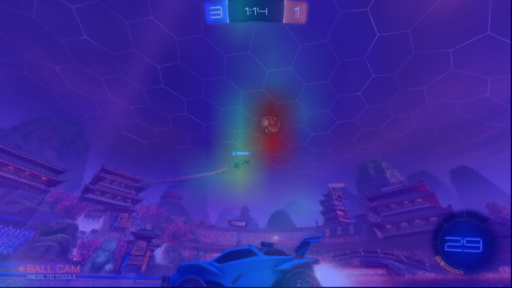} & \im{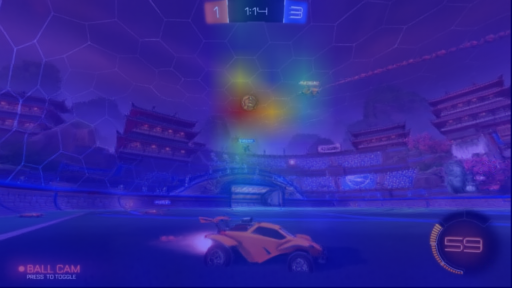} & \im{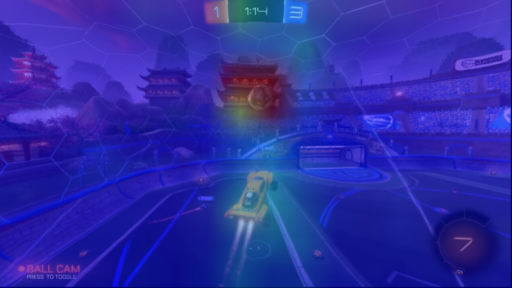} \\
    {\scriptsize the timer, through Player 1's view} & \im{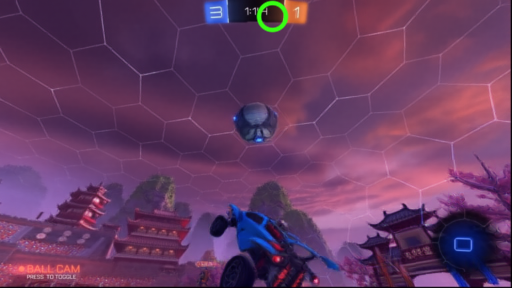} & \im{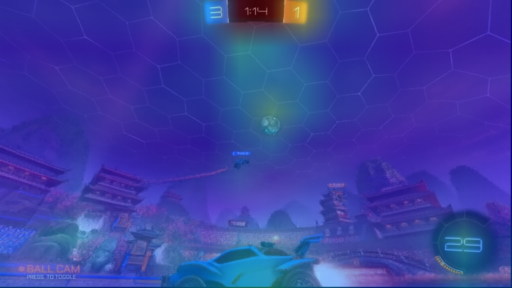} & \im{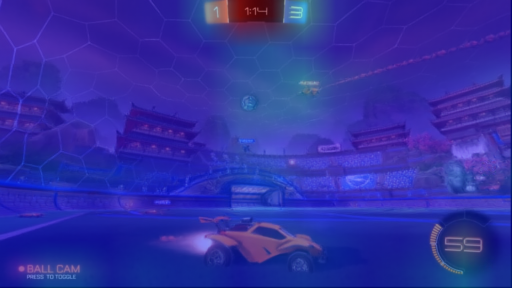} & \im{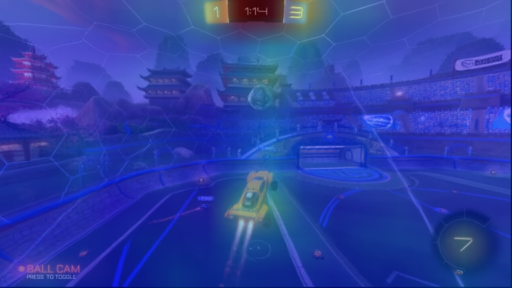} \\
  \end{tabular}
  \caption{\textbf{The multiplayer model attends across views to shared objects and state.} Cross-view
  spatial attention on a rollout generated by the multiplayer world model. Each column is one player's
  generated view. After the generated views, each row queries the token named on the left and
  shows where its attention lands in the other views, warm colors marking strong attention and cool
  colors weak. Player~1's car, queried through Player~2's view,
  is picked out in the other views that see it and stays cooler where it is off-screen. The ball, queried
  through Player~1's view, activates in every view. The timer, queried through Player~1's view, attends
  to the shared top-center HUD band of every view. Cross-view attention is spatially structured; averaged over layers~$4$--$7$.}
  \label{fig:mp-attention}
\end{figure}

\paragraph{The model keeps the four views mutually consistent.} The multiplayer model renders all
four players' views jointly and keeps them agreeing on the shared world: a car that leaves one view
stays present in the others, and when it re-enters that view it reappears where those views place it.
The same coherence holds for salient game events: when the ball is driven into the net, the
``SCORED!'' callout and the goal explosion appear together across all four views (\cref{fig:goal}),
and a demolition is rendered consistently from every camera: the ``DEMOLITION'' callout for the
attacker and the ``BOOM!'' blast filling the victim's own view, seen as a distant burst by the others
(\cref{fig:rollout_destroy}).
A single-player model sees only one view, so it has no cross-view signal to fall back on and loses or
merges off-screen cars, a limitation we discuss in \cref{sec:exp-failure}. This consistency is
visible in the model's attention: a query token placed on an object in one view attends to the same
object in the other three views (\cref{fig:mp-attention}).

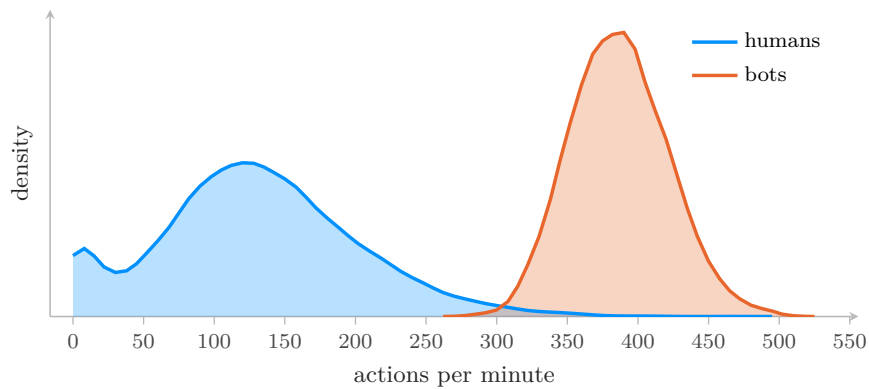
\begin{figure}[htbp]
  \centering
  \begin{tikzpicture}
  \begin{axis}[rsplot, width=0.74\linewidth, height=0.34\linewidth,
    xmin=0, xmax=540, ymin=0, ymax=1.08,
    xlabel={actions per minute}, ylabel={density}, ytick=\empty,
    legend pos=north east, legend style={font=\footnotesize, draw=none, fill=none}]
    \addplot[draw=none, fill=plotblue, fill opacity=0.28, forget plot] coordinates {(0,0) (0,0.2146) (8,0.2395) (15,0.2130) (22,0.1744) (30,0.1552) (38,0.1611) (45,0.1859) (52,0.2226) (60,0.2662) (68,0.3140) (75,0.3647) (82,0.4151) (90,0.4595) (98,0.4928) (105,0.5156) (112,0.5319) (120,0.5412) (128,0.5394) (135,0.5267) (142,0.5077) (150,0.4845) (158,0.4553) (165,0.4203) (172,0.3827) (180,0.3465) (188,0.3135) (195,0.2838) (202,0.2563) (210,0.2303) (218,0.2060) (225,0.1827) (232,0.1594) (240,0.1375) (248,0.1183) (255,0.1007) (262,0.0845) (270,0.0714) (278,0.0616) (285,0.0535) (292,0.0458) (300,0.0387) (308,0.0324) (315,0.0265) (322,0.0213) (330,0.0177) (338,0.0155) (345,0.0139) (352,0.0117) (360,0.0089) (368,0.0060) (375,0.0037) (382,0.0024) (390,0.0019) (398,0.0018) (405,0.0017) (412,0.0014) (420,0.0009) (428,0.0004) (435,0.0002) (442,0.0001) (450,0.0001) (458,0.0003) (465,0.0004) (472,0.0003) (480,0.0002) (488,0.0001) (495,0.0000) (540,0)} \closedcycle;
    \addplot[draw=none, fill=plotorange, fill opacity=0.28, forget plot] coordinates {(0,0) (262,0.0001) (270,0.0009) (278,0.0035) (285,0.0082) (292,0.0131) (300,0.0232) (308,0.0529) (315,0.1076) (322,0.1825) (330,0.2822) (338,0.4120) (345,0.5520) (352,0.6809) (360,0.8149) (368,0.9240) (375,0.9703) (382,0.9929) (390,1.0000) (398,0.9410) (405,0.8253) (412,0.7273) (420,0.6229) (428,0.4924) (435,0.3801) (442,0.2828) (450,0.1945) (458,0.1320) (465,0.0903) (472,0.0626) (480,0.0400) (488,0.0277) (495,0.0194) (502,0.0085) (510,0.0028) (518,0.0008) (525,0.0001) (540,0)} \closedcycle;
    \addplot[draw=plotblue, line width=1.3pt] coordinates {(0,0.2146) (8,0.2395) (15,0.2130) (22,0.1744) (30,0.1552) (38,0.1611) (45,0.1859) (52,0.2226) (60,0.2662) (68,0.3140) (75,0.3647) (82,0.4151) (90,0.4595) (98,0.4928) (105,0.5156) (112,0.5319) (120,0.5412) (128,0.5394) (135,0.5267) (142,0.5077) (150,0.4845) (158,0.4553) (165,0.4203) (172,0.3827) (180,0.3465) (188,0.3135) (195,0.2838) (202,0.2563) (210,0.2303) (218,0.2060) (225,0.1827) (232,0.1594) (240,0.1375) (248,0.1183) (255,0.1007) (262,0.0845) (270,0.0714) (278,0.0616) (285,0.0535) (292,0.0458) (300,0.0387) (308,0.0324) (315,0.0265) (322,0.0213) (330,0.0177) (338,0.0155) (345,0.0139) (352,0.0117) (360,0.0089) (368,0.0060) (375,0.0037) (382,0.0024) (390,0.0019) (398,0.0018) (405,0.0017) (412,0.0014) (420,0.0009) (428,0.0004) (435,0.0002) (442,0.0001) (450,0.0001) (458,0.0003) (465,0.0004) (472,0.0003) (480,0.0002) (488,0.0001) (495,0.0000)};
    \addlegendentry{humans}
    \addplot[draw=plotorange, line width=1.3pt] coordinates {(262,0.0001) (270,0.0009) (278,0.0035) (285,0.0082) (292,0.0131) (300,0.0232) (308,0.0529) (315,0.1076) (322,0.1825) (330,0.2822) (338,0.4120) (345,0.5520) (352,0.6809) (360,0.8149) (368,0.9240) (375,0.9703) (382,0.9929) (390,1.0000) (398,0.9410) (405,0.8253) (412,0.7273) (420,0.6229) (428,0.4924) (435,0.3801) (442,0.2828) (450,0.1945) (458,0.1320) (465,0.0903) (472,0.0626) (480,0.0400) (488,0.0277) (495,0.0194) (502,0.0085) (510,0.0028) (518,0.0008) (525,0.0001)};
    \addlegendentry{bots}
  \end{axis}
  \end{tikzpicture}
  \caption{\textbf{Humans play unlike the training policies.}
  Distribution of actions per minute across clips for the nine controls (forward, backward,
  left, right, jump, boost, powerslide, air-roll left and right). Both curves come from recorded
  gameplay, and neither is produced by the world model: the automated policies behind the training
  data (\emph{bots}) act fast and consistently, in a narrow band near $390$ actions per minute, while
  \emph{humans} act more slowly and far more variably, from near-idle to bursts and spread broadly
  below $350$. The human curve comes from a separate corpus of publicly shared human gameplay clips
  on the Medal.tv platform \citep{medaltv} (roughly $5{,}000$ hours). The plot documents the
  distribution shift the model must handle; in our live demo it stays coherent under human control.}
  \label{fig:apm}
\end{figure}

\paragraph{The model generalizes beyond the training action distribution.} At rollout the model
is routinely driven in ways whose statistics differ from training, and it stays stable under the
shift. For instance, a scene in which every car sits still never appears in the data, since
recorded play is always in motion; the model nonetheless holds such a scene stable, with cars
that neither drift nor accelerate on their own. It is equally robust to who is at the controls:
the automated policies that generate the training clips (\cref{sec:data-physics}) act fast and
uniformly, concentrating near $390$ actions per minute, while a person plays far more slowly and
variably, from near-idle to bursts and spread broadly below $350$ (\cref{fig:apm}); in live play we
observe that the model stays responsive and coherent under human control all the same.
Driven well outside the training
action distribution, it still obeys the underlying rule that a car moves only when acted on,
evidence that it has captured the game's dynamics themselves and transfers across playing styles.

\paragraph{Desynchronized views recover on their own.} The multiplayer model renders the four
players' views jointly. On rare occasions one view gradually desynchronizes from the others and
degrades into noise, typically when a player enters a regime the bots never produce,
such as staying still inside the goal, which is far out of distribution. The model then pulls it back, within moments, to a view consistent with
the other three and with the player's actions. We do not train for this recovery; it points
to a joint representation that stays self-consistent across views (\cref{fig:recovery}).

\subsection{Failure cases}
\label{sec:exp-failure}

The model is stable over long rollouts and follows the player's actions closely, so the
failures we describe here are rare and localized. Most stem from two limits: a context too
short to hold state across a match, and imbalances in the training data, where rare events
are under-modeled and a few very common ones over-learned.

\paragraph{The ball, left untouched, tends to move on its own.} While in play the ball
behaves correctly, taking impulses from the cars and bouncing off the walls. When it is
left untouched for a while, however, it tends to gain speed and roll toward a goal. A still ball is a rare event in the data, since play rarely stops, so the model's learned
prior favors continued motion and does not keep a resting ball still.

\paragraph{The player's car sometimes acts uncommanded.} Occasionally, it jumps or boosts
without the corresponding key being pressed: over roughly forty minutes of human play
(about $48{,}000$ frames at $20$\,fps) we counted on the order of $80$ uncommanded boosts
and $30$ jumps. The kickoff is the clearest
example: the bots open almost every match by boosting toward the ball in much the same way,
so the kickoff phase carries little behavioral diversity and the model reproduces that boost
even when the player holds back.

\begin{figure}[htbp]
    \centering
    \begin{minipage}[t]{0.315\linewidth}\centering
      \includegraphics[width=\linewidth]{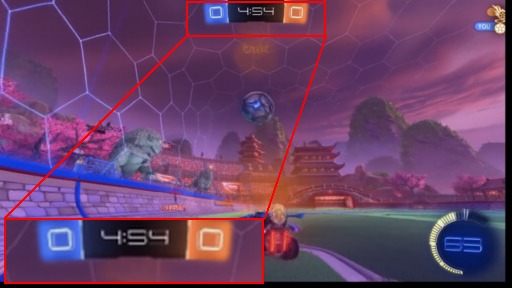}\\[1.5pt]{\scriptsize $t$}
    \end{minipage}\hfill
    \begin{minipage}[t]{0.315\linewidth}\centering
      \includegraphics[width=\linewidth]{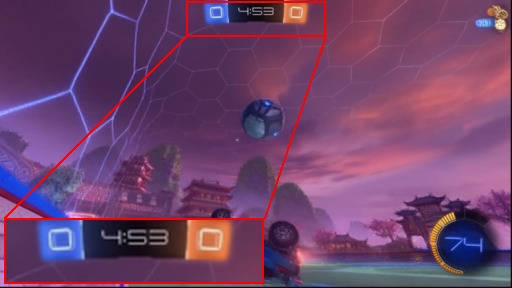}\\[1.5pt]{\scriptsize $t{+}1$\,s}
    \end{minipage}\hfill
    \begin{minipage}[t]{0.315\linewidth}\centering
      \includegraphics[width=\linewidth]{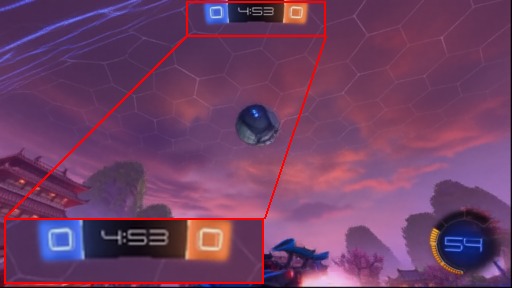}\\[1.5pt]{\scriptsize $t{+}2$\,s}
    \end{minipage}\\[6pt]
    \begin{minipage}[t]{0.315\linewidth}\centering
      \includegraphics[width=\linewidth]{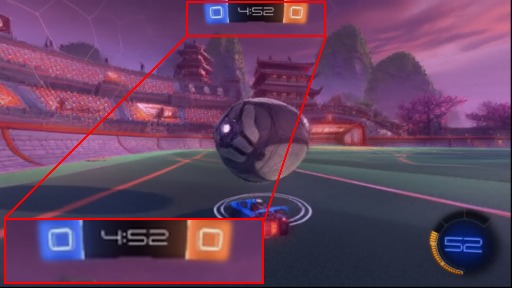}\\[1.5pt]{\scriptsize $t{+}3$\,s}
    \end{minipage}\hfill
    \begin{minipage}[t]{0.315\linewidth}\centering
      \includegraphics[width=\linewidth]{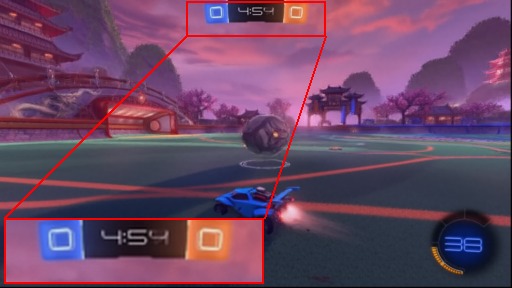}\\[1.5pt]{\scriptsize $t{+}4$\,s}
    \end{minipage}\hfill
    \begin{minipage}[t]{0.315\linewidth}\centering
      \includegraphics[width=\linewidth]{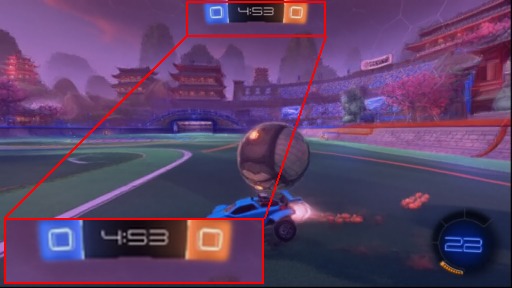}\\[1.5pt]{\scriptsize $t{+}5$\,s}
    \end{minipage}
    \caption{\textbf{The clock drifts from real time.} A generated rollout (single view),
    sampled once per second, with the score-and-clock HUD boxed in red and magnified at the lower
    left of each frame. Over five seconds of real rollout the clock should count down five seconds,
    yet it barely moves, reading $4{:}54$, $4{:}53$, $4{:}53$, $4{:}52$, $4{:}54$, $4{:}53$: it
    advances far too slowly and even ticks back up, one instance of the clock losing track.}
    \label{fig:clock}
\end{figure}

\paragraph{The clock and score lose track.} For most of a match the clock and score are
correct. They slip mainly at transitions, the score updating after a goal or the clock
crossing a milestone, each of which happens once per game and is thinly represented. The
errors stay internally consistent, though: when the model's clock reaches thirty seconds it
still triggers the on-screen ``thirty seconds remaining'' prompt, in step with its own
count even when that count is wrong (\cref{fig:clock}).

\paragraph{Goal replays diverge.} After a goal the game cuts to a scripted replay. The
replay never reproduces the goal that just happened, since the action sequence it shows is
always a different one. Occasionally the replayed ball never reaches the net, and because a
full replay outlasts the context, the rollout then stalls. In single-player the frames stay
clean and only the content is wrong, whereas in multiplayer the replay also breaks
visually.

\begin{figure}[htbp]
    \centering
    \includegraphics[width=0.245\linewidth]{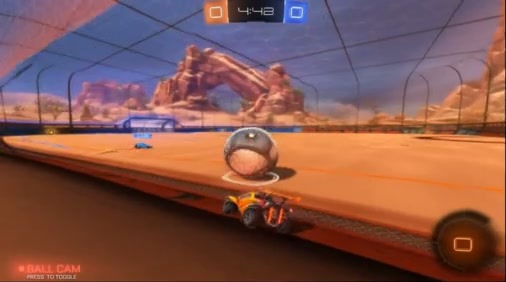}%
    \includegraphics[width=0.245\linewidth]{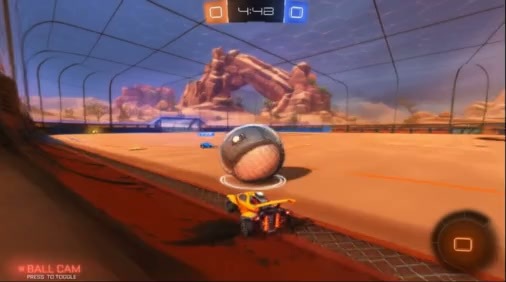}%
    \includegraphics[width=0.245\linewidth]{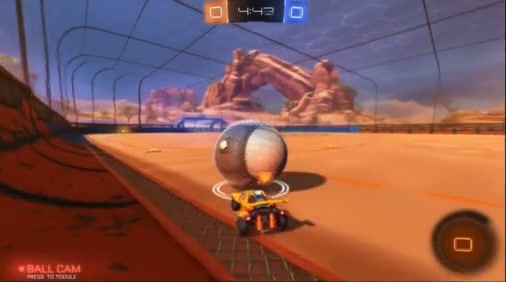}%
    \includegraphics[width=0.245\linewidth]{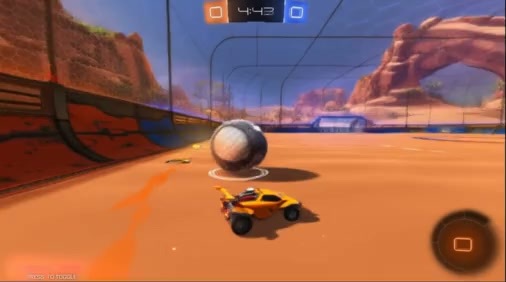}\\[3pt]
    \begin{tikzpicture}
      \draw[-{Stealth[length=2.4mm,width=2.4mm]}, line width=0.9pt, draw=black!65] (0,0) -- (\linewidth,0);
      \node[fill=white, inner xsep=4pt, font=\footnotesize\itshape, text=black!65] at (0.5\linewidth,0) {time};
    \end{tikzpicture}
    \caption{\textbf{A car merges into the ball in a single-player world model.} A rollout from a $1$B
    single-player model, shown left to right. A non-playable car (blue) sits beside the ball and is drawn
    into it as the two overlap; once the ball moves on, the car does not re-emerge as a distinct object.
    We do not observe this failure in the multiplayer model.}
    \label{fig:carmerge}
\end{figure}

\paragraph{Single-player rollouts mishandle the other cars.} A single-player model is
conditioned on one player's actions alone, so to render the other three cars it must model
their behavior even while they are off-screen, a form of theory of mind, so that they
re-enter the frame correctly. It slips in two distinct ways. A car that leaves the view is
often gone when the camera returns to it, a context-length limit: once a car stays off-screen
longer than the rolling window, the model has no trace of it left to bring back. Separately,
cars sometimes merge into the ball, a modeling error that conflates the two (\cref{fig:carmerge}). The multiplayer
model, conditioned on all four cars, largely avoids the first failure, and we do not observe the car--ball merging in its rollouts.

\FloatBarrier
\section{Conclusion}
\label{sec:conclusion}

We introduced MIRA, a real-time world model of 2v2 Rocket League for four
simultaneous players. It is a diffusion model that predicts in the
latent space of a video representation codec. Conditioned on every player's
actions and past video context, it renders each player's own view, keeps the four
views consistent, stays coherent over hour-long rollouts, and generates $20$
frames per second on a single Nvidia B200 GPU.

Three design choices carried the result. Predicting in a latent
space worked best when that latent was built on a frozen, pretrained feature
extractor: it bought nothing in reconstruction, yet it was what stopped long
rollouts from drifting and kept the four views consistent. For the objective,
diffusion forcing with few-step distillation gave long-horizon stability and
interactive speed where teacher forcing collapsed. And conditioning on all four
action streams beat the single-player model, as the extra views resolved
uncertainty about the shared state. Since visual quality hides all of this, we evaluated on action adherence,
physical understanding, and rollout stationarity alongside image fidelity and
human preference, and traced how each scales with data and model size.

Trained only to predict the next frame, the model develops behavior it was never
supervised on: it keeps
goals, demolitions, and the HUD agreeing across all four views, stays coherent
under human control unlike the bots it learned from, and recovers on its own when
a view is pushed out of distribution.

MIRA has clear limitations. We study a single game with data from one family of bot
policies, so the model inherits that style of play. Its rare failures trace mostly
to a context window too short to carry match state, which lets the clock and score
slip and off-screen cars be forgotten, and to data imbalance, which leaves rare
events such as a resting ball under-modeled. We leave to future work extending
MIRA to more varied, human-like behavior, longer memory, and real environments.

We hope MIRA is a useful step toward world models faithful enough to train and
evaluate interacting agents. To that end, we release our dataset, our full
training and inference code, and a live demo at \\ \url{https://mira-wm.com}.

\clearpage

\section*{Acknowledgments}
We thank David Picard, Nicolas Dufour, Paula Wehmeyer, and Trevor Gravely for insightful discussions and support throughout the course of this project.

\bibliographystyle{plainnat}
\bibliography{references}

\begin{thebibliography}{158}
\providecommand{\natexlab}[1]{#1}
\providecommand{\url}[1]{\texttt{#1}}
\expandafter\ifx\csname urlstyle\endcsname\relax
  \providecommand{\doi}[1]{doi: #1}\else
  \providecommand{\doi}{doi: \begingroup \urlstyle{rm}\Url}\fi

\bibitem[Abramson et~al.(2022)Abramson, Ahuja, Carnevale, Georgiev, Goldin,
  Hung, Landon, Lhotka, Lillicrap, Muldal, Powell, Santoro, Scully, Srivastava,
  von Glehn, Wayne, Wong, Yan, and Zhu]{abramson2022improving}
Josh Abramson, Arun Ahuja, Federico Carnevale, Petko Georgiev, Alex Goldin,
  Alden Hung, Jessica Landon, Jirka Lhotka, Timothy Lillicrap, Alistair Muldal,
  George Powell, Adam Santoro, Guy Scully, Sanjana Srivastava, Tamara von
  Glehn, Greg Wayne, Nathaniel Wong, Chen Yan, and Rui Zhu.
\newblock Improving multimodal interactive agents with reinforcement learning
  from human feedback, 2022.

\bibitem[Ainslie et~al.(2023)Ainslie, Lee-Thorp, de~Jong, Zemlyanskiy,
  Lebr{\'o}n, and Sanghai]{ainslie2023gqa}
Joshua Ainslie, James Lee-Thorp, Michiel de~Jong, Yury Zemlyanskiy, Federico
  Lebr{\'o}n, and Sumit Sanghai.
\newblock {GQA}: Training generalized multi-query transformer models from
  multi-head checkpoints.
\newblock In \emph{Conference on Empirical Methods in Natural Language
  Processing (EMNLP)}, 2023.

\bibitem[Alain and Bengio(2017)]{alain2017understanding}
Guillaume Alain and Yoshua Bengio.
\newblock Understanding intermediate layers using linear classifier probes.
\newblock In \emph{International Conference on Learning Representations (ICLR)
  Workshop}, 2017.

\bibitem[Albrecht et~al.(2024)Albrecht, Christianos, and
  Sch{\"a}fer]{albrecht2024marl}
Stefano~V. Albrecht, Filippos Christianos, and Lukas Sch{\"a}fer.
\newblock \emph{Multi-Agent Reinforcement Learning: Foundations and Modern
  Approaches}.
\newblock MIT Press, 2024.

\bibitem[Alonso et~al.(2024)Alonso, Jelley, Micheli, Kanervisto, Storkey,
  Pearce, and Fleuret]{alonso2024diffusion}
Eloi Alonso, Adam Jelley, Vincent Micheli, Anssi Kanervisto, Amos Storkey, Tim
  Pearce, and Fran{\c{c}}ois Fleuret.
\newblock Diffusion for world modeling: Visual details matter in {Atari}.
\newblock In \emph{Advances in Neural Information Processing Systems
  (NeurIPS)}, 2024.

\bibitem[Ansel et~al.(2024)Ansel, Yang, He, Gimelshein, Jain, Voznesensky, Bao,
  Bell, Berard, Burovski, Chauhan, Chourdia, Constable, Desmaison, DeVito,
  Ellison, Feng, Gong, Gschwind, Hirsh, Huang, Kalambarkar, Kirsch, Lazos,
  Lezcano, Liang, Liang, Lu, Luk, Maher, Pan, Puhrsch, Reso, Saroufim,
  Siraichi, Suk, Zhang, Suo, Tillet, Zhao, Wang, Zhou, Zou, Wang, Mathews, Wen,
  Chanan, Wu, and Chintala]{10.1145/3620665.3640366}
Jason Ansel, Edward Yang, Horace He, Natalia Gimelshein, Animesh Jain, Michael
  Voznesensky, Bin Bao, Peter Bell, David Berard, Evgeni Burovski, Geeta
  Chauhan, Anjali Chourdia, Will Constable, Alban Desmaison, Zachary DeVito,
  Elias Ellison, Will Feng, Jiong Gong, Michael Gschwind, Brian Hirsh, Sherlock
  Huang, Kshiteej Kalambarkar, Laurent Kirsch, Michael Lazos, Mario Lezcano,
  Yanbo Liang, Jason Liang, Yinghai Lu, C.~K. Luk, Bert Maher, Yunjie Pan,
  Christian Puhrsch, Matthias Reso, Mark Saroufim, Marcos~Yukio Siraichi, Helen
  Suk, Shunting Zhang, Michael Suo, Phil Tillet, Xu~Zhao, Eikan Wang, Keren
  Zhou, Richard Zou, Xiaodong Wang, Ajit Mathews, William Wen, Gregory Chanan,
  Peng Wu, and Soumith Chintala.
\newblock Pytorch 2: Faster machine learning through dynamic python bytecode
  transformation and graph compilation.
\newblock In \emph{Proceedings of the 29th ACM International Conference on
  Architectural Support for Programming Languages and Operating Systems, Volume
  2}, ASPLOS '24, page 929–947, New York, NY, USA, 2024. Association for
  Computing Machinery.
\newblock ISBN 9798400703850.
\newblock \doi{10.1145/3620665.3640366}.
\newblock \url{https://doi.org/10.1145/3620665.3640366}.

\bibitem[Arnab et~al.(2021)Arnab, Dehghani, Heigold, Sun, Lu{\v{c}}i{\'c}, and
  Schmid]{arnab2021vivit}
Anurag Arnab, Mostafa Dehghani, Georg Heigold, Chen Sun, Mario Lu{\v{c}}i{\'c},
  and Cordelia Schmid.
\newblock {ViViT}: A video vision transformer.
\newblock In \emph{IEEE/CVF International Conference on Computer Vision
  (ICCV)}, 2021.

\bibitem[Assran et~al.(2023)Assran, Duval, Misra, Bojanowski, Vincent, Rabbat,
  LeCun, and Ballas]{assran2023self}
Mahmoud Assran, Quentin Duval, Ishan Misra, Piotr Bojanowski, Pascal Vincent,
  Michael Rabbat, Yann LeCun, and Nicolas Ballas.
\newblock Self-supervised learning from images with a joint-embedding
  predictive architecture.
\newblock In \emph{IEEE/CVF Conference on Computer Vision and Pattern
  Recognition (CVPR)}, 2023.

\bibitem[Assran et~al.(2025)]{assran2025vjepa2}
Mahmoud Assran et~al.
\newblock {V-JEPA 2}: Self-supervised video models enable understanding,
  prediction and planning.
\newblock \emph{arXiv preprint arXiv:2506.09985}, 2025.

\bibitem[Bansal et~al.(2024)Bansal, Lin, Xie, Zong, Yarom, Bitton, Jiang, Sun,
  Chang, and Grover]{bansal2024videophy}
Hritik Bansal, Zongyu Lin, Tianyi Xie, Zeshun Zong, Michal Yarom, Yonatan
  Bitton, Chenfanfu Jiang, Yizhou Sun, Kai-Wei Chang, and Aditya Grover.
\newblock {VideoPhy}: Evaluating physical commonsense for video generation.
\newblock \emph{arXiv preprint arXiv:2406.03520}, 2024.

\bibitem[Bar et~al.(2025)Bar, Zhou, Tran, Darrell, and
  LeCun]{bar2025navigation}
Amir Bar, Gaoyue Zhou, Danny Tran, Trevor Darrell, and Yann LeCun.
\newblock Navigation world models.
\newblock In \emph{IEEE/CVF Conference on Computer Vision and Pattern
  Recognition (CVPR)}, 2025.

\bibitem[Bar-Tal et~al.(2024)Bar-Tal, Chefer, Tov, Herrmann, Paiss, Zada,
  Ephrat, Hur, Liu, Raj, et~al.]{bartal2024lumiere}
Omer Bar-Tal, Hila Chefer, Omer Tov, Charles Herrmann, Roni Paiss, Shiran Zada,
  Ariel Ephrat, Junhwa Hur, Guanghui Liu, Amit Raj, et~al.
\newblock {Lumiere}: A space-time diffusion model for video generation.
\newblock In \emph{SIGGRAPH Asia}, 2024.

\bibitem[Bardes et~al.(2024)Bardes, Garrido, Ponce, Chen, Rabbat, LeCun,
  Assran, and Ballas]{bardes2024revisiting}
Adrien Bardes, Quentin Garrido, Jean Ponce, Xinlei Chen, Michael Rabbat, Yann
  LeCun, Mahmoud Assran, and Nicolas Ballas.
\newblock Revisiting feature prediction for learning visual representations
  from video.
\newblock \emph{Transactions on Machine Learning Research (TMLR)}, 2024.
\newblock arXiv:2404.08471.

\bibitem[Bartoccioni et~al.(2025)Bartoccioni, Ramzi, Besnier,
  et~al.]{bartoccioni2025vavim}
Florent Bartoccioni, Elias Ramzi, Victor Besnier, et~al.
\newblock {VaViM} and {VaVAM}: Autonomous driving through video generative
  modeling.
\newblock \emph{arXiv preprint arXiv:2502.15672}, 2025.

\bibitem[Bengio et~al.(2015)Bengio, Vinyals, Jaitly, and
  Shazeer]{bengio2015scheduled}
Samy Bengio, Oriol Vinyals, Navdeep Jaitly, and Noam Shazeer.
\newblock Scheduled sampling for sequence prediction with recurrent neural
  networks.
\newblock In \emph{Advances in Neural Information Processing Systems
  (NeurIPS)}, 2015.

\bibitem[Bertasius et~al.(2021)Bertasius, Wang, and
  Torresani]{bertasius2021space}
Gedas Bertasius, Heng Wang, and Lorenzo Torresani.
\newblock Is space-time attention all you need for video understanding?
\newblock In \emph{International Conference on Machine Learning (ICML)}, 2021.

\bibitem[Bi et~al.(2025)Bi, Zhang, Lu, and Zheng]{bi2025vfmvae}
Tianci Bi, Xiaoyi Zhang, Yan Lu, and Nanning Zheng.
\newblock {VFM-VAE}: Vision foundation models can be good tokenizers for latent
  diffusion models.
\newblock \emph{arXiv preprint arXiv:2510.18457}, 2025.

\bibitem[Blattmann et~al.(2023)Blattmann, Dockhorn, Kulal, Mendelevitch,
  Kilian, Lorenz, Levi, English, Voleti, Letts, et~al.]{blattmann2023stable}
Andreas Blattmann, Tim Dockhorn, Sumith Kulal, Daniel Mendelevitch, Maciej
  Kilian, Dominik Lorenz, Yam Levi, Zion English, Vikram Voleti, Adam Letts,
  et~al.
\newblock Stable video diffusion: Scaling latent video diffusion models to
  large datasets.
\newblock \emph{arXiv preprint arXiv:2311.15127}, 2023.

\bibitem[Boffi et~al.(2025)Boffi, Albergo, and
  Vanden-Eijnden]{boffi2025consistency}
Nicholas~M. Boffi, Michael~S. Albergo, and Eric Vanden-Eijnden.
\newblock How to build a consistency model: Learning flow maps via
  self-distillation.
\newblock \emph{arXiv preprint arXiv:2505.18825}, 2025.

\bibitem[Braaten and {Necto contributors}(2022)]{nexto}
Rolv-Arild Braaten and {Necto contributors}.
\newblock Necto/nexto: A rocket league bot trained with deep reinforcement
  learning.
\newblock \url{https://github.com/Rolv-Arild/Necto}, 2022.

\bibitem[Brooks et~al.(2024)Brooks, Peebles, Holmes, DePue, Guo, Jing, Schnurr,
  Taylor, Luhman, Luhman, et~al.]{brooks2024video}
Tim Brooks, Bill Peebles, Connor Holmes, Will DePue, Yufei Guo, Li~Jing, David
  Schnurr, Joe Taylor, Troy Luhman, Eric Luhman, et~al.
\newblock Video generation models as world simulators, 2024.
\newblock OpenAI technical report,
  \url{https://openai.com/index/video-generation-models-as-world-simulators/}.

\bibitem[Bruce et~al.(2024)Bruce, Dennis, Edwards, Parker-Holder, Shi, Hughes,
  Lai, Mavalankar, Steigerwald, Apps, et~al.]{bruce2024genie}
Jake Bruce, Michael Dennis, Ashley Edwards, Jack Parker-Holder, Yuge Shi,
  Edward Hughes, Matthew Lai, Aditi Mavalankar, Richie Steigerwald, Chris Apps,
  et~al.
\newblock {Genie}: Generative interactive environments.
\newblock In \emph{International Conference on Machine Learning (ICML)}, 2024.

\bibitem[Caron et~al.(2021)Caron, Touvron, Misra, J{\'e}gou, Mairal,
  Bojanowski, and Joulin]{caron2021emerging}
Mathilde Caron, Hugo Touvron, Ishan Misra, Herv{\'e} J{\'e}gou, Julien Mairal,
  Piotr Bojanowski, and Armand Joulin.
\newblock Emerging properties in self-supervised vision transformers.
\newblock In \emph{IEEE/CVF International Conference on Computer Vision
  (ICCV)}, 2021.

\bibitem[Che et~al.(2025)Che, He, Liu, Jin, and Chen]{che2025gamegenx}
Haoxuan Che, Xuanhua He, Quande Liu, Cheng Jin, and Hao Chen.
\newblock {GameGen-X}: Interactive open-world game video generation.
\newblock In \emph{International Conference on Learning Representations
  (ICLR)}, 2025.

\bibitem[Chen et~al.(2024)Chen, Mart{\'\i}~Mons{\'o}, Du, Simchowitz, Tedrake,
  and Sitzmann]{chen2024diffusion}
Boyuan Chen, Diego Mart{\'\i}~Mons{\'o}, Yilun Du, Max Simchowitz, Russ
  Tedrake, and Vincent Sitzmann.
\newblock Diffusion forcing: Next-token prediction meets full-sequence
  diffusion.
\newblock In \emph{Advances in Neural Information Processing Systems
  (NeurIPS)}, 2024.

\bibitem[Chen et~al.(2025{\natexlab{a}})Chen, Han, Chen, Li, Wang, Wang, Wang,
  Liu, Zou, and Raj]{chen2025maetok}
Hao Chen, Yujin Han, Fangyi Chen, Xiang Li, Yidong Wang, Jindong Wang, Ze~Wang,
  Zicheng Liu, Difan Zou, and Bhiksha Raj.
\newblock Masked autoencoders are effective tokenizers for diffusion models.
\newblock In \emph{International Conference on Machine Learning (ICML)},
  2025{\natexlab{a}}.

\bibitem[Chen et~al.(2025{\natexlab{b}})Chen, Zou, He, Chen, Xie, Han, and
  Cai]{chen2025dcae15}
Junyu Chen, Dongyun Zou, Wenkun He, Junsong Chen, Enze Xie, Song Han, and Han
  Cai.
\newblock {DC-AE 1.5}: Accelerating diffusion model convergence with structured
  latent space.
\newblock In \emph{IEEE/CVF International Conference on Computer Vision
  (ICCV)}, 2025{\natexlab{b}}.

\bibitem[Chen et~al.(2022)Chen, Wang, Chen, Wu, Liu, Chen, Li, Kanda, Yoshioka,
  Xiao, et~al.]{chen2022wavlm}
Sanyuan Chen, Chengyi Wang, Zhengyang Chen, Yu~Wu, Shujie Liu, Zhuo Chen, Jinyu
  Li, Naoyuki Kanda, Takuya Yoshioka, Xiong Xiao, et~al.
\newblock {WavLM}: Large-scale self-supervised pre-training for full stack
  speech processing.
\newblock \emph{IEEE Journal of Selected Topics in Signal Processing}, 2022.

\bibitem[Chetlur et~al.(2014)Chetlur, Woolley, Vandermersch, Cohen, Tran,
  Catanzaro, and Shelhamer]{DBLP:journals/corr/ChetlurWVCTCS14}
Sharan Chetlur, Cliff Woolley, Philippe Vandermersch, Jonathan Cohen, John
  Tran, Bryan Catanzaro, and Evan Shelhamer.
\newblock cudnn: Efficient primitives for deep learning.
\newblock \emph{CoRR}, abs/1410.0759, 2014.
\newblock \url{http://arxiv.org/abs/1410.0759}.

\bibitem[Darcet et~al.(2024)Darcet, Oquab, Mairal, and
  Bojanowski]{darcet2024vision}
Timoth{\'e}e Darcet, Maxime Oquab, Julien Mairal, and Piotr Bojanowski.
\newblock Vision transformers need registers.
\newblock In \emph{International Conference on Learning Representations
  (ICLR)}, 2024.

\bibitem[{Decart and Etched}(2024)]{decart2024oasis}
{Decart and Etched}.
\newblock {Oasis}: A universe in a transformer, 2024.
\newblock \url{https://oasis-model.github.io/}.

\bibitem[D{\'e}fossez et~al.(2024)D{\'e}fossez, Mazar{\'e}, Orsini, Royer,
  P{\'e}rez, J{\'e}gou, Grave, and Zeghidour]{defossez2024moshi}
Alexandre D{\'e}fossez, Laurent Mazar{\'e}, Manu Orsini, Am{\'e}lie Royer,
  Patrick P{\'e}rez, Herv{\'e} J{\'e}gou, Edouard Grave, and Neil Zeghidour.
\newblock {Moshi}: a speech-text foundation model for real-time dialogue.
\newblock \emph{arXiv preprint arXiv:2410.00037}, 2024.

\bibitem[Dehghani et~al.(2023)Dehghani, Djolonga, Mustafa, Padlewski, Heek,
  et~al.]{dehghani2023scaling}
Mostafa Dehghani, Josip Djolonga, Basil Mustafa, Piotr Padlewski, Jonathan
  Heek, et~al.
\newblock Scaling vision transformers to 22 billion parameters.
\newblock In \emph{International Conference on Machine Learning (ICML)}, 2023.

\bibitem[Dosovitskiy et~al.(2021)Dosovitskiy, Beyer, Kolesnikov, Weissenborn,
  Zhai, Unterthiner, Dehghani, Minderer, Heigold, Gelly, Uszkoreit, and
  Houlsby]{dosovitskiy2021image}
Alexey Dosovitskiy, Lucas Beyer, Alexander Kolesnikov, Dirk Weissenborn,
  Xiaohua Zhai, Thomas Unterthiner, Mostafa Dehghani, Matthias Minderer, Georg
  Heigold, Sylvain Gelly, Jakob Uszkoreit, and Neil Houlsby.
\newblock An image is worth 16x16 words: Transformers for image recognition at
  scale.
\newblock In \emph{International Conference on Learning Representations
  (ICLR)}, 2021.

\bibitem[Duan et~al.(2025)Duan, Guo, Zhao, Wu, et~al.]{duan2025worldscore}
Haoyi Duan, Hong-Xing Guo, Xiaoshuai Zhao, Jiajun Wu, et~al.
\newblock {WorldScore}: A unified evaluation benchmark for world generation.
\newblock In \emph{IEEE/CVF International Conference on Computer Vision
  (ICCV)}, 2025.

\bibitem[Esser et~al.(2021)Esser, Rombach, and Ommer]{esser2021taming}
Patrick Esser, Robin Rombach, and Bj{\"o}rn Ommer.
\newblock Taming transformers for high-resolution image synthesis.
\newblock In \emph{CVPR}, 2021.

\bibitem[Feng et~al.(2024)Feng, Zhang, Yang, Xiao, Shu, Liu, Zheng, Huang, Liu,
  and Zhang]{feng2024matrix}
Ruili Feng, Han Zhang, Zhantao Yang, Jie Xiao, Zhilei Shu, Zhiheng Liu, Andy
  Zheng, Yukun Huang, Yu~Liu, and Hongyang Zhang.
\newblock The {Matrix}: Infinite-horizon world generation with real-time moving
  control.
\newblock \emph{arXiv preprint arXiv:2412.03568}, 2024.

\bibitem[Frans et~al.(2025)Frans, Hafner, Levine, and Abbeel]{frans2025one}
Kevin Frans, Danijar Hafner, Sergey Levine, and Pieter Abbeel.
\newblock One step diffusion via shortcut models.
\newblock In \emph{International Conference on Learning Representations
  (ICLR)}, 2025.

\bibitem[Gao et~al.(2024)Gao, Yang, Chen, Chitta, Qiu, Geiger, Zhang, and
  Li]{gao2024vista}
Shenyuan Gao, Jiazhi Yang, Li~Chen, Kashyap Chitta, Yihang Qiu, Andreas Geiger,
  Jun Zhang, and Hongyang Li.
\newblock {Vista}: A generalizable driving world model with high fidelity and
  versatile controllability.
\newblock In \emph{Advances in Neural Information Processing Systems
  (NeurIPS)}, 2024.

\bibitem[Gao et~al.(2025)Gao, Chen, Chen, and Gu]{gao2025fae}
Yuan Gao, Chen Chen, Tianrong Chen, and Jiatao Gu.
\newblock One layer is enough: Adapting pretrained visual encoders for image
  generation.
\newblock \emph{arXiv preprint arXiv:2512.07829}, 2025.

\bibitem[{Gemini Robotics Team}(2025)]{geminirobotics2025veo}
{Gemini Robotics Team}.
\newblock Evaluating {Gemini} robotics policies in a {Veo} world simulator.
\newblock 2025.

\bibitem[{Google DeepMind}(2024)]{parkerholder2024genie2}
{Google DeepMind}.
\newblock {Genie 2}: A large-scale foundation world model, 2024.
\newblock
  \url{https://deepmind.google/blog/genie-2-a-large-scale-foundation-world-model/}.

\bibitem[{Google DeepMind}(2025)]{deepmind2025genie3}
{Google DeepMind}.
\newblock {Genie 3}: A new frontier for world models, 2025.
\newblock
  \url{https://deepmind.google/blog/genie-3-a-new-frontier-for-world-models/}.

\bibitem[Gui et~al.(2025)Gui, Schusterbauer, Phan, Krause, Susskind, Bautista,
  and Ommer]{gui2025reptok}
Ming Gui, Johannes Schusterbauer, Timy Phan, Felix Krause, Josh Susskind,
  Miguel~Angel Bautista, and Bj{\"o}rn Ommer.
\newblock Adapting self-supervised representations as a latent space for
  efficient generation.
\newblock \emph{arXiv preprint arXiv:2510.14630}, 2025.

\bibitem[Guo et~al.(2025)Guo, Ye, He, Wu, Jiang, Pearce, and
  Bian]{guo2025mineworld}
Junliang Guo, Yang Ye, Tianyu He, Haoyu Wu, Yushu Jiang, Tim Pearce, and Jiang
  Bian.
\newblock {MineWorld}: A real-time and open-source interactive world model on
  {Minecraft}.
\newblock \emph{arXiv preprint arXiv:2504.08388}, 2025.

\bibitem[Gupta et~al.(2024)Gupta, Yu, Sohn, Gu, Hahn, Fei-Fei, Essa, Jiang, and
  Lezama]{gupta2024photorealistic}
Agrim Gupta, Lijun Yu, Kihyuk Sohn, Xiuye Gu, Meera Hahn, Li~Fei-Fei, Irfan
  Essa, Lu~Jiang, and Jos{\'e} Lezama.
\newblock Photorealistic video generation with diffusion models.
\newblock In \emph{European Conference on Computer Vision (ECCV)}, 2024.

\bibitem[Ha and Schmidhuber(2018)]{ha2018world}
David Ha and J{\"u}rgen Schmidhuber.
\newblock Recurrent world models facilitate policy evolution.
\newblock In \emph{Advances in Neural Information Processing Systems
  (NeurIPS)}, 2018.
\newblock arXiv:1803.10122.

\bibitem[HaCohen et~al.(2025)]{hacohen2025ltxvideo}
Yoav HaCohen et~al.
\newblock {LTX-Video}: Realtime video latent diffusion.
\newblock \emph{arXiv preprint arXiv:2501.00103}, 2025.

\bibitem[HaCohen et~al.(2026)]{hacohen2026ltx2}
Yoav HaCohen et~al.
\newblock {LTX-2}: Efficient joint audio-visual foundation model.
\newblock \emph{arXiv preprint arXiv:2601.03233}, 2026.

\bibitem[Hafner et~al.(2019)Hafner, Lillicrap, Fischer, Villegas, Ha, Lee, and
  Davidson]{hafner2019planet}
Danijar Hafner, Timothy Lillicrap, Ian Fischer, Ruben Villegas, David Ha,
  Honglak Lee, and James Davidson.
\newblock Learning latent dynamics for planning from pixels.
\newblock In \emph{International Conference on Machine Learning (ICML)}, 2019.

\bibitem[Hafner et~al.(2020)Hafner, Lillicrap, Ba, and
  Norouzi]{hafner2020dream}
Danijar Hafner, Timothy Lillicrap, Jimmy Ba, and Mohammad Norouzi.
\newblock Dream to control: Learning behaviors by latent imagination.
\newblock In \emph{International Conference on Learning Representations
  (ICLR)}, 2020.

\bibitem[Hafner et~al.(2021)Hafner, Lillicrap, Norouzi, and
  Ba]{hafner2021mastering}
Danijar Hafner, Timothy Lillicrap, Mohammad Norouzi, and Jimmy Ba.
\newblock Mastering atari with discrete world models.
\newblock In \emph{International Conference on Learning Representations
  (ICLR)}, 2021.

\bibitem[Hafner et~al.(2025{\natexlab{a}})Hafner, Pasukonis, Ba, and
  Lillicrap]{hafner2025mastering}
Danijar Hafner, Jurgis Pasukonis, Jimmy Ba, and Timothy Lillicrap.
\newblock Mastering diverse domains through world models.
\newblock \emph{Nature}, 2025{\natexlab{a}}.
\newblock arXiv:2301.04104.

\bibitem[Hafner et~al.(2025{\natexlab{b}})Hafner, Pasukonis, Ba, and
  Lillicrap]{hafner2025nature}
Danijar Hafner, Jurgis Pasukonis, Jimmy Ba, and Timothy Lillicrap.
\newblock Mastering diverse control tasks through world models.
\newblock \emph{Nature}, 2025{\natexlab{b}}.

\bibitem[Hafner et~al.(2025{\natexlab{c}})Hafner, Yan, and
  Lillicrap]{hafner2025dreamerv4}
Danijar Hafner, Wilson Yan, and Timothy Lillicrap.
\newblock Training agents inside of scalable world models.
\newblock \emph{arXiv preprint arXiv:2509.24527}, 2025{\natexlab{c}}.

\bibitem[Hansen et~al.(2022)Hansen, Wang, and Su]{hansen2022temporal}
Nicklas Hansen, Xiaolong Wang, and Hao Su.
\newblock Temporal difference learning for model predictive control.
\newblock In \emph{International Conference on Machine Learning (ICML)}, 2022.

\bibitem[Hansen et~al.(2024)Hansen, Su, and Wang]{hansen2024tdmpc2}
Nicklas Hansen, Hao Su, and Xiaolong Wang.
\newblock {TD-MPC2}: Scalable, robust world models for continuous control.
\newblock In \emph{International Conference on Learning Representations
  (ICLR)}, 2024.

\bibitem[Hansen-Estruch et~al.(2026)Hansen-Estruch, Chen, Ramanujan, Zohar,
  Ping, Sinha, Georgopoulos, Schoenfeld, Hou, Juefei-Xu, Vishwanath, and
  Thabet]{hansen2026vitokv2}
Philippe Hansen-Estruch, Jiahui Chen, Vivek Ramanujan, Orr Zohar, Yan Ping,
  Animesh Sinha, Markos Georgopoulos, Edgar Schoenfeld, Ji~Hou, Felix
  Juefei-Xu, Sriram Vishwanath, and Ali Thabet.
\newblock {ViTok-v2}: Scaling native resolution auto-encoders to 5 billion
  parameters.
\newblock \emph{arXiv preprint arXiv:2605.05331}, 2026.

\bibitem[He et~al.(2024)He, Xu, Guo, Wetzstein, Dai, Li, and
  Yang]{he2024cameractrl}
Hao He, Yinghao Xu, Yuwei Guo, Gordon Wetzstein, Bo~Dai, Hongsheng Li, and
  Ceyuan Yang.
\newblock {CameraCtrl}: Enabling camera control for text-to-video generation.
\newblock \emph{arXiv preprint arXiv:2404.02101}, 2024.

\bibitem[He et~al.(2022)He, Chen, Xie, Li, Doll{\'a}r, and
  Girshick]{he2022masked}
Kaiming He, Xinlei Chen, Saining Xie, Yanghao Li, Piotr Doll{\'a}r, and Ross
  Girshick.
\newblock Masked autoencoders are scalable vision learners.
\newblock In \emph{IEEE/CVF Conference on Computer Vision and Pattern
  Recognition (CVPR)}, 2022.

\bibitem[He et~al.(2025)He, Zhang, Wu, Zhang, Xu, et~al.]{he2025matrixgame2}
Xianglong He, Chunli Zhang, Dongdong Wu, Yifan Zhang, Yiqun Xu, et~al.
\newblock {Matrix-Game 2.0}: An open-source, real-time, and streaming
  interactive world model.
\newblock \emph{arXiv preprint arXiv:2508.13009}, 2025.

\bibitem[Henry et~al.(2020)Henry, Dachapally, Pawar, and Chen]{henry2020query}
Alex Henry, Prudhvi~Raj Dachapally, Shubham Pawar, and Yuxuan Chen.
\newblock Query-key normalization for transformers.
\newblock In \emph{Findings of the Association for Computational Linguistics:
  EMNLP}, 2020.

\bibitem[Heusel et~al.(2017)Heusel, Ramsauer, Unterthiner, Nessler, and
  Hochreiter]{heusel2017gans}
Martin Heusel, Hubert Ramsauer, Thomas Unterthiner, Bernhard Nessler, and Sepp
  Hochreiter.
\newblock {GANs} trained by a two time-scale update rule converge to a local
  {Nash} equilibrium.
\newblock In \emph{Advances in Neural Information Processing Systems
  (NeurIPS)}, 2017.

\bibitem[Ho et~al.(2020)Ho, Jain, and Abbeel]{ho2020denoising}
Jonathan Ho, Ajay Jain, and Pieter Abbeel.
\newblock Denoising diffusion probabilistic models.
\newblock In \emph{Advances in Neural Information Processing Systems
  (NeurIPS)}, 2020.

\bibitem[Ho et~al.(2022{\natexlab{a}})Ho, Chan, Saharia, Whang, Gao, Gritsenko,
  Kingma, Poole, Norouzi, Fleet, and Salimans]{ho2022imagen}
Jonathan Ho, William Chan, Chitwan Saharia, Jay Whang, Ruiqi Gao, Alexey
  Gritsenko, Diederik~P. Kingma, Ben Poole, Mohammad Norouzi, David~J. Fleet,
  and Tim Salimans.
\newblock {Imagen Video}: High definition video generation with diffusion
  models.
\newblock \emph{arXiv preprint arXiv:2210.02303}, 2022{\natexlab{a}}.

\bibitem[Ho et~al.(2022{\natexlab{b}})Ho, Salimans, Gritsenko, Chan, Norouzi,
  and Fleet]{ho2022video}
Jonathan Ho, Tim Salimans, Alexey Gritsenko, William Chan, Mohammad Norouzi,
  and David~J. Fleet.
\newblock Video diffusion models.
\newblock In \emph{Advances in Neural Information Processing Systems
  (NeurIPS)}, 2022{\natexlab{b}}.

\bibitem[Hu et~al.(2022)Hu, Corrado, Griffiths, Murez, Gurau, Yeo, Kendall,
  Cipolla, and Shotton]{hu2022mile}
Anthony Hu, Gianluca Corrado, Nicolas Griffiths, Zak Murez, Corina Gurau,
  Hudson Yeo, Alex Kendall, Roberto Cipolla, and Jamie Shotton.
\newblock Model-based imitation learning for urban driving.
\newblock In \emph{Advances in Neural Information Processing Systems
  (NeurIPS)}, 2022.

\bibitem[Hu et~al.(2023)Hu, Russell, Yeo, Murez, Fedoseev, Kendall, Shotton,
  and Corrado]{hu2023gaia}
Anthony Hu, Lloyd Russell, Hudson Yeo, Zak Murez, George Fedoseev, Alex
  Kendall, Jamie Shotton, and Gianluca Corrado.
\newblock {GAIA-1}: A generative world model for autonomous driving.
\newblock \emph{arXiv preprint arXiv:2309.17080}, 2023.

\bibitem[Hu et~al.(2026)]{hu2026metaworld}
Teng Hu et~al.
\newblock {MetaWorld}: Scaling multi-agent video world model from single-view
  video data.
\newblock \emph{arXiv preprint arXiv:2606.02753}, 2026.

\bibitem[Huang et~al.(2025)Huang, Li, He, Zhou, and Shechtman]{huang2025self}
Xun Huang, Zhengqi Li, Guande He, Mingyuan Zhou, and Eli Shechtman.
\newblock Self forcing: Bridging the train-test gap in autoregressive video
  diffusion.
\newblock In \emph{Advances in Neural Information Processing Systems
  (NeurIPS)}, 2025.

\bibitem[Huang et~al.(2024)Huang, He, Yu, Zhang, Si, Jiang, Zhang, Wu, Jin,
  Chanpaisit, et~al.]{huang2024vbench}
Ziqi Huang, Yinan He, Jiashuo Yu, Fan Zhang, Chenyang Si, Yuming Jiang, Yuanhan
  Zhang, Tianxing Wu, Qingyang Jin, Nattapol Chanpaisit, et~al.
\newblock {VBench}: Comprehensive benchmark suite for video generative models.
\newblock In \emph{IEEE/CVF Conference on Computer Vision and Pattern
  Recognition (CVPR)}, 2024.

\bibitem[Kanervisto et~al.(2025)Kanervisto, Bignell, Wen, Grayson, Georgescu,
  Valcarcel~Macua, Tan, Rashid, Pearce, Cant{\'o}n~Ferrer,
  et~al.]{kanervisto2025world}
Anssi Kanervisto, Dave Bignell, Linda~Yilin Wen, Martin Grayson, Raluca
  Georgescu, Sergio Valcarcel~Macua, Shan~Zheng Tan, Tabish Rashid, Tim Pearce,
  Cristian Cant{\'o}n~Ferrer, et~al.
\newblock World and human action models towards gameplay ideation.
\newblock \emph{Nature}, 638:\penalty0 656--663, 2025.

\bibitem[Kim et~al.(2024)Kim, Kang, Choi, and Han]{kim2024fifo}
Jihwan Kim, Junoh Kang, Jinyoung Choi, and Bohyung Han.
\newblock {FIFO-Diffusion}: Generating infinite videos from text without
  training.
\newblock In \emph{Advances in Neural Information Processing Systems
  (NeurIPS)}, 2024.

\bibitem[Kim et~al.(2020)Kim, Zhou, Philion, Torralba, and
  Fidler]{kim2020learning}
Seung~Wook Kim, Yuhao Zhou, Jonah Philion, Antonio Torralba, and Sanja Fidler.
\newblock Learning to simulate dynamic environments with {GameGAN}.
\newblock In \emph{IEEE/CVF Conference on Computer Vision and Pattern
  Recognition (CVPR)}, 2020.

\bibitem[Koh et~al.(2021)Koh, Lee, Yang, Baldridge, and
  Anderson]{koh2021pathdreamer}
Jing~Yu Koh, Honglak Lee, Yinfei Yang, Jason Baldridge, and Peter Anderson.
\newblock {Pathdreamer}: A world model for indoor navigation.
\newblock In \emph{IEEE/CVF International Conference on Computer Vision
  (ICCV)}, 2021.

\bibitem[Kondratyuk et~al.(2024)Kondratyuk, Yu, Gu, Lezama, Huang, Schindler,
  Hornung, Birodkar, Yan, Chiu, et~al.]{kondratyuk2024videopoet}
Dan Kondratyuk, Lijun Yu, Xiuye Gu, Jos{\'e} Lezama, Jonathan Huang, Grant
  Schindler, Rachel Hornung, Vighnesh Birodkar, Jimmy Yan, Ming-Chang Chiu,
  et~al.
\newblock {VideoPoet}: A large language model for zero-shot video generation.
\newblock In \emph{International Conference on Machine Learning (ICML)}, 2024.

\bibitem[Kong et~al.(2024)]{kong2024hunyuanvideo}
Weijie Kong et~al.
\newblock {HunyuanVideo}: A systematic framework for large video generative
  models.
\newblock \emph{arXiv preprint arXiv:2412.03603}, 2024.

\bibitem[Kouzelis et~al.(2025)Kouzelis, Kakogeorgiou, Gidaris, and
  Komodakis]{kouzelis2025eqvae}
Theodoros Kouzelis, Ioannis Kakogeorgiou, Spyros Gidaris, and Nikos Komodakis.
\newblock {EQ-VAE}: Equivariance regularized latent space for improved
  generative image modeling.
\newblock In \emph{International Conference on Machine Learning (ICML)}, 2025.

\bibitem[Lamb et~al.(2016)Lamb, Goyal, Zhang, Zhang, Courville, and
  Bengio]{lamb2016professor}
Alex Lamb, Anirudh Goyal, Ying Zhang, Saizheng Zhang, Aaron Courville, and
  Yoshua Bengio.
\newblock Professor forcing: A new algorithm for training recurrent networks.
\newblock In \emph{Advances in Neural Information Processing Systems
  (NeurIPS)}, 2016.

\bibitem[LeCun(2022)]{lecun2022path}
Yann LeCun.
\newblock A path towards autonomous machine intelligence, 2022.
\newblock Position paper, version 0.9.2, OpenReview.

\bibitem[Li et~al.(2025{\natexlab{a}})Li, Zhang, Lin, Xie,
  et~al.]{li2025worldmodelbench}
Dacheng Li, Yunhao Zhang, Ji~Lin, Enze Xie, et~al.
\newblock {WorldModelBench}: Judging video generation models as world models.
\newblock \emph{arXiv preprint arXiv:2502.20694}, 2025{\natexlab{a}}.

\bibitem[Li et~al.(2025{\natexlab{b}})Li, Zhang, Jiang, Wang, Zhao,
  et~al.]{li2025hunyuangamecraft}
Jiaqi Li, Junshu Zhang, Boyuan Jiang, Yuxuan Wang, Yujie Zhao, et~al.
\newblock {Hunyuan-GameCraft}: High-dynamic interactive game video generation
  with hybrid history condition.
\newblock \emph{arXiv preprint arXiv:2506.17201}, 2025{\natexlab{b}}.

\bibitem[Li et~al.(2023)Li, Hopkins, Bau, Vi{\'e}gas, Pfister, and
  Wattenberg]{li2023emergent}
Kenneth Li, Aspen~K. Hopkins, David Bau, Fernanda Vi{\'e}gas, Hanspeter
  Pfister, and Martin Wattenberg.
\newblock Emergent world representations: Exploring a sequence model trained on
  a synthetic task.
\newblock In \emph{International Conference on Learning Representations
  (ICLR)}, 2023.

\bibitem[Li and He(2025)]{li2025jit}
Tianhong Li and Kaiming He.
\newblock Back to basics: Let denoising generative models denoise.
\newblock \emph{arXiv preprint arXiv:2511.13720}, 2025.

\bibitem[Li et~al.(2022)Li, Wang, Snavely, and
  Kanazawa]{li2022infinitenaturezero}
Zhengqi Li, Qianqian Wang, Noah Snavely, and Angjoo Kanazawa.
\newblock {InfiniteNature-Zero}: Learning perpetual view generation of natural
  scenes from single images.
\newblock In \emph{European Conference on Computer Vision (ECCV)}, 2022.

\bibitem[Lipman et~al.(2023)Lipman, Chen, Ben-Hamu, Nickel, and
  Le]{lipman2023flow}
Yaron Lipman, Ricky T.~Q. Chen, Heli Ben-Hamu, Maximilian Nickel, and Matt Le.
\newblock Flow matching for generative modeling.
\newblock In \emph{International Conference on Learning Representations
  (ICLR)}, 2023.

\bibitem[Liu et~al.(2021)Liu, Tucker, Jampani, Makadia, Snavely, and
  Kanazawa]{liu2021infinite}
Andrew Liu, Richard Tucker, Varun Jampani, Ameesh Makadia, Noah Snavely, and
  Angjoo Kanazawa.
\newblock {Infinite Nature}: Perpetual view generation of natural scenes from a
  single image.
\newblock In \emph{IEEE/CVF International Conference on Computer Vision
  (ICCV)}, 2021.

\bibitem[Liu et~al.(2026)Liu, He, Ren, et~al.]{liu2026gammaworld}
Fangfu Liu, Kai He, Xuanchi Ren, et~al.
\newblock {$\gamma$-World}: Generative multi-agent world modeling beyond two
  players.
\newblock \emph{arXiv preprint arXiv:2605.28816}, 2026.

\bibitem[Liu et~al.(2025)Liu, Li, Zhao, Liu, Lu, et~al.]{liu2025rolling}
Kunhao Liu, Wenbo Li, Jiale Zhao, Ziwei Liu, Shijian Lu, et~al.
\newblock Rolling forcing: Autoregressive long video diffusion in real time.
\newblock \emph{arXiv preprint arXiv:2509.25161}, 2025.

\bibitem[Liu et~al.(2023{\natexlab{a}})Liu, Wu, Van~Hoorick, Tokmakov,
  Zakharov, and Vondrick]{liu2023zero1to3}
Ruoshi Liu, Rundi Wu, Basile Van~Hoorick, Pavel Tokmakov, Sergey Zakharov, and
  Carl Vondrick.
\newblock {Zero-1-to-3}: Zero-shot one image to 3d object.
\newblock In \emph{IEEE/CVF International Conference on Computer Vision
  (ICCV)}, 2023{\natexlab{a}}.

\bibitem[Liu et~al.(2023{\natexlab{b}})Liu, Gong, and Liu]{liu2023rectified}
Xingchao Liu, Chengyue Gong, and Qiang Liu.
\newblock Flow straight and fast: Learning to generate and transfer data with
  rectified flow.
\newblock In \emph{International Conference on Learning Representations
  (ICLR)}, 2023{\natexlab{b}}.

\bibitem[Luo et~al.(2023)Luo, Tan, Huang, Li, and Li]{luo2023latent}
Simian Luo, Yiqin Tan, Longbo Huang, Jian Li, and Hongsheng Li.
\newblock Latent consistency models: Synthesizing high-resolution images with
  few-step inference.
\newblock \emph{arXiv preprint arXiv:2310.04378}, 2023.

\bibitem[Ma et~al.(2024)Ma, Goldstein, Albergo, Boffi, Vanden-Eijnden, and
  Xie]{ma2024sit}
Nanye Ma, Mark Goldstein, Michael~S. Albergo, Nicholas~M. Boffi, Eric
  Vanden-Eijnden, and Saining Xie.
\newblock {SiT}: Exploring flow and diffusion-based generative models with
  scalable interpolant transformers.
\newblock In \emph{European Conference on Computer Vision (ECCV)}, 2024.

\bibitem[Ma et~al.(2026)]{ma2026pixelgen}
Zehong Ma et~al.
\newblock {PixelGen}: Improving pixel diffusion with perceptual supervision.
\newblock \emph{arXiv preprint arXiv:2602.02493}, 2026.

\bibitem[Maes et~al.(2026)Maes, Le~Lidec, Scieur, LeCun, and
  Balestriero]{maes2026leworldmodel}
Lucas Maes, Quentin Le~Lidec, Damien Scieur, Yann LeCun, and Randall
  Balestriero.
\newblock {LeWorldModel}: Stable end-to-end joint-embedding predictive
  architecture from pixels.
\newblock \emph{arXiv preprint arXiv:2603.19312}, 2026.

\bibitem[{Medal.tv}(2024)]{medaltv}
{Medal.tv}.
\newblock {Medal}: A gameplay clip capture and sharing platform.
\newblock \url{https://medal.tv}, 2024.

\bibitem[Menapace et~al.(2021)Menapace, Lathuili{\`e}re, Tulyakov, Siarohin,
  and Ricci]{menapace2021playable}
Willi Menapace, St{\'e}phane Lathuili{\`e}re, Sergey Tulyakov, Aliaksandr
  Siarohin, and Elisa Ricci.
\newblock Playable video generation.
\newblock In \emph{IEEE/CVF Conference on Computer Vision and Pattern
  Recognition (CVPR)}, 2021.

\bibitem[Meng et~al.(2025)Meng, Liao, Tan, Shao, Lu, Zhang, Cheng, Li, Qiao,
  and Luo]{meng2025phygenbench}
Fanqing Meng, Jiaqi Liao, Xinyu Tan, Wenqi Shao, Quanfeng Lu, Kaipeng Zhang,
  Yu~Cheng, Dianqi Li, Yu~Qiao, and Ping Luo.
\newblock Towards world simulator: Crafting physical commonsense-based
  benchmark for video generation.
\newblock In \emph{International Conference on Machine Learning (ICML)}, 2025.

\bibitem[{Meta AI}(2026)]{eupe2026}
{Meta AI}.
\newblock {EUPE}: Efficient universal perception encoder.
\newblock \emph{arXiv preprint arXiv:2603.22387}, 2026.

\bibitem[Micheli et~al.(2023)Micheli, Alonso, and
  Fleuret]{micheli2023transformers}
Vincent Micheli, Eloi Alonso, and Fran{\c{c}}ois Fleuret.
\newblock Transformers are sample-efficient world models.
\newblock In \emph{International Conference on Learning Representations
  (ICLR)}, 2023.

\bibitem[Micheli et~al.(2024)Micheli, Alonso, and
  Fleuret]{micheli2024efficient}
Vincent Micheli, Eloi Alonso, and Fran{\c{c}}ois Fleuret.
\newblock Efficient world models with context-aware tokenization.
\newblock In \emph{International Conference on Machine Learning (ICML)}, 2024.

\bibitem[Motamed et~al.(2025)Motamed, Culp, Swersky, Jaini, and
  Geirhos]{motamed2025physicsiq}
Saman Motamed, Laura Culp, Kevin Swersky, Priyank Jaini, and Robert Geirhos.
\newblock Do generative video models understand physical principles?
\newblock \emph{arXiv preprint arXiv:2501.09038}, 2025.

\bibitem[Mulder and {BakkesMod contributors}(2016)]{bakkesmod}
Chris Mulder and {BakkesMod contributors}.
\newblock {BakkesMod}: A rocket league modding framework.
\newblock \url{https://bakkesmod.com}, 2016.

\bibitem[Oquab et~al.(2024)Oquab, Darcet, Moutakanni, Vo, Szafraniec, Khalidov,
  Fernandez, Haziza, Massa, El-Nouby, et~al.]{oquab2024dinov2}
Maxime Oquab, Timoth{\'e}e Darcet, Th{\'e}o Moutakanni, Huy Vo, Marc
  Szafraniec, Vasil Khalidov, Pierre Fernandez, Daniel Haziza, Francisco Massa,
  Alaaeldin El-Nouby, et~al.
\newblock {DINOv2}: Learning robust visual features without supervision.
\newblock \emph{Transactions on Machine Learning Research (TMLR)}, 2024.

\bibitem[Peebles and Xie(2023)]{peebles2023scalable}
William Peebles and Saining Xie.
\newblock Scalable diffusion models with transformers.
\newblock In \emph{IEEE/CVF International Conference on Computer Vision
  (ICCV)}, 2023.

\bibitem[Peng et~al.(2022)Peng, Dong, Bao, Ye, and Wei]{peng2022beit2}
Zhiliang Peng, Li~Dong, Hangbo Bao, Qixiang Ye, and Furu Wei.
\newblock {BEiT v2}: Masked image modeling with vector-quantized visual
  tokenizers.
\newblock \emph{arXiv preprint arXiv:2208.06366}, 2022.

\bibitem[Po et~al.(2026)Po, Zhang, Hertz, Wetzstein, Wadhwa, and
  Ruiz]{po2026multigen}
Ryan Po, Kai Zhang, Amir Hertz, Gordon Wetzstein, Neal Wadhwa, and Nataniel
  Ruiz.
\newblock {MultiGen}: Level-design for editable multiplayer worlds in diffusion
  game engines.
\newblock \emph{arXiv preprint arXiv:2603.06679}, 2026.

\bibitem[Polyak et~al.(2024)]{polyak2024moviegen}
Adam Polyak et~al.
\newblock {Movie Gen}: A cast of media foundation models.
\newblock \emph{arXiv preprint arXiv:2410.13720}, 2024.

\bibitem[Pondaven et~al.(2026)Pondaven, Wu, Gilitschenski, Torr, Tulyakov,
  Pizzati, and Siarohin]{pondaven2026actionparty}
Alexander Pondaven, Haoyu Wu, Igor Gilitschenski, Philip Torr, Sergey Tulyakov,
  Fabio Pizzati, and Aliaksandr Siarohin.
\newblock {ActionParty}: Multi-subject action binding in generative video
  games.
\newblock \emph{arXiv preprint arXiv:2604.02330}, 2026.

\bibitem[Qiu et~al.(2025)]{qiu2025gated}
Zihan Qiu et~al.
\newblock Gated attention for large language models: Non-linearity, sparsity,
  and attention-sink-free.
\newblock In \emph{Advances in Neural Information Processing Systems
  (NeurIPS)}, 2025.

\bibitem[Quevedo et~al.(2025)Quevedo, Sharma, Sun, Suryavanshi, Liang, and
  Yang]{quevedo2025worldgymworldmodelenvironment}
Julian Quevedo, Ansh~Kumar Sharma, Yixiang Sun, Varad Suryavanshi, Percy Liang,
  and Sherry Yang.
\newblock Worldgym: World model as an environment for policy evaluation, 2025.

\bibitem[Ramanana~Rahary et~al.(2026)Ramanana~Rahary, Dufour, P{\'e}rez, and
  Picard]{rahary2026oneview}
Adrien Ramanana~Rahary, Nicolas Dufour, Patrick P{\'e}rez, and David Picard.
\newblock {One View Is Enough}! monocular training for in-the-wild novel view
  generation.
\newblock \emph{arXiv preprint arXiv:2603.23488}, 2026.

\bibitem[{RLGym contributors}(2021)]{rlgym}
{RLGym contributors}.
\newblock {RLGym}: A python api for reinforcement learning in rocket league.
\newblock \url{https://rlgym.org}, 2021.

\bibitem[Robine et~al.(2023)Robine, H{\"o}ftmann, Uelwer, and
  Harmeling]{robine2023transformer}
Jan Robine, Marc H{\"o}ftmann, Tobias Uelwer, and Stefan Harmeling.
\newblock Transformer-based world models are happy with 100k interactions.
\newblock In \emph{International Conference on Learning Representations
  (ICLR)}, 2023.

\bibitem[Rombach et~al.(2022)Rombach, Blattmann, Lorenz, Esser, and
  Ommer]{rombach2022high}
Robin Rombach, Andreas Blattmann, Dominik Lorenz, Patrick Esser, and Bj{\"o}rn
  Ommer.
\newblock High-resolution image synthesis with latent diffusion models.
\newblock In \emph{IEEE/CVF Conference on Computer Vision and Pattern
  Recognition (CVPR)}, 2022.

\bibitem[Ruhe et~al.(2024)Ruhe, Heek, Salimans, and Hoogeboom]{ruhe2024rolling}
David Ruhe, Jonathan Heek, Tim Salimans, and Emiel Hoogeboom.
\newblock Rolling diffusion models.
\newblock In \emph{International Conference on Machine Learning (ICML)}, 2024.

\bibitem[Russell et~al.(2025)Russell, Hu, Bertoni, Fedoseev, Shotton, Arani,
  and Corrado]{russell2025gaia2}
Lloyd Russell, Anthony Hu, Lorenzo Bertoni, George Fedoseev, Jamie Shotton,
  Elahe Arani, and Gianluca Corrado.
\newblock {GAIA-2}: A controllable multi-view generative world model for
  autonomous driving.
\newblock \emph{arXiv preprint arXiv:2503.20523}, 2025.

\bibitem[Salimans et~al.(2016)Salimans, Goodfellow, Zaremba, Cheung, Radford,
  and Chen]{salimans2016improved}
Tim Salimans, Ian Goodfellow, Wojciech Zaremba, Vicki Cheung, Alec Radford, and
  Xi~Chen.
\newblock Improved techniques for training {GANs}.
\newblock In \emph{Advances in Neural Information Processing Systems
  (NeurIPS)}, 2016.

\bibitem[Sargent et~al.(2024)Sargent, Li, Shah, Herrmann, Yu, Zhang, Chan,
  Lagun, Fei-Fei, Sun, and Wu]{sargent2024zeronvs}
Kyle Sargent, Zizhang Li, Tanmay Shah, Charles Herrmann, Hong-Xing Yu, Yunzhi
  Zhang, Eric~Ryan Chan, Dmitry Lagun, Li~Fei-Fei, Deqing Sun, and Jiajun Wu.
\newblock {ZeroNVS}: Zero-shot 360-degree view synthesis from a single image.
\newblock In \emph{IEEE/CVF Conference on Computer Vision and Pattern
  Recognition (CVPR)}, 2024.

\bibitem[Savva et~al.(2026)Savva, Michel, Xie, et~al.]{savva2026solaris}
Georgy Savva, Oscar Michel, Saining Xie, et~al.
\newblock {Solaris}: Building a multiplayer video world model in {Minecraft}.
\newblock \emph{arXiv preprint arXiv:2602.22208}, 2026.

\bibitem[Schrittwieser et~al.(2020)Schrittwieser, Antonoglou, Hubert, Simonyan,
  Sifre, Schmitt, Guez, Lockhart, Hassabis, Graepel, Lillicrap, and
  Silver]{schrittwieser2020mastering}
Julian Schrittwieser, Ioannis Antonoglou, Thomas Hubert, Karen Simonyan,
  Laurent Sifre, Simon Schmitt, Arthur Guez, Edward Lockhart, Demis Hassabis,
  Thore Graepel, Timothy Lillicrap, and David Silver.
\newblock Mastering {Atari}, {Go}, chess and shogi by planning with a learned
  model.
\newblock \emph{Nature}, 2020.

\bibitem[Shazeer(2019)]{shazeer2019fast}
Noam Shazeer.
\newblock Fast transformer decoding: One write-head is all you need.
\newblock \emph{arXiv preprint arXiv:1911.02150}, 2019.

\bibitem[Shazeer(2020)]{shazeer2020glu}
Noam Shazeer.
\newblock {GLU} variants improve transformer.
\newblock \emph{arXiv preprint arXiv:2002.05202}, 2020.

\bibitem[Silver and Sutton(2025)]{silver2025welcome}
David Silver and Richard~S Sutton.
\newblock Welcome to the era of experience.
\newblock \emph{Google AI}, 2025.

\bibitem[Sim{\'e}oni et~al.(2025)Sim{\'e}oni, Vo, Seitzer, Baldassarre, Oquab,
  Jose, Khalidov, Szafraniec, Yi, Ramamonjisoa, et~al.]{simeoni2025dinov3}
Oriane Sim{\'e}oni, Huy~V. Vo, Maximilian Seitzer, Federico Baldassarre, Maxime
  Oquab, Cijo Jose, Vasil Khalidov, Marc Szafraniec, Seungeun Yi, Micha{\"e}l
  Ramamonjisoa, et~al.
\newblock {DINOv3}.
\newblock \emph{arXiv preprint arXiv:2508.10104}, 2025.

\bibitem[Singer et~al.(2023)Singer, Polyak, Hayes, Yin, An, Zhang, Hu, Yang,
  Ashual, Gafni, Parikh, Gupta, and Taigman]{singer2023makeavideo}
Uriel Singer, Adam Polyak, Thomas Hayes, Xi~Yin, Jie An, Songyang Zhang, Qiyuan
  Hu, Harry Yang, Oron Ashual, Oran Gafni, Devi Parikh, Sonal Gupta, and Yaniv
  Taigman.
\newblock {Make-A-Video}: Text-to-video generation without text-video data.
\newblock In \emph{International Conference on Learning Representations
  (ICLR)}, 2023.

\bibitem[Singh et~al.(2026)Singh, Zheng, Wu, Zhang, Shechtman, and
  Xie]{singh2026raev2}
Jaskirat Singh, Boyang Zheng, Zongze Wu, Richard Zhang, Eli Shechtman, and
  Saining Xie.
\newblock Improved baselines with representation autoencoders.
\newblock \emph{arXiv preprint arXiv:2605.18324}, 2026.

\bibitem[Song et~al.(2025)Song, Chen, Simchowitz, Du, Tedrake, and
  Sitzmann]{song2025history}
Kiwhan Song, Boyuan Chen, Max Simchowitz, Yilun Du, Russ Tedrake, and Vincent
  Sitzmann.
\newblock History-guided video diffusion.
\newblock In \emph{International Conference on Machine Learning (ICML)}, 2025.

\bibitem[Song and Dhariwal(2024)]{song2024improved}
Yang Song and Prafulla Dhariwal.
\newblock Improved techniques for training consistency models.
\newblock In \emph{International Conference on Learning Representations
  (ICLR)}, 2024.

\bibitem[Song et~al.(2023)Song, Dhariwal, Chen, and
  Sutskever]{song2023consistency}
Yang Song, Prafulla Dhariwal, Mark Chen, and Ilya Sutskever.
\newblock Consistency models.
\newblock In \emph{International Conference on Machine Learning (ICML)}, 2023.

\bibitem[Su et~al.(2024)Su, Lu, Pan, Murtadha, Wen, and Liu]{su2024roformer}
Jianlin Su, Yu~Lu, Shengfeng Pan, Ahmed Murtadha, Bo~Wen, and Yunfeng Liu.
\newblock {RoFormer}: Enhanced transformer with rotary position embedding.
\newblock \emph{Neurocomputing}, 2024.

\bibitem[Tillet et~al.(2019)Tillet, Kung, and Cox]{10.1145/3315508.3329973}
Philippe Tillet, H.~T. Kung, and David Cox.
\newblock Triton: an intermediate language and compiler for tiled neural
  network computations.
\newblock In \emph{Proceedings of the 3rd ACM SIGPLAN International Workshop on
  Machine Learning and Programming Languages}, MAPL 2019, page 10–19, New
  York, NY, USA, 2019. Association for Computing Machinery.
\newblock ISBN 9781450367196.
\newblock \doi{10.1145/3315508.3329973}.
\newblock \url{https://doi.org/10.1145/3315508.3329973}.

\bibitem[Tong et~al.(2026)Tong, Zheng, Wang, Tang, Ma, Brown, Yang, Fergus,
  LeCun, and Xie]{tong2026scalerae}
Shengbang Tong, Boyang Zheng, Ziteng Wang, Bingda Tang, Nanye Ma, Ellis Brown,
  Jihan Yang, Rob Fergus, Yann LeCun, and Saining Xie.
\newblock Scaling text-to-image diffusion transformers with representation
  autoencoders.
\newblock \emph{arXiv preprint arXiv:2601.16208}, 2026.

\bibitem[Tschannen et~al.(2025)Tschannen, Gritsenko, Wang, Naeem,
  Alabdulmohsin, et~al.]{tschannen2025siglip2}
Michael Tschannen, Alexey Gritsenko, Xiao Wang, Muhammad~Ferjad Naeem, Ibrahim
  Alabdulmohsin, et~al.
\newblock {SigLIP 2}: Multilingual vision-language encoders with improved
  semantic understanding, localization, and dense features.
\newblock \emph{arXiv preprint arXiv:2502.14786}, 2025.

\bibitem[Unterthiner et~al.(2018)Unterthiner, van Steenkiste, Kurach, Marinier,
  Michalski, and Gelly]{unterthiner2018towards}
Thomas Unterthiner, Sjoerd van Steenkiste, Karol Kurach, Raphael Marinier,
  Marcin Michalski, and Sylvain Gelly.
\newblock Towards accurate generative models of video: A new metric and
  challenges.
\newblock \emph{arXiv preprint arXiv:1812.01717}, 2018.

\bibitem[Valevski et~al.(2025)Valevski, Leviathan, Arar, and
  Fruchter]{valevski2025diffusion}
Dani Valevski, Yaniv Leviathan, Moab Arar, and Shlomi Fruchter.
\newblock Diffusion models are real-time game engines.
\newblock In \emph{International Conference on Learning Representations
  (ICLR)}, 2025.

\bibitem[Wang et~al.(2026)Wang, Jung, Monnier, Sohn, Zou, Xiang, Yeh, Liu,
  Huang, Nguyen-Phuoc, et~al.]{wang2026worldgen}
Dilin Wang, Hyunyoung Jung, Tom Monnier, Kihyuk Sohn, Chuhang Zou, Xiaoyu
  Xiang, Yu-Ying Yeh, Di~Liu, Zixuan Huang, Thu Nguyen-Phuoc, et~al.
\newblock Worldgen: From text to traversable and interactive 3d worlds.
\newblock In \emph{IEEE/CVF Conference on Computer Vision and Pattern
  Recognition (CVPR)}, 2026.

\bibitem[Wang et~al.(2024)Wang, Zhu, Huang, Chen, Zhu, and
  Lu]{wang2024drivedreamer}
Xiaofeng Wang, Zheng Zhu, Guan Huang, Xinze Chen, Jiagang Zhu, and Jiwen Lu.
\newblock {DriveDreamer}: Towards real-world-driven world models for autonomous
  driving.
\newblock In \emph{European Conference on Computer Vision (ECCV)}, 2024.

\bibitem[Wang et~al.(2004)Wang, Bovik, Sheikh, and Simoncelli]{wang2004image}
Zhou Wang, Alan~C. Bovik, Hamid~R. Sheikh, and Eero~P. Simoncelli.
\newblock Image quality assessment: From error visibility to structural
  similarity.
\newblock \emph{IEEE Transactions on Image Processing}, 13\penalty0 (4), 2004.

\bibitem[{Wayve}(2025)]{wayve2025gaia3}
{Wayve}.
\newblock {GAIA-3}: Advancing world models from simulation to evaluation, 2025.
\newblock \url{https://wayve.ai/press/wayve-launches-gaia3/}.

\bibitem[{World Labs}(2024)]{worldlabs2024}
{World Labs}.
\newblock Generating worlds, 2024.
\newblock \url{https://www.worldlabs.ai/blog/generating-worlds}.

\bibitem[{World Labs}(2025)]{worldlabs2025rtfm}
{World Labs}.
\newblock {RTFM}: A real-time frame model, 2025.
\newblock \url{https://www.worldlabs.ai/blog/rtfm}.

\bibitem[Wu et~al.(2026)Wu, Yu, Zou, and Liu]{wu2026multiworld}
Haoyu Wu, Jiwen Yu, Yingtian Zou, and Xihui Liu.
\newblock {MultiWorld}: Scalable multi-agent multi-view video world models.
\newblock \emph{arXiv preprint arXiv:2604.18564}, 2026.

\bibitem[Xu et~al.(2026)]{xu2026ifid}
Tongda Xu et~al.
\newblock Making reconstruction {FID} predictive of diffusion generation {FID}.
\newblock \emph{arXiv preprint arXiv:2603.05630}, 2026.

\bibitem[Yang et~al.(2025{\natexlab{a}})Yang, Li, Fan, Tian, and
  Wang]{yang2025ldetok}
Jiawei Yang, Tianhong Li, Lijie Fan, Yonglong Tian, and Yue Wang.
\newblock Latent denoising makes good tokenizers.
\newblock \emph{arXiv preprint arXiv:2507.15856}, 2025{\natexlab{a}}.

\bibitem[Yang et~al.(2024)Yang, Gao, Qiu, Chen, et~al.]{yang2024genad}
Jiazhi Yang, Shenyuan Gao, Yihang Qiu, Li~Chen, et~al.
\newblock Generalized predictive model for autonomous driving.
\newblock In \emph{IEEE/CVF Conference on Computer Vision and Pattern
  Recognition (CVPR)}, 2024.

\bibitem[Yang et~al.(2025{\natexlab{b}})Yang, Teng, Zheng, Ding, Huang, Xu,
  Yang, Hong, Zhang, Feng, et~al.]{yang2025cogvideox}
Zhuoyi Yang, Jiayan Teng, Wendi Zheng, Ming Ding, Shiyu Huang, Jiazheng Xu,
  Yuanming Yang, Wenyi Hong, Xiaohan Zhang, Guanyu Feng, et~al.
\newblock {CogVideoX}: Text-to-video diffusion models with an expert
  transformer.
\newblock In \emph{International Conference on Learning Representations
  (ICLR)}, 2025{\natexlab{b}}.

\bibitem[Yao et~al.(2025)Yao, Yang, and Wang]{yao2025reconstruction}
Jingfeng Yao, Bin Yang, and Xinggang Wang.
\newblock Reconstruction vs. generation: Taming optimization dilemma in latent
  diffusion models.
\newblock In \emph{IEEE/CVF Conference on Computer Vision and Pattern
  Recognition (CVPR)}, 2025.

\bibitem[Yin et~al.(2025)Yin, Zhang, Zhang, Freeman, Durand, Shechtman, and
  Huang]{yin2025causvid}
Tianwei Yin, Qiang Zhang, Richard Zhang, William~T. Freeman, Fr{\'e}do Durand,
  Eli Shechtman, and Xun Huang.
\newblock From slow bidirectional to fast autoregressive video diffusion
  models.
\newblock In \emph{IEEE/CVF Conference on Computer Vision and Pattern
  Recognition (CVPR)}, 2025.

\bibitem[Yu et~al.(2025)Yu, Qin, Liu, Yu, Zhang, Wang, et~al.]{yu2025position}
Jiwen Yu, Yiran Qin, Haoxuan Liu, Xiao Yu, Bowen Zhang, Xihui Wang, et~al.
\newblock Position: Interactive generative video as next-generation game
  engine.
\newblock \emph{arXiv preprint arXiv:2503.17359}, 2025.

\bibitem[Zhai et~al.(2023)Zhai, Mustafa, Kolesnikov, and
  Beyer]{zhai2023sigmoid}
Xiaohua Zhai, Basil Mustafa, Alexander Kolesnikov, and Lucas Beyer.
\newblock Sigmoid loss for language image pre-training.
\newblock In \emph{IEEE/CVF International Conference on Computer Vision
  (ICCV)}, 2023.

\bibitem[Zhang et~al.(2018)Zhang, Isola, Efros, Shechtman, and
  Wang]{zhang2018unreasonable}
Richard Zhang, Phillip Isola, Alexei~A. Efros, Eli Shechtman, and Oliver Wang.
\newblock The unreasonable effectiveness of deep features as a perceptual
  metric.
\newblock In \emph{IEEE/CVF Conference on Computer Vision and Pattern
  Recognition (CVPR)}, 2018.

\bibitem[Zhang et~al.(2023)Zhang, Wang, Sun, Yuan, and Huang]{zhang2023storm}
Weipu Zhang, Gang Wang, Jian Sun, Yetian Yuan, and Gao Huang.
\newblock {STORM}: Efficient stochastic transformer based world models for
  reinforcement learning.
\newblock In \emph{Advances in Neural Information Processing Systems
  (NeurIPS)}, 2023.

\bibitem[Zhang et~al.(2024)Zhang, Zhang, Li, Zhou, and
  Qiu]{zhang2023speechtokenizer}
Xin Zhang, Dong Zhang, Shimin Li, Yaqian Zhou, and Xipeng Qiu.
\newblock {SpeechTokenizer}: Unified speech tokenizer for speech language
  models.
\newblock In \emph{International Conference on Learning Representations
  (ICLR)}, 2024.

\bibitem[Zhang et~al.(2025)Zhang, Wei, Wu, He, Xu, Lyu, Zhang, Liu, Chen,
  et~al.]{zhang2025matrixgame}
Yifan Zhang, Chunli Wei, Dongdong Wu, Xianglong He, Yiqun Xu, Xinjie Lyu,
  Yongchao Zhang, Hao Liu, Yang Chen, et~al.
\newblock {Matrix-Game}: Interactive world foundation model.
\newblock \emph{arXiv preprint arXiv:2506.18701}, 2025.

\bibitem[Zheng et~al.(2025)Zheng, Ma, Tong, and Xie]{zheng2025rae}
Boyang Zheng, Nanye Ma, Shengbang Tong, and Saining Xie.
\newblock Diffusion transformers with representation autoencoders.
\newblock \emph{arXiv preprint arXiv:2510.11690}, 2025.

\bibitem[Zhou et~al.(2025)Zhou, Pan, LeCun, and Pinto]{zhou2025dino}
Gaoyue Zhou, Hengkai Pan, Yann LeCun, and Lerrel Pinto.
\newblock {DINO-WM}: World models on pre-trained visual features enable
  zero-shot planning.
\newblock In \emph{International Conference on Learning Representations
  (ICLR)}, 2025.

\bibitem[Zhu et~al.(2026)Zhu, Peng, Feng, et~al.]{zhu2026incantation}
Shangwen Zhu, Yiran Peng, Ruili Feng, et~al.
\newblock {Incantation}: Natural language as the action interface for
  multi-entity video world models.
\newblock \emph{arXiv preprint arXiv:2605.18601}, 2026.

\end{thebibliography}

\clearpage
\appendix
% The \appendix command is issued in main.tex before this file is included.

\section{Implementation Details}
\label{app:impl}

\paragraph{Codec.} \Cref{tab:codec-hparams} gives the full training, architecture, and loss
configuration of the baseline representation codec (\cref{sec:latent}) used in the codec
ablations, the frozen temporally-downsampled DINOv3 RAE.

\begin{table}[htbp]
\centering\small
\setlength{\tabcolsep}{8pt}\renewcommand{\arraystretch}{1.18}
\caption{\textbf{Codec training configuration.} Optimization, architecture, and loss
hyperparameters for the frozen tdown RAE used as the baseline codec in the ablations
(\cref{sec:exp-latent}). The codec is trained to $250$k steps; the ablations use its
$125$k-step checkpoint.}
\label{tab:codec-hparams}
\begin{tabular}{l l}
\toprule
Hyperparameter & Value \\
\midrule
\multicolumn{2}{l}{\emph{Compute and schedule}}\\
GPUs (DDP) & $8$ H100 ($1$ node) \\
Global batch size & $32$ clips ($4$ per device) \\
Training steps & $250$k (ablations use the $125$k checkpoint) \\
\texttt{torch.compile} & enabled \\
\midrule
\multicolumn{2}{l}{\emph{Optimizer (AdamW)}}\\
Learning rate & $2\times10^{-4}$ \\
$(\beta_1,\beta_2)$ & $(0.9,\,0.95)$ \\
Weight decay & $0$ \\
Learning-rate schedule & $1$k-step warmup, then decay to $1\times10^{-6}$ over $249$k steps \\
Weight EMA & none \\
Latent-statistics EMA & $0.99$ \\
\midrule
\multicolumn{2}{l}{\emph{Encoder (feature extractor + bottleneck)}}\\
Feature extractor & DINOv3-L, frozen \\
Aggregated blocks & $\{11,13,15,17,19,21,23\}$ \\
Latent channels & $32$ \\
Spatial / temporal downsampling & $\times2$ / $\times2$ (net $/32$ spatial, $10$\,Hz latent) \\
Input & $288\times512$, $40$ frames, $20$\,fps \\
\midrule
\multicolumn{2}{l}{\emph{Decoder (space-time ViT)}}\\
Width / depth / heads & $1152$ / $28$ / $16$ \\
MLP multiplier & $4$ \\
QK normalization & LayerNorm \\
Patch size (space / time) & $16$ / $2$ \\
\midrule
\multicolumn{2}{l}{\emph{Loss (adaptive gradient-norm weighting)}}\\
$L_1$ reconstruction & $1.0$ \\
LPIPS perceptual & $1.0$ (on $25\%$ of frames) \\
DINO latent consistency & $1.0$ (on $25\%$ of frames) \\
Perceptual DINO backbone & DINOv3-B \\
Adversarial (GAN) loss & none \\
\midrule
\multicolumn{2}{l}{\emph{Data}}\\
Players & $4$ \\
\bottomrule
\end{tabular}
\end{table}

\paragraph{World model.} \Cref{tab:wm-hparams} gives the training and architecture configuration of
the baseline single-player and multiplayer world models (\cref{sec:exp-multiplayer}). Both share the
same latent world model, optimizer, and frozen codec (\cref{tab:codec-hparams}); they differ only in
the number of players and the multiplayer conditioning.

\begin{table}[htbp]
\centering\small
\setlength{\tabcolsep}{7pt}\renewcommand{\arraystretch}{1.18}
\caption{\textbf{World model training configuration.} Baseline single-player and multiplayer world
models. Values that are shared span both columns; rows that differ are split. Both decode through the
frozen codec of \cref{tab:codec-hparams}.}
\label{tab:wm-hparams}
\begin{tabular}{l cc}
\toprule
Hyperparameter & Single-player & Multiplayer \\
\midrule
\multicolumn{3}{l}{\emph{Compute and schedule}}\\
GPUs (DDP) & $16$ H100 ($8\times2$ nodes) & $32$ H100 ($8\times4$ nodes) \\
Global batch & $16$ views & $32$ tiled ($128$ player-views) \\
Training steps & \multicolumn{2}{c}{$150$k} \\
\texttt{torch.compile} & \multicolumn{2}{c}{enabled} \\
\midrule
\multicolumn{3}{l}{\emph{Optimizer (AdamW)}}\\
Learning rate & \multicolumn{2}{c}{$1\times10^{-4}$} \\
$(\beta_1,\beta_2)$ & \multicolumn{2}{c}{$(0.9,\,0.99)$} \\
Weight decay & \multicolumn{2}{c}{$0.1$} \\
Schedule & \multicolumn{2}{c}{$1$k-step warmup, then constant} \\
Weight EMA decay & \multicolumn{2}{c}{$0.9999$} \\
\midrule
\multicolumn{3}{l}{\emph{Architecture (latent world model)}}\\
Hidden dimension & \multicolumn{2}{c}{$2048$} \\
Layers & \multicolumn{2}{c}{$16$} \\
Attention heads (query / KV) & \multicolumn{2}{c}{$16$ / $4$ (GQA)} \\
Temporal attention & \multicolumn{2}{c}{every $4$ layers} \\
Positional encoding & \multicolumn{2}{c}{RoPE (space and time)} \\
Patch size & \multicolumn{2}{c}{$1$} \\
Context length (training) & \multicolumn{2}{c}{$78$ latent frames} \\
Attention gating / AdaLN & \multicolumn{2}{c}{yes / yes} \\
Objective & \multicolumn{2}{c}{flow matching, causal (diffusion forcing)} \\
Input & \multicolumn{2}{c}{$288\times512$, $80$ frames, $20$\,fps} \\
\midrule
\multicolumn{3}{l}{\emph{Multiplayer conditioning}}\\
Players & $1$ & $4$ \\
View tiling & --- & $4$ views tiled, joint spatial attention \\
Per-player action dropout & --- & yes ($0.5$ subset-drop) \\
Action dropout (per step) & \multicolumn{2}{c}{$0.1$} \\
\bottomrule
\end{tabular}
\end{table}

\subsection{Live-demo model configuration}
\label{app:demo-config}

\Cref{tab:app-demo-hparams} lists the full set of hyperparameters used to train the
$5$B-parameter multiplayer world model served in the live demo
(\url{https://mira-wm.com}). This is the largest configuration in the paper: a
$4$-player, action-conditioned flow-matching transformer
(\cref{sec:architecture}) trained with diffusion forcing (\cref{sec:objective}) in
the latent space of the RAE codec (\cref{sec:latent}), on the $\approx 10{,}000$
clean match-hours of the multiplayer corpus (\Cref{app:data}). The model runs in
real time, emitting $20$ frames per second on a single Nvidia B200 GPU
(\cref{sec:streaming_inference}). The demo checkpoint is warm-started from a
single-player model pretrained for $30{,}000$ steps and then trained for a further
$100{,}000$ steps on multiplayer data; this is well short of the $2$M-step schedule
ceiling in the configuration.

\begin{table}[htbp]
\centering\small
\setlength{\tabcolsep}{8pt}\renewcommand{\arraystretch}{1.2}
\caption{\textbf{Live-demo model hyperparameters.} Full training configuration of the
$5$B multiplayer world model served at \url{https://mira-wm.com}.}
\label{tab:app-demo-hparams}
\begin{tabular}{l l}
\toprule
Hyperparameter & Value \\
\midrule
\multicolumn{2}{l}{\emph{Transformer architecture}} \\
Parameters                & $\approx 5$\,B \\
Hidden dimension          & $4096$ \\
Layers                    & $16$ \\
Attention heads           & $32$ \\
Key/value heads (GQA)     & $8$ \\
Patch size                & $1$ (one token per latent vector) \\
Positional encoding       & factorized spatio-temporal RoPE \\
Attention                 & QK-norm, sigmoid output gate \\
Feed-forward              & SwiGLU \\
Register tokens           & spatial + temporal \\
Action conditioning       & AdaLN (flow time $+$ action embedding) \\
\midrule
\multicolumn{2}{l}{\emph{Multiplayer}} \\
Players                   & $4$ \\
Per-player embedding      & yes \\
Action dropout            & per-player, $p = 0.1$ \\
\midrule
\multicolumn{2}{l}{\emph{Inputs}} \\
Resolution                & $288 \times 512 \times 3$ \\
Frame rate                & $20$\,fps \\
Training clip length      & $80$ frames ($4$\,s) \\
Context window (inference) & $20$ latent frames \\
Codec                     & RAE (\cref{sec:latent}) \\
Latent mean / std         & $0.457$ / $10.688$ \\
Actions (kbm)             & \texttt{W A S D Q E SPACE LSHIFT LCTRL} \\
\midrule
\multicolumn{2}{l}{\emph{Optimization}} \\
Optimizer                 & AdamW \\
Learning rate             & $1\times 10^{-4}$ \\
Betas                     & $(0.9, 0.99)$ \\
Weight decay              & $0.1$ \\
Gradient clipping         & $1.0$ \\
LR schedule               & $1000$ warmup steps, then constant \\
Minimum LR                & $1\times 10^{-6}$ \\
EMA decay ($\gamma$)      & $0.999$ \\
\midrule
\multicolumn{2}{l}{\emph{Training}} \\
Single-player pretraining & $30\,000$ steps \\
Multiplayer training      & $100\,000$ steps \\
Batch size (per device)   & $1$ \\
Train-set proportion      & $0.995$ \\
Training data             & $10{,}000$ match-hours \\
\bottomrule
\end{tabular}
\end{table}

\paragraph{Live-demo codec.} The demo model decodes through a temporally-downsampled
DINOv3 RAE (\cref{sec:latent}) that shares the architecture and loss of the baseline
codec in \cref{tab:codec-hparams}: $288\times512$ input at $20$\,fps, temporal patch
size $2$ for a $10$\,Hz latent, and $32$ latent channels. Relative to the ablation
codec it is trained longer and on more compute: $400$k steps on $4$ nodes
($32$ B200, DDP) at a global batch of $32$ clips, with \texttt{torch.compile}
enabled and the same AdamW optimizer and warmup schedule as \cref{tab:codec-hparams}.

\section{Data Details}
\label{app:data}

This appendix gives the action representation used during collection
(\cref{sec:data-collection}), the cross-player synchronization signals, the
action-noise injection procedure and full noise-composition breakdown, the
processed dataset scale, and the collection-time quality control.

\paragraph{Keybindings.} The nine-key vocabulary maps to different in-game
actions depending on whether the car is grounded or airborne
(\cref{tab:app-keybinds}).

\begin{table}[htbp]
\centering\small
\setlength{\tabcolsep}{8pt}\renewcommand{\arraystretch}{1.2}
\caption{\textbf{Keybindings.} In-game meaning of each key on the ground and in
the air.}
\label{tab:app-keybinds}
\begin{tabular}{l l l}
\toprule
Key & In-game action (ground) & In-game action (air) \\
\midrule
\texttt{W}      & Throttle forward (accelerate) & Pitch down (nose down) \\
\texttt{S}      & Throttle back (reverse)       & Pitch up (nose up) \\
\texttt{A}      & Steer left                    & Yaw left \\
\texttt{D}      & Steer right                   & Yaw right \\
\texttt{SPACE}  & Jump                          & Jump / flip / double-jump \\
\texttt{LSHIFT} & Boost                         & Boost \\
\texttt{LCTRL}  & Handbrake (powerslide)        & --- \\
\texttt{Q}      & ---                           & Air-roll left \\
\texttt{E}      & ---                           & Air-roll right \\
\bottomrule
\end{tabular}
\end{table}

\paragraph{Policy-to-keyboard mapping.} Nexto's three-valued axis commands and
binary buttons are translated to keypresses by thresholding each axis at
$\pm 0.5$ and conditioning on whether the car is grounded or airborne
(\cref{tab:app-mapping}).

\begin{table}[htbp]
\centering\small
\setlength{\tabcolsep}{8pt}\renewcommand{\arraystretch}{1.2}
\caption{\textbf{Policy-to-keyboard mapping.} Translation from Nexto output to
the nine-key vocabulary. Axes are thresholded at $\pm 0.5$; buttons fire on any
non-zero value.}
\label{tab:app-mapping}
\begin{tabular}{l l l l}
\toprule
Nexto output & Condition & On ground & In air \\
\midrule
Throttle  & $\geq +0.5$ & \texttt{W}      & --- \\
          & $\leq -0.5$ & \texttt{S}      & --- \\
Steer     & $\leq -0.5$ & \texttt{A}      & --- \\
          & $\geq +0.5$ & \texttt{D}      & --- \\
Pitch     & $\leq -0.5$ & ---             & \texttt{W} \\
          & $\geq +0.5$ & ---             & \texttt{S} \\
Yaw       & $\leq -0.5$ & ---             & \texttt{A} \\
          & $\geq +0.5$ & ---             & \texttt{D} \\
Roll      & $\leq -0.5$ & ---             & \texttt{Q} \\
          & $\geq +0.5$ & ---             & \texttt{E} \\
Jump      & $\neq 0$    & \texttt{SPACE}  & \texttt{SPACE} \\
Boost     & $\neq 0$    & \texttt{LSHIFT} & \texttt{LSHIFT} \\
Handbrake & $\neq 0$    & \texttt{LCTRL}  & \texttt{LCTRL} \\
\bottomrule
\end{tabular}
\end{table}

\paragraph{Synchronization signals.} Each VM records on its own local clock, so
raw timestamps are not directly comparable across the four players. To make
cross-player alignment possible later, we record the raw signals it needs: a
shared match identifier, per-frame video timestamps on the VM's clock, and
networked game events (kickoff countdown start, kickoff countdown end, goal scored, goal replay end) that fire at
the same logical instant in every player's recording. The alignment procedure
itself is described in \Cref{sec:data-processing}.

\paragraph{Action-noise injection.} To widen the behavioral distribution beyond
what a single deterministic policy produces, we optionally inject noise into a
player's actions. Noise was off for the first $21$ days of collection and on for
the final $5$ days. When it is on, each of the four players independently has a
$50\%$ chance of being made noisy; the rest play the bot policy unchanged. A
noisy player alternates between policy-driven and noise spans within the match,
with each round starting policy-driven so the car re-positions before noise
resumes. The noise type is fixed per player per match and is either no-op (zero
input) or uniformly random ($\{-1,0,1\}$ for axes, $\{0,1\}$ for buttons). Across
the full corpus, $83.48\%$ of matches are fully clean (all four views play the
unaltered policy) and $16.52\%$ contain at least one noisy view. Every recording
is tagged with its noise condition (mode and per-view) so the dataset can be
filtered or reweighted. All models in this paper train on the clean,
unaltered-policy recordings only ($\approx 10{,}000$ match-hours); the full
collected corpus is larger ($\approx 12{,}500$ match-hours, \Cref{tab:data-scale})
and additionally includes the noisy views, which we release in full but do not
train on.

\paragraph{Noise-composition breakdown.} \Cref{tab:app-noise} gives the full
distribution of matches over per-view noise compositions. A match is described by
how many of its four views are no-op (zero input) and how many are uniformly
random; the remaining views play the unaltered policy.

\begin{table}[htbp]
\centering\small
\setlength{\tabcolsep}{8pt}\renewcommand{\arraystretch}{1.2}
\caption{\textbf{Noise composition.} Match counts by per-view noise composition,
out of $99{,}408$ matches. ``Perfect'' means all four views play the unaltered
policy. The $16{,}425$ noisy matches ($16.52\%$) split across the remaining rows.}
\label{tab:app-noise}
\begin{tabular}{l rr}
\toprule
Composition & Matches & \% \\
\midrule
Perfect            & $82{,}983$ & $83.48$ \\
$1$ random, $1$ noop & $3{,}309$ & $3.33$ \\
$1$ random          & $2{,}184$ & $2.20$ \\
$1$ noop            & $2{,}150$ & $2.16$ \\
$2$ random          & $1{,}672$ & $1.68$ \\
$1$ random, $2$ noop & $1{,}672$ & $1.68$ \\
$2$ noop            & $1{,}630$ & $1.64$ \\
$2$ random, $1$ noop & $1{,}583$ & $1.59$ \\
$3$ noop            & $583$      & $0.59$ \\
$3$ random          & $573$      & $0.58$ \\
$2$ random, $2$ noop & $402$     & $0.40$ \\
$3$ random, $1$ noop & $272$     & $0.27$ \\
$1$ random, $3$ noop & $256$     & $0.26$ \\
$4$ noop            & $73$       & $0.07$ \\
$4$ random          & $66$       & $0.07$ \\
\bottomrule
\end{tabular}
\end{table}

\paragraph{Scale.} The figures below describe the processed $720$p\,/\,$20$\,fps
dataset (\Cref{sec:data-processing}), not the larger raw recordings; counts are
taken after the collection-time quality gates. The corpus comprises $99{,}408$
matches, or $397{,}632$ per-view recordings ($4$ views per match), totalling
roughly $12{,}454$ match-hours of gameplay ($\approx 49{,}816$ view-hours), with
a median match length of about $7.5$\,minutes ($\approx 451$\,s). Of the
match-hours, $9{,}998$ are fully clean and $2{,}456$ are noisy. The three maps are
near-uniformly represented, at roughly $33{,}100$ matches and $4{,}150$\,hours
each (\Cref{tab:data-scale}). The data was collected over about $26$ days
($2026\text{-}05\text{-}08$ to $2026\text{-}06\text{-}02$) across $120$
concurrent VMs, with some downtime and restarts.

\begin{table}[htbp]
\centering\small
\setlength{\tabcolsep}{6pt}\renewcommand{\arraystretch}{1.2}
\caption{\textbf{Dataset scale, per map.} Matches and gameplay hours after
processing, split by noise condition. A match is \emph{clean} when all four views
play the unaltered policy and \emph{noisy} otherwise.}
\label{tab:data-scale}
\begin{tabular}{l rrr rrr}
\toprule
& \multicolumn{3}{c}{Matches} & \multicolumn{3}{c}{Hours} \\
\cmidrule(lr){2-4}\cmidrule(lr){5-7}
Map & Clean & Noisy & Total & Clean & Noisy & Total \\
\midrule
Champions Field   & $27{,}670$ & $5{,}448$ & $33{,}118$ & $3{,}314.7$ & $814.6$ & $4{,}129.2$ \\
Forbidden Temple  & $27{,}658$ & $5{,}481$ & $33{,}139$ & $3{,}364.5$ & $821.5$ & $4{,}186.0$ \\
Deadeye Canyon    & $27{,}655$ & $5{,}496$ & $33{,}151$ & $3{,}318.9$ & $820.3$ & $4{,}139.2$ \\
\midrule
Total             & $82{,}983$ & $16{,}425$ & $99{,}408$ & $9{,}998.1$ & $2{,}456.3$ & $12{,}454.5$ \\
\bottomrule
\end{tabular}
\end{table}

\paragraph{Quality control.} Capture stalls, mid-match player dropouts, and
network lag cause the entire match to be discarded rather than shipped as partial
data. Long-running game instances are periodically restarted to prevent a slow
drift in capture quality.
\label{sec:data-qc}

\subsection{Data processing}
\label{sec:data-processing}

Processing turns the raw per-player recordings into the frame-aligned chunks the
model consumes. It aligns the four views onto a shared timeline, selects the
usable gameplay window, tiles it into fixed-length chunks, and emits
frame-aligned video, action, and physics files for each.

\paragraph{Cross-player alignment.} Because each VM records on its own clock, raw
timestamps are not comparable across players. We align on the end of the first
kickoff countdown, a networked game event that fires at the same logical instant
in all four recordings, which places every view on a shared master timeline used
by all downstream steps. The anchor is tight in practice: all VMs run in the same
network region, and the game's client-server networking keeps networked-event
presentation synchronized across clients even though one VM acts as host.

\paragraph{Gameplay-window selection.} We decode the per-match event log for the
match anchors (kickoff start and end, goal scored, goal-replay end). We then drop
the initial recorder warm-up window (about the first $8$\,s of each recording),
which covers the asset-streaming artifacts at the start of a fresh game process
(the opening video is slightly degraded while game assets are still streaming in),
and keep the remainder end-to-end, including kickoff
countdowns and goal replays.

\paragraph{Chunking.} The kept gameplay interval is tiled into fixed-length
$4$-second chunks, identified by $(\texttt{match\_id}, \texttt{player\_id},
\texttt{chunk\_idx})$. The same \texttt{chunk\_idx} across the four players covers
the same interval on the master timeline, so the four views of a chunk are
frame-aligned.

\paragraph{Video transcoding.} Per-player video is re-encoded into standardized
$720$p\,/\,$20$\,fps chunks, downsampled from the $720$p\,/\,$30$\,fps source. We
encode with H.264 and disable B-frames to avoid any non-causal temporal leakage,
and we seek frame-accurately per chunk with a two-stage approach (coarse to the
nearest keyframe, then fine to the target frame). No audio is kept.

\paragraph{Per-frame action labels.} The keypress event stream defines a step
function over the nine-key vocabulary (\Cref{sec:data-collection}). At each
video-frame instant on the master timeline we emit the set of keys held at that
moment, reading the step function with a hold-last-seen rule and no interpolation
between events. This produces one action row per video frame, written as one
action file per chunk and aligned with the corresponding video chunk.

\paragraph{Output format.} Each chunk ships as a triple of frame-aligned files:
the video (\texttt{mp4}), the action labels (one row per video frame), and the
physics state (one row per video frame, where the $120$\,Hz source telemetry is
hold-last-seen sampled at each frame timestamp, the same step-function
reconstruction used for actions). The three files share a
$(\text{match}, \text{player}, \text{chunk\_idx})$ key, so every video frame has
exactly one action row and one physics row. Fixing \texttt{match\_id} and
\texttt{chunk\_idx} and varying \texttt{player\_id} recovers the four
frame-aligned views of the same interval. Each chunk also carries per-(match,
player) metadata: which chunks were kept, their timestamps on the master
timeline, the bot identifier, the decoded match anchors, and the noise condition
propagated from \Cref{sec:data-collection}.

\section{Physics Game State Schema}
\label{app:physics}

This appendix gives the full schema of the physics game state logged during
collection (\cref{sec:data-physics}): one record per physics tick ($120$\,Hz),
comprising match-level state (\cref{tab:app-phys-match}), ball state
(\cref{tab:app-phys-ball}), and one car state per car on the field
(\cref{tab:app-phys-car}; four cars in 2v2).

\paragraph{Conventions.} \texttt{Vec3} is $(x, y, z)$ floats and \texttt{Quat} is
$(x, y, z, w)$ floats, both in the game's world frame; Rocket League uses a
$Z$-up coordinate system. Positions are in Unreal units (uu, with $1\,\text{uu} =
1\,\text{cm}$), velocities in uu/s, angular velocities in rad/s, and times in
seconds. Boost is normalized to $[0, 1]$ in the game code (shown as $0$--$100$ to
players during gameplay). A ``\,--\,'' in the unit column denotes a dimensionless
field.

\begin{table}[htbp]
\centering\small
\setlength{\tabcolsep}{8pt}\renewcommand{\arraystretch}{1.2}
\caption{\textbf{Match-level game state.}}
\label{tab:app-phys-match}
\begin{tabular}{l l l l}
\toprule
Field & Type & Unit & Description \\
\midrule
\texttt{time\_remaining} & float & s & seconds left in the match \\
\texttt{score\_blue}     & int   & -- & blue team score \\
\texttt{score\_orange}   & int   & -- & orange team score \\
\texttt{is\_overtime}    & bool  & -- & overtime active \\
\bottomrule
\end{tabular}
\end{table}

\begin{table}[htbp]
\centering\small
\setlength{\tabcolsep}{8pt}\renewcommand{\arraystretch}{1.2}
\caption{\textbf{Ball state.}}
\label{tab:app-phys-ball}
\begin{tabular}{l l l l}
\toprule
Field & Type & Unit & Description \\
\midrule
\texttt{location}         & Vec3 & uu    & world position \\
\texttt{velocity}         & Vec3 & uu/s  & linear velocity \\
\texttt{rotation}         & Quat & --    & orientation (unit quaternion) \\
\texttt{angular\_velocity} & Vec3 & rad/s & spin \\
\bottomrule
\end{tabular}
\end{table}

\begin{table}[htbp]
\centering\small
\setlength{\tabcolsep}{6pt}\renewcommand{\arraystretch}{1.2}
\caption{\textbf{Car state} (one entry per car).}
\label{tab:app-phys-car}
\begin{tabular}{l l l l}
\toprule
Field & Type & Unit & Description \\
\midrule
\texttt{player\_id}          & int   & --    & player identifier \\
\texttt{team}                & ubyte & --    & $0$ = blue, $1$ = orange \\
\texttt{is\_local}           & bool  & --    & this is the recording player's car \\
\texttt{location}            & Vec3  & uu    & world position \\
\texttt{velocity}            & Vec3  & uu/s  & linear velocity \\
\texttt{rotation}            & Quat  & --    & orientation (unit quaternion) \\
\texttt{angular\_velocity}   & Vec3  & rad/s & rotational velocity \\
\texttt{boost\_amount}       & float & --    & current boost, normalized ($0$--$1$) \\
\texttt{is\_on\_ground}      & bool  & --    & car is on the ground \\
\texttt{is\_on\_wall}        & bool  & --    & car is driving on a wall \\
\texttt{is\_supersonic}      & bool  & --    & car is at supersonic speed \\
\texttt{is\_demolished}      & bool  & --    & always false (not populated; see below) \\
\texttt{attacker\_player\_id} & int  & --    & demolishing player while demoed, else $-1$ \\
\texttt{has\_jumped}         & bool  & --    & first jump performed \\
\texttt{has\_double\_jumped} & bool  & --    & second jump used \\
\texttt{has\_flip}           & bool  & --    & flip/dodge still available \\
\texttt{is\_dodging}         & bool  & --    & currently in a flip/dodge \\
\texttt{jump\_time}          & float & s     & time since first jump (up to $0.2$\,s while jump held) \\
\texttt{dodge\_time}         & float & s     & seconds into current dodge \\
\texttt{max\_dodge\_time}    & float & s     & flip-window duration (static $1.25$\,s) \\
\bottomrule
\end{tabular}
\end{table}

\paragraph{Demolitions.} The \texttt{is\_demolished} flag is never populated (it
is always false) due to a broken game hook during data collection. A demolition
is instead recoverable from \texttt{attacker\_player\_id}, which holds the
demolishing player's id while the car is demoed and reverts to $-1$ shortly
afterwards.

\paragraph{Physical value ranges.} \Cref{tab:app-phys-ranges} lists engine caps
and arena geometry for the main physical quantities, useful for normalization and
sanity-checking by downstream users. These are engine limits and arena bounds for
vanilla 2v2 with no mutators, not empirical per-field minima and maxima of our
recordings. Values are drawn from the RLBot community wiki (``Useful Game
Values'', \texttt{wiki.rlbot.org}).

\begin{table}[htbp]
\centering\small
\setlength{\tabcolsep}{6pt}\renewcommand{\arraystretch}{1.2}
\caption{\textbf{Physical value ranges} (vanilla 2v2, no mutators). Engine caps
and arena geometry, not empirical ranges of the recordings.}
\label{tab:app-phys-ranges}
\begin{tabular}{l l l l}
\toprule
Quantity & Range / cap & Unit & Notes \\
\midrule
position $x$      & $[-4096, +4096]$ & uu      & side walls \\
position $y$      & $[-5120, +5120]$ & uu      & back walls; goal mouth extends to ${\sim}\pm6000$ \\
position $z$      & $[0, 2044]$      & uu      & floor to ceiling \\
car speed         & $\leq 2300$      & uu/s    & supersonic $\geq 2200$; no-boost drive $\leq 1410$ \\
ball speed        & $\leq 6000$      & uu/s    & \\
car ang.\ vel.\   & $\leq 5.5$       & rad/s   & per-axis \\
ball ang.\ vel.\  & $\leq 6.0$       & rad/s   & \\
\texttt{boost\_amount} & $[0, 1]$    & --      & in-game $0$--$100$; drains $33.3$/s \\
rotation          & unit quaternion  & --      & $|q| = 1$ \\
\texttt{jump\_time} & $[0, 0.2]$     & s       & accrues while jump held \\
\texttt{max\_dodge\_time} & $1.25$   & s       & constant (flip window) \\
ball radius       & $91.25$          & uu      & rest center $z \approx 93.15$ \\
gravity           & $650$            & uu/s$^2$ & downward ($-z$) \\
\bottomrule
\end{tabular}
\end{table}

\paragraph{Authority and per-view differences.} Every player's recording stores
the entire game state from that client's point of view. Because clients apply
client-side prediction to hide network latency, predicted values can differ
slightly across the four views; the effect is minor as matches run on a LAN. The
host machine holds the authoritative state and can be identified as the player
with the lowest \texttt{player\_id}. During goal replays the physics values are
frozen at the last state before the replay began. When we downsample the
$120$\,Hz telemetry to frame timestamps during processing
(\Cref{sec:data-processing}), we use the same hold-last-seen rule applied to
actions.

\begin{figure}[htbp]
  \centering
  \includegraphics[width=\linewidth]{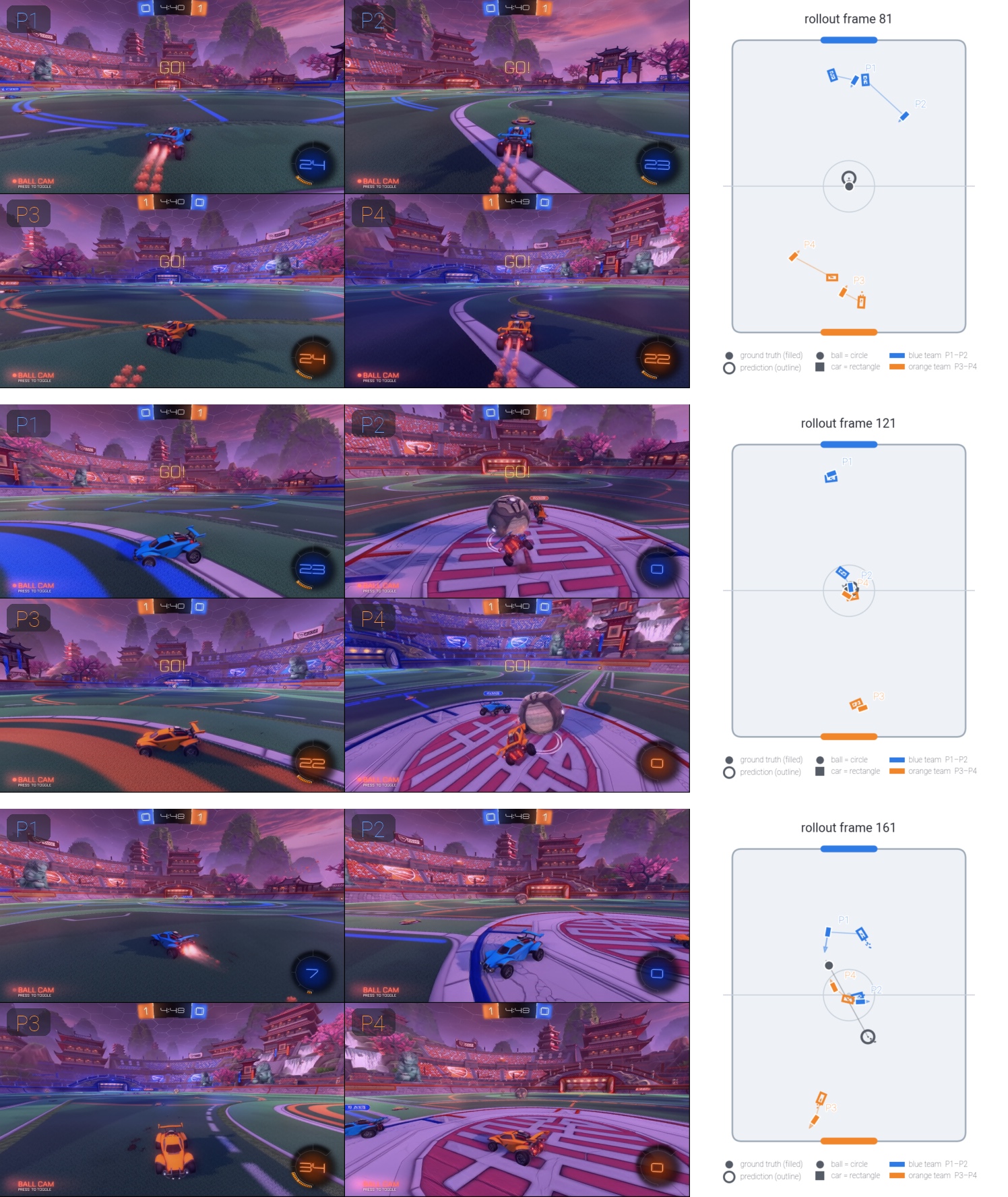}
  \caption{\textbf{Multiplayer rollout with game-state probe.} A rollout from the multiplayer world
  model; the game-state probe (\cref{sec:metrics}) reads ball and car positions and velocities from
  the model's pre-head activations.}
  \label{fig:rollout2}
\end{figure}

\section{Action Probe for Controllability}
\label{app:action-probe}

The action recoverability ratio (\cref{sec:metrics}) relies on a learned action probe. We report its
architecture and its accuracy on real held-out video so its calibration can be judged.

\paragraph{Architecture and training.} The probe is a frozen DINOv3-B (ViT-B/16) encoder with a small
(${\approx}4$M-parameter) attention-pooling head. Each $8$-frame ($0.4$\,s at $20$\,fps) chunk is
resized to $224$\,px and encoded frame by frame into a $14\times14$ grid of patch tokens; learnable
queries cross-attend over all tokens, with temporal and spatial positional embeddings, and an MLP maps
the pooled output to nine action logits. Training is multi-label: a control is positive if it is
pressed in any frame of the chunk, and the head is optimized with per-action binary cross-entropy on
the single-player training split. The DINOv3-B backbone stays frozen so the probe transfers across
real, reconstructed, and generated frames.

\paragraph{Accuracy on real data.} On $35{,}000$ held-out windows the probe reaches $0.838$ mAP and
$0.896$ mean AUROC (\cref{tab:action-probe}). Accuracy is high for common controls and lower for rare
ones such as reverse, mirroring the action-frequency trend that ARR reveals in the world model.

\begin{table}[htbp]
\centering\small
\caption{\textbf{Action-probe accuracy on real held-out video} ($35{,}000$ windows, nine controls).
Positive rate is the fraction of windows in which the control is pressed.}
\label{tab:action-probe}
\begin{tabular}{l c c c c}
\toprule
Action & Positive rate & F1 & AP & AUROC \\
\midrule
Forward     & $82.6\%$ & $0.903$ & $0.965$ & $0.850$ \\
Powerslide  & $47.2\%$ & $0.850$ & $0.941$ & $0.940$ \\
Right       & $46.9\%$ & $0.758$ & $0.856$ & $0.861$ \\
Left        & $44.7\%$ & $0.711$ & $0.829$ & $0.843$ \\
Boost       & $34.5\%$ & $0.718$ & $0.831$ & $0.875$ \\
Jump        & $22.2\%$ & $0.800$ & $0.894$ & $0.956$ \\
Air-roll Left  & $20.9\%$ & $0.738$ & $0.813$ & $0.946$ \\
Reverse     & $20.4\%$ & $0.542$ & $0.653$ & $0.858$ \\
Air-roll Right  & $18.7\%$ & $0.684$ & $0.756$ & $0.934$ \\
\midrule
Aggregate   & ---      & $0.745$ & $0.838$ & $0.896$ \\
\bottomrule
\end{tabular}
\end{table}

\section{Multiplayer Data Allocation at a Larger Budget}
\label{app:mp-larger}

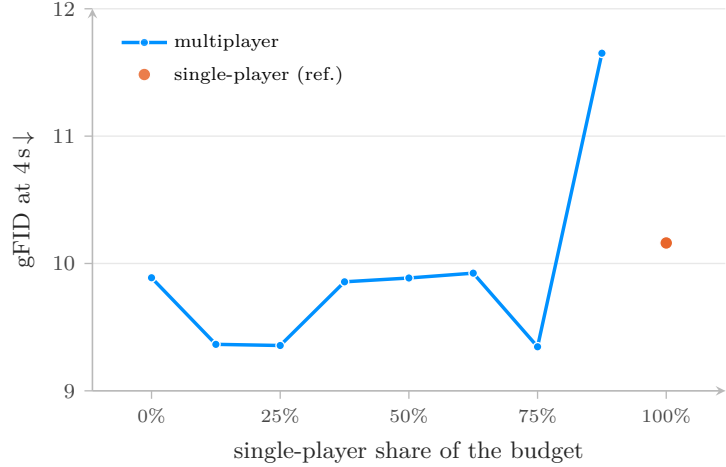
\begin{figure}[htbp]
  \centering
  \begin{tikzpicture}
  \begin{axis}[rsplot, width=0.6\linewidth, height=0.4\linewidth,
    xmin=-8, xmax=108, xtick={0,25,50,75,100},
    xticklabels={0\%,25\%,50\%,75\%,100\%}, x tick label style={font=\scriptsize},
    xlabel={single-player share of the budget},
    ymin=9, ymax=12, ytick={9,10,11,12}, ylabel={gFID at $4$\,s\,$\downarrow$},
    legend style={at={(0.03,0.97)}, anchor=north west, font=\scriptsize, draw=none,
                  fill=white, fill opacity=0.85, text opacity=1, cells={anchor=west}}]
    \addplot[plotblue, mark=*, mark size=1.6pt, line width=1.3pt]
      coordinates {(0,9.888)(12.5,9.365)(25,9.356)(37.5,9.856)(50,9.886)(62.5,9.924)(75,9.346)(87.5,11.651)};
    \addlegendentry{multiplayer}
    \addplot[plotorange, only marks, mark=*, mark size=2.4pt] coordinates {(100,10.161)};
    \addlegendentry{single-player (ref.)}
  \end{axis}
  \end{tikzpicture}
  \caption{\textbf{Multiplayer data allocation at the larger ($3.2$M-clip) budget.} gFID of one
  player's view at the $4$\,s horizon across the single-player share. With enough multiplayer steps,
  multiplayer trains adequately even from scratch ($0\%$), and multiplayer-heavy mixes reach the best
  gFID, beating both pure single-player ($100\%$) and pure multiplayer ($0\%$). Most mixes improve on
  pure single-player. Compare the single-player-budget sweep of \cref{fig:mp-data}.}
  \label{fig:mp-data-3p2m}
\end{figure}

The single-player--multiplayer allocation sweep of \cref{fig:mp-data} uses the single-player model's
own budget, where multiplayer training is starved of steps. At twice that budget ($3.2$M
player-views), multiplayer is no longer undertrained (\cref{fig:mp-data-3p2m}): it trains adequately
even from scratch, and multiplayer-heavy mixes fare best. A $75\%$-multiplayer blend ($25\%$
single-player) matches the lowest gFID and beats both pure multiplayer and pure single-player.

\section{Codec latent-space ablations}
\label{sec:codec-ablations}

These experiments detail the codec latent-space design choices deferred from \cref{sec:exp-latent}. The ablation protocol, metric suite, and baseline codec are all described there: each codec is judged by a $1$B-parameter world model trained on it, generation columns are reported at the $4$\,s horizon, and in every table the best value in each column is bold and the second best underlined.

\begin{takeaway}{Latent-space ablations, in brief.} 
\begin{itemize}[nosep, leftmargin=1.3em, topsep=3pt]
\item \textbf{Any reasonable pretrained extractor works.} Smaller DINOv3 variants and EUPE-B trail DINOv3-L only slightly and all resist drift (\cref{tab:exp-enc-family}).
\item \textbf{Read several feature-extractor layers, not just the last.} Averaging a spread of the feature extractor's intermediate blocks keeps spatial detail that the deepest block discards (\cref{tab:exp-enc-layers}).
\item \textbf{Compression can go in the codec or the world model.} The $2\times$ reduction works in either, so we do all of it in the codec (\cref{tab:exp-where}).
\item \textbf{Adaptive loss balancing helps.} Gradient-matched weights beat fixed weights on nearly all reconstruction and generation metrics (\cref{tab:exp-balance}).
\item \textbf{Upsample before decoding, not after.} Expanding the compact latent to the feature-extractor grid at the decoder's input, so the decoder runs on more tokens, far outperforms upsampling only at the output (\cref{tab:exp-upconv}).
\end{itemize}
\end{takeaway}

\paragraph{Which pretrained feature extractor?} We keep the feature extractor frozen and pretrained but swap DINOv3-L for the smaller DINOv3-B and DINOv3-S, and for EUPE-B, a compact feature extractor distilled from several vision foundation models into one \citep{eupe2026} and distinct from the DINOv3 family. We aggregate each feature extractor's intermediate blocks the same way (block indices rescaled to its depth), with the bottleneck's input width set to its feature dimension. Because P-DINO and the Fréchet DINO Distances are computed with a DINOv3-B backbone, we gray out those three metrics for the DINOv3-B feature extractor, which shares the metric's backbone and therefore has an unfair advantage (\cref{tab:exp-enc-family}).

\begin{table}[htbp]
\centering\scriptsize
\setlength{\tabcolsep}{3pt}\renewcommand{\arraystretch}{1.2}
\caption{\textbf{Pretrained feature extractor.} Smaller DINOv3 variants and a different pretrained family (EUPE-B). P-DINO and the FDD distances use a DINOv3-B backbone, so the DINOv3-B row is biased on those (grayed).}
\label{tab:exp-enc-family}
\begin{tabular}{l ccccccc | ccc}
\toprule
& \multicolumn{7}{c}{Codec (reconstruction)} & \multicolumn{3}{c}{World model (generation)} \\
\cmidrule(lr){2-8}\cmidrule(lr){9-11}
Feature extractor & PSNR\,$\uparrow$ & SSIM\,$\uparrow$ & LPIPS\,$\downarrow$ & P-DINO\,$\downarrow$ & rFID\,$\downarrow$ & rFVD\,$\downarrow$ & rFDD\,$\downarrow$ & gFID\,$\downarrow$ & gFVD\,$\downarrow$ & gFDD\,$\downarrow$ \\
\midrule
\textbf{DINOv3-L (ours)} & \textbf{29.7} & \textbf{0.891} & \textbf{0.051} & \textbf{0.021} & \textbf{5.4} & \textbf{38.6} & \textbf{0.17} & \textbf{10.7} & \textbf{163.1} & \textbf{0.55} \\
DINOv3-B & \textbf{29.7} & \textbf{0.891} & \textbf{0.051} & \textcolor{gray}{0.018} & \underline{5.8} & \underline{38.9} & \textcolor{gray}{0.08} & \underline{10.9} & \underline{178.8} & \textcolor{gray}{0.52} \\
DINOv3-S & \underline{29.1} & \underline{0.880} & 0.059 & 0.025 & 7.0 & 53.9 & 0.26 & 12.3 & 190.3 & 0.71 \\
EUPE-B & 29.0 & 0.876 & \underline{0.058} & \underline{0.023} & 6.2 & 52.2 & \underline{0.18} & 11.4 & 180.4 & \underline{0.56} \\
\bottomrule
\end{tabular}
\end{table}

\begin{takeaway}{Which pretrained feature extractor?} Any reasonable one. Smaller DINOv3 variants and EUPE-B trail DINOv3-L only slightly, which suggests richer representations may help downstream, and all of them stay equally low-drift over long rollouts (\cref{fig:drift}): resistance to drift is a property of pretraining, shared across feature extractors.\end{takeaway}

\paragraph{How many feature-extractor layers should the latent use?} The baseline averages features from seven DINOv3-L blocks ($\{11,13,15,17,19,21,23\}$) and adds the deepest as a residual, a choice adopted from RAEv2 \citep{singh2026raev2} (\cref{sec:latent}). We compare against reading only the final block (\cref{tab:exp-enc-layers}).

\begin{table}[htbp]
\centering\scriptsize
\setlength{\tabcolsep}{3pt}\renewcommand{\arraystretch}{1.2}
\caption{\textbf{Layer aggregation.} Aggregating seven DINOv3-L blocks (ours) against reading only the final block.}
\label{tab:exp-enc-layers}
\begin{tabular}{l ccccccc | ccc}
\toprule
& \multicolumn{7}{c}{Codec (reconstruction)} & \multicolumn{3}{c}{World model (generation)} \\
\cmidrule(lr){2-8}\cmidrule(lr){9-11}
Features read & PSNR\,$\uparrow$ & SSIM\,$\uparrow$ & LPIPS\,$\downarrow$ & P-DINO\,$\downarrow$ & rFID\,$\downarrow$ & rFVD\,$\downarrow$ & rFDD\,$\downarrow$ & gFID\,$\downarrow$ & gFVD\,$\downarrow$ & gFDD\,$\downarrow$ \\
\midrule
\textbf{Multi-layer mean (ours)} & \textbf{29.7} & \textbf{0.891} & \textbf{0.051} & \textbf{0.021} & \textbf{5.4} & \textbf{38.6} & \textbf{0.17} & \textbf{10.7} & \textbf{163.1} & \textbf{0.55} \\
Last block only & 29.4 & 0.887 & 0.062 & 0.024 & 7.8 & 55.0 & 0.25 & 12.2 & 172.4 & 0.62 \\
\bottomrule
\end{tabular}
\end{table}

\begin{takeaway}{How many feature-extractor layers should the latent use?} Several. Aggregating a spread of DINOv3 blocks beats the final block alone on every metric, in line with RAEv2 \citep{singh2026raev2}: earlier blocks keep spatial detail the deepest layer discards.\end{takeaway}

\paragraph{Where should the spatio-temporal compression happen?} Given that the latent is compressed, the $2\times$ spatial and $2\times$ temporal reduction can be done by the codec or by the world model, and we ask which is better. The baseline compresses in the codec, and the world model reads the latent with patch size $1$. We instead keep the codec at the higher resolution and let the world model's patch embedding do the $2\times$, in time or in space and time, compressing the latent as it reads it (\cref{tab:exp-where}).

\begin{table}[htbp]
\centering\scriptsize
\setlength{\tabcolsep}{3pt}\renewcommand{\arraystretch}{1.2}
\caption{\textbf{Where the compression happens.} Whether the $2\times$ spatial and $2\times$ temporal compression is done in the codec or in the world model's patch embedding.}
\label{tab:exp-where}
\begin{tabular}{ll ccccccc | ccc}
\toprule
\multicolumn{2}{c}{Compression done in} & \multicolumn{7}{c}{Codec (reconstruction)} & \multicolumn{3}{c}{World model (generation)} \\
\cmidrule(lr){1-2}\cmidrule(lr){3-9}\cmidrule(lr){10-12}
Spatial $2\times$ & Temporal $2\times$ & PSNR\,$\uparrow$ & SSIM\,$\uparrow$ & LPIPS\,$\downarrow$ & P-DINO\,$\downarrow$ & rFID\,$\downarrow$ & rFVD\,$\downarrow$ & rFDD\,$\downarrow$ & gFID\,$\downarrow$ & gFVD\,$\downarrow$ & gFDD\,$\downarrow$ \\
\midrule
\textbf{Codec} & \textbf{Codec} & 29.7 & 0.891 & 0.051 & 0.021 & 5.4 & 38.6 & 0.17 & \underline{10.7} & \underline{163.1} & \textbf{0.55} \\
Codec & World model & \underline{30.6} & \underline{0.909} & \underline{0.039} & \underline{0.016} & \underline{4.3} & \underline{23.0} & \underline{0.13} & \textbf{10.3} & \textbf{158.0} & \underline{0.56} \\
World model & World model & \textbf{31.9} & \textbf{0.926} & \textbf{0.029} & \textbf{0.013} & \textbf{3.8} & \textbf{13.4} & \textbf{0.12} & 11.8 & 214.3 & 0.63 \\
\bottomrule
\end{tabular}
\end{table}

\begin{takeaway}{Where should the spatio-temporal compression happen?} Either place works for the temporal reduction; the spatial reduction belongs in the codec. Moving the temporal $2\times$ into the world model's patch embedding matches doing it in the codec on generation, so that choice is free, whereas moving the spatial $2\times$ there too is worse on every generation metric. We therefore perform all of the compression in the codec.\end{takeaway}

\paragraph{Does adaptive loss balancing matter?} The baseline weights each perceptual term by the adaptive gradient-norm balancing of the codec loss (\cref{sec:latent}). We compare against fixing every weight to $1$, with no adaptive rescaling (\cref{tab:exp-balance}).

\begin{table}[htbp]
\centering\scriptsize
\setlength{\tabcolsep}{3pt}\renewcommand{\arraystretch}{1.2}
\caption{\textbf{Loss balancing.} Adaptive gradient-matched weights (ours) against fixed weights.}
\label{tab:exp-balance}
\begin{tabular}{l ccccccc | ccc}
\toprule
& \multicolumn{7}{c}{Codec (reconstruction)} & \multicolumn{3}{c}{World model (generation)} \\
\cmidrule(lr){2-8}\cmidrule(lr){9-11}
Balancing & PSNR\,$\uparrow$ & SSIM\,$\uparrow$ & LPIPS\,$\downarrow$ & P-DINO\,$\downarrow$ & rFID\,$\downarrow$ & rFVD\,$\downarrow$ & rFDD\,$\downarrow$ & gFID\,$\downarrow$ & gFVD\,$\downarrow$ & gFDD\,$\downarrow$ \\
\midrule
\textbf{Adaptive (ours)} & \textbf{29.7} & \textbf{0.891} & \textbf{0.051} & \textbf{0.021} & \textbf{5.4} & 38.6 & \textbf{0.17} & \textbf{10.7} & 163.1 & \textbf{0.55} \\
Fixed weights & 29.1 & 0.882 & 0.052 & 0.031 & 6.7 & \textbf{37.7} & 0.64 & 11.9 & \textbf{160.3} & 0.96 \\
\bottomrule
\end{tabular}
\end{table}

\begin{takeaway}{Does adaptive loss balancing matter?} Yes. Fixed weights worsen nearly all reconstruction and generation metrics, so rebalancing the perceptual terms against reconstruction during training is worth its small cost.\end{takeaway}

\paragraph{Where should the decoder upsample?} The decoder first upsamples the latent spatially, from the $/32$ latent grid back to the feature extractor's $/16$ patch grid, with a strided transposed convolution. We test placing this upconv as the decoder's first layer, before the ViT blocks (the baseline, so attention runs at $/16$), or after them (so attention runs at the coarser $/32$ latent and upsampling comes last) (\cref{tab:exp-upconv}).

\begin{table}[htbp]
\centering\scriptsize
\setlength{\tabcolsep}{3pt}\renewcommand{\arraystretch}{1.2}
\caption{\textbf{Upsampling placement.} The spatial upconv before the ViT blocks (ours) against after them.}
\label{tab:exp-upconv}
\begin{tabular}{l ccccccc | ccc}
\toprule
& \multicolumn{7}{c}{Codec (reconstruction)} & \multicolumn{3}{c}{World model (generation)} \\
\cmidrule(lr){2-8}\cmidrule(lr){9-11}
Upconv & PSNR\,$\uparrow$ & SSIM\,$\uparrow$ & LPIPS\,$\downarrow$ & P-DINO\,$\downarrow$ & rFID\,$\downarrow$ & rFVD\,$\downarrow$ & rFDD\,$\downarrow$ & gFID\,$\downarrow$ & gFVD\,$\downarrow$ & gFDD\,$\downarrow$ \\
\midrule
\textbf{Before ViT (ours)} & \textbf{29.7} & \textbf{0.891} & \textbf{0.051} & \textbf{0.021} & \textbf{5.4} & \textbf{38.6} & \textbf{0.17} & \textbf{10.7} & \textbf{163.1} & \textbf{0.55} \\
After ViT & 27.6 & 0.842 & 0.105 & 0.038 & 11.6 & 125.2 & 0.53 & 15.7 & 232.2 & 0.89 \\
\bottomrule
\end{tabular}
\end{table}

\begin{takeaway}{Where should the decoder upsample?} At its input, not its output. Expanding the latent to the feature-extractor grid at the decoder's input, so the decoder runs on more tokens at the finer grid, far outperforms upsampling last, the worst codec in the study.\end{takeaway}

\section{Pixel-space rollouts}
\label{sec:pixel-drift}

Without a codec, the pixel-space world model (\cref{tab:exp-pixels}) holds together for only a few
frames before its rollout drifts, a failure the latent-space models do not share.

\begin{figure}[htbp]
    \centering
    \includegraphics[width=0.166\linewidth]{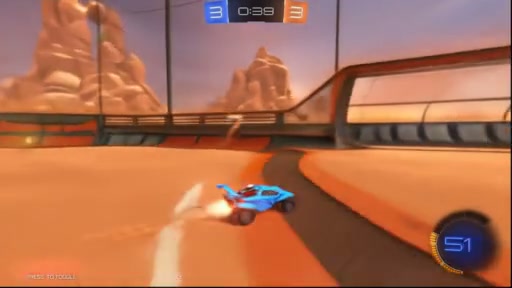}%
    \includegraphics[width=0.166\linewidth]{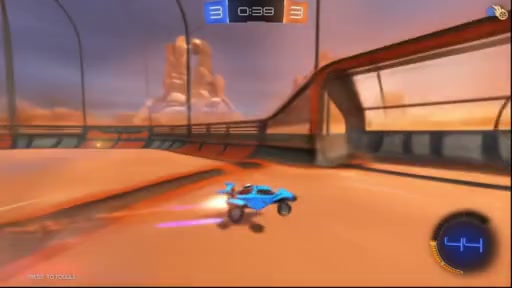}%
    \includegraphics[width=0.166\linewidth]{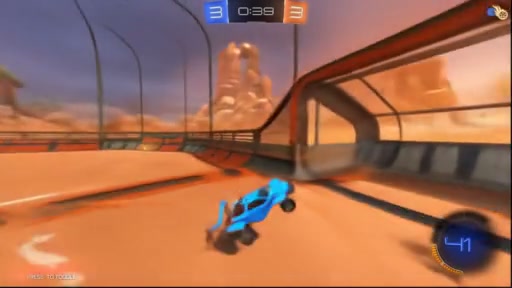}%
    \includegraphics[width=0.166\linewidth]{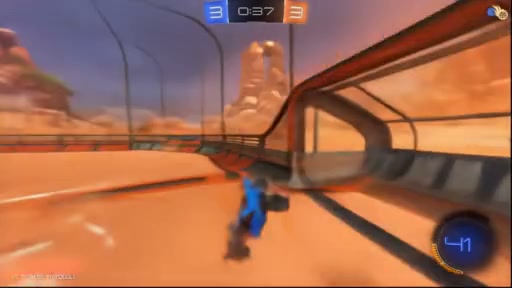}%
    \includegraphics[width=0.166\linewidth]{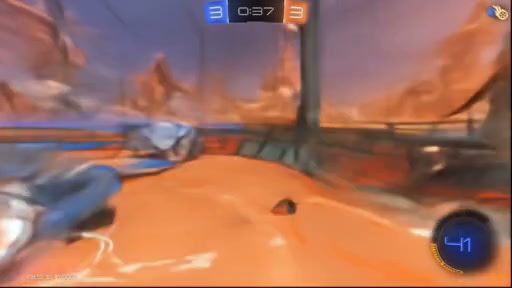}%
    \includegraphics[width=0.166\linewidth]{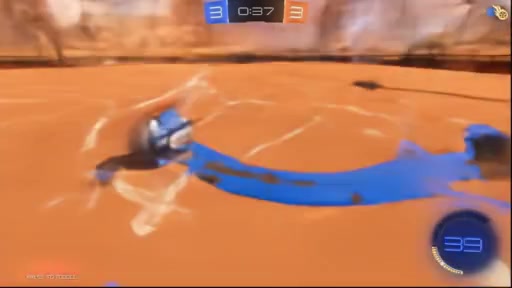}\\[3pt]
    \begin{tikzpicture}
      \draw[-{Stealth[length=2.4mm,width=2.4mm]}, line width=0.9pt, draw=black!65] (0,0) -- (\linewidth,0);
      \node[fill=white, inner xsep=4pt, font=\footnotesize\itshape, text=black!65] at (0.5\linewidth,0) {time};
    \end{tikzpicture}
    \caption{\textbf{Pixel-space rollouts drift.} A rollout from the plain pixel-space world
    model, shown left to right. The scene stays coherent for the first frames, then rapidly
    degrades into warped, unstructured texture, illustrating why pixel-space generation trails the
    latent baseline (\cref{tab:exp-pixels}).}
    \label{fig:pixeldrift}
\end{figure}

\section{Context Noise Across Metrics}
\label{app:noise-heatmaps}

\begin{figure}[htbp]
  \centering
  \resizebox{\linewidth}{!}{\input{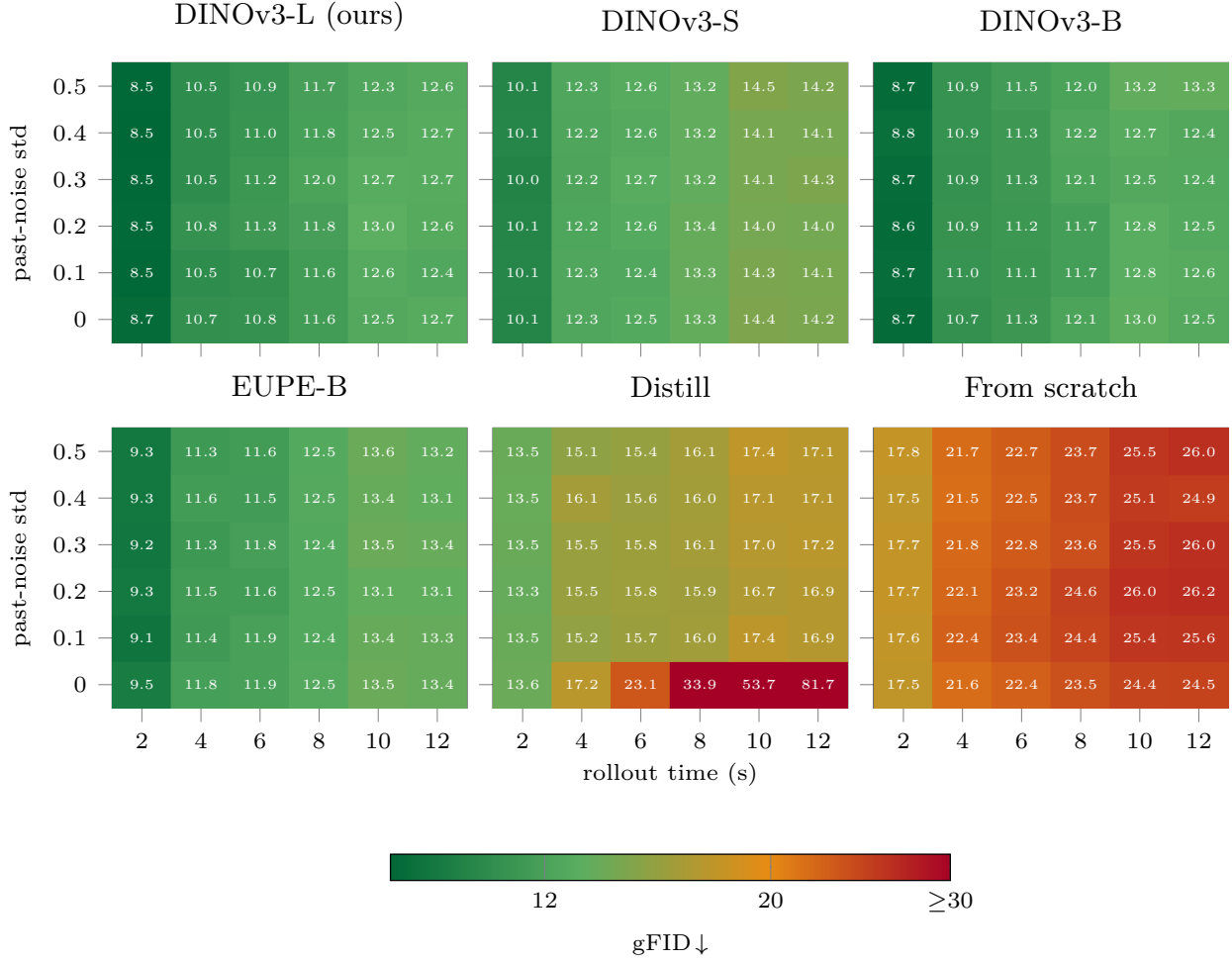}}
  \caption{\textbf{Context noise at inference, gFID (all codecs).} gFID against the past-noise
  standard deviation (vertical) and rollout time (horizontal), for all six codecs; color is
  clipped at $30$.}
  \label{fig:noise-gfid-full}
\end{figure}

\begin{figure}[htbp]
  \centering
  \resizebox{\linewidth}{!}{\input{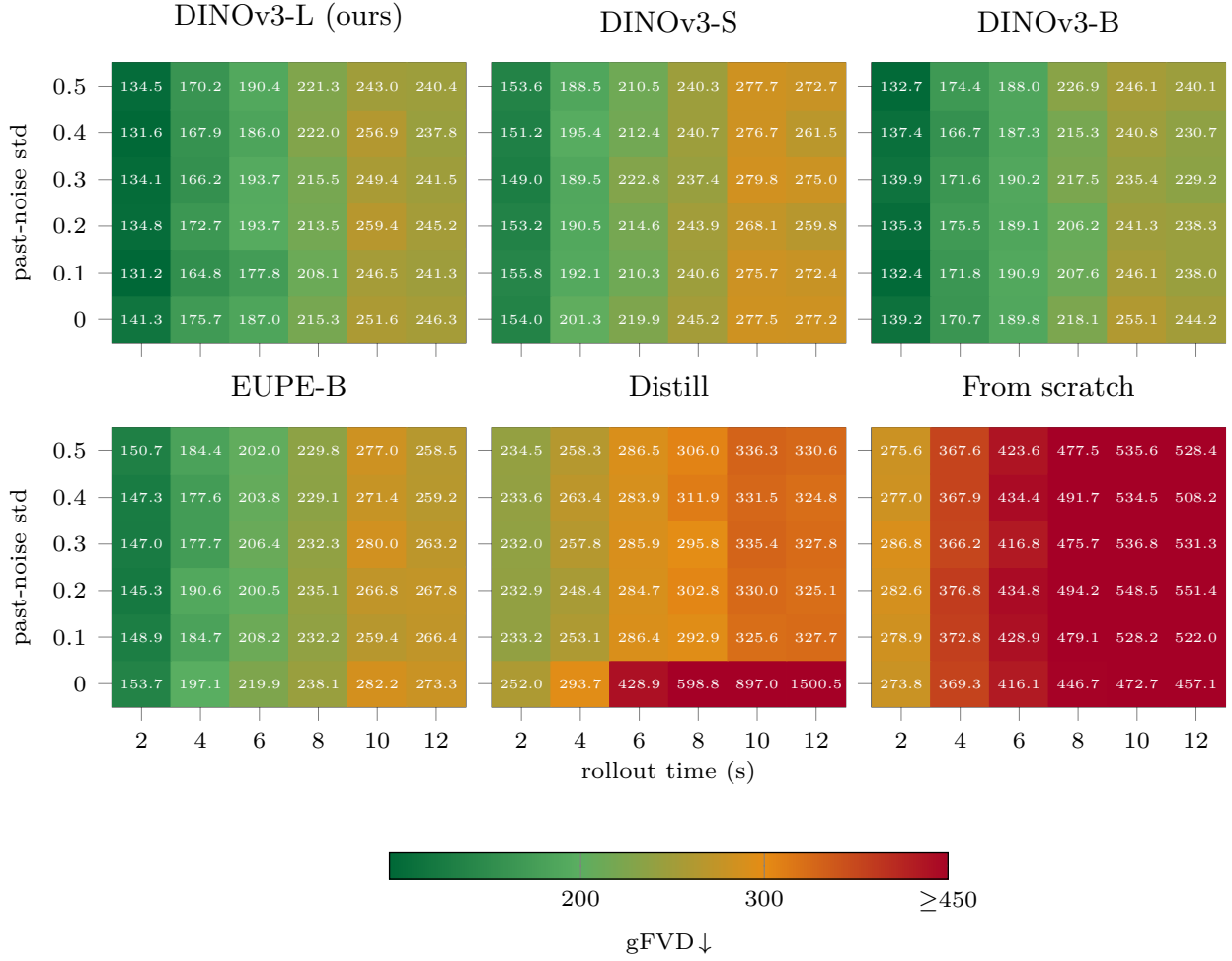}}
  \caption{\textbf{Context noise at inference, gFVD.} gFVD against the past-noise standard
  deviation (vertical) and rollout time (horizontal), for all six codecs; color is clipped at $450$.}
  \label{fig:noise-gfvd}
\end{figure}

\begin{figure}[htbp]
  \centering
  \resizebox{\linewidth}{!}{\input{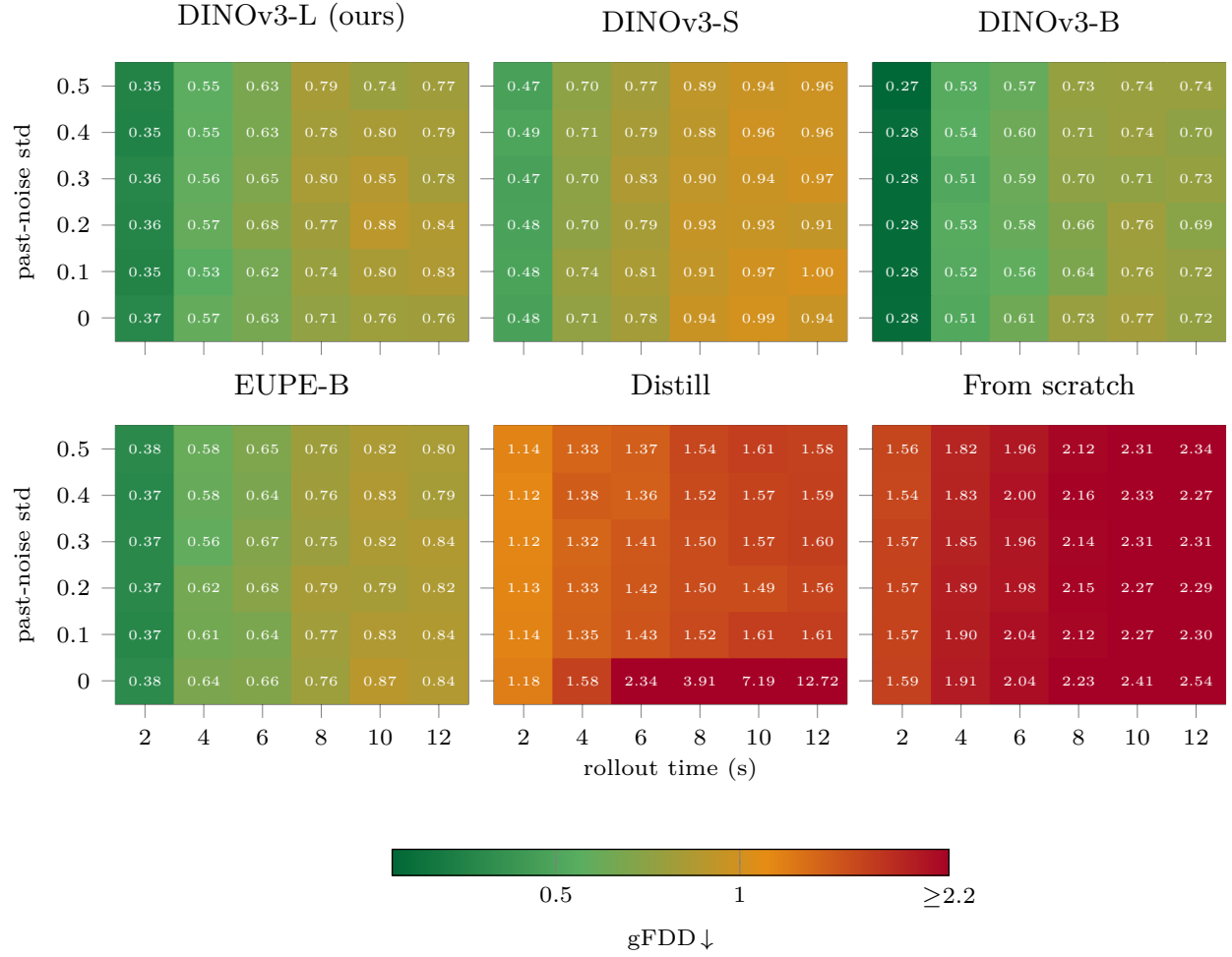}}
  \caption{\textbf{Context noise at inference, gFDD.} gFDD against the past-noise standard
  deviation (vertical) and rollout time (horizontal), for each codec; color is clipped at $2.2$.}
  \label{fig:noise-gfdd}
\end{figure}

\Cref{fig:exp-noise} reports gFID against the past-noise standard deviation and rollout time for
three representative codecs. Here we give the full set of six codecs (\cref{fig:noise-gfid-full})
and the other two generation metrics, gFVD (\cref{fig:noise-gfvd}) and gFDD
(\cref{fig:noise-gfdd}). All follow the same pattern: every codec is flat in noise except the
distilled one, whose quality collapses over the rollout when the past is fed clean and is
rescued by a small amount of noise.

\section{Human Evaluation}
\label{sec:human-eval}

\begin{figure}[htbp]
    \centering
    \includegraphics[width=0.95\linewidth]{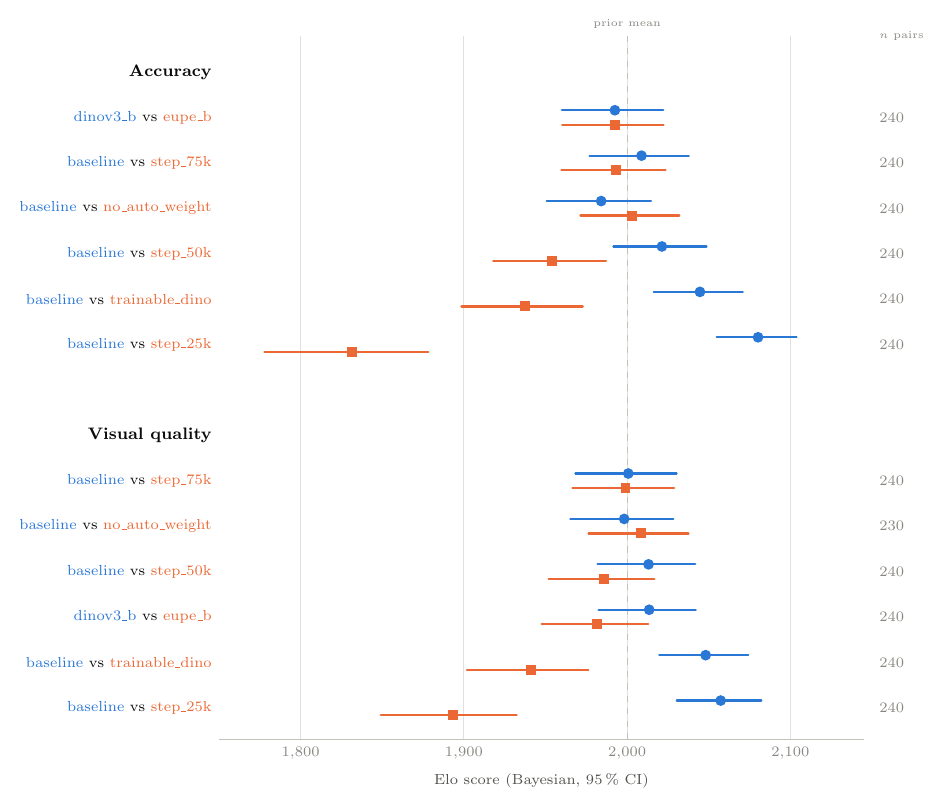}
    \caption{\textbf{Human evaluation: quality.} Pairwise Elo (Bayesian, $95\%$ CI) from the quality
    study of \cref{sec:metrics}, on accuracy and on visual quality, across a subset of comparisons.}
    \label{fig:he1}
\end{figure}

\begin{figure}[htbp]
    \centering
    \includegraphics[width=0.95\linewidth]{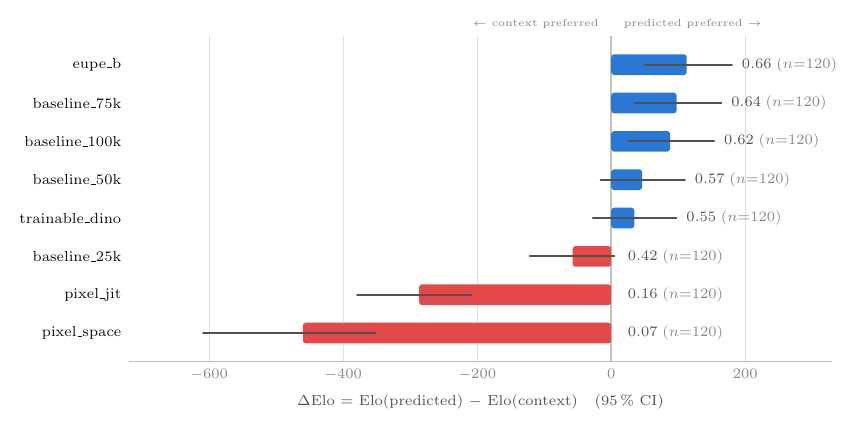}
    \caption{\textbf{Human evaluation: action adherence.} Elo delta between the action-conditioned
    prediction and the context from the adherence study of \cref{sec:metrics}, across a subset of
    comparisons; higher means raters more often prefer the predicted video. This ranking correlates
    strongly with ARR (\cref{sec:exp-controllability}, \cref{fig:arr-human}).}
    \label{fig:he2}
\end{figure}

We complement the automatic metrics with the two human preference studies described in
\cref{sec:metrics}, run on a subset of the comparisons. \cref{fig:he1} reports the pairwise quality
Elo and \cref{fig:he2} the action-adherence Elo delta. The action-adherence ranking correlates
strongly with ARR, validating it as a controllability proxy (\cref{sec:exp-controllability},
\cref{fig:arr-human}).

\section{Out-of-distribution recovery}
\label{app:recovery}

\begin{figure}[htbp]
    \centering
    \begin{minipage}[t]{0.48\linewidth}\centering
      \includegraphics[width=\linewidth]{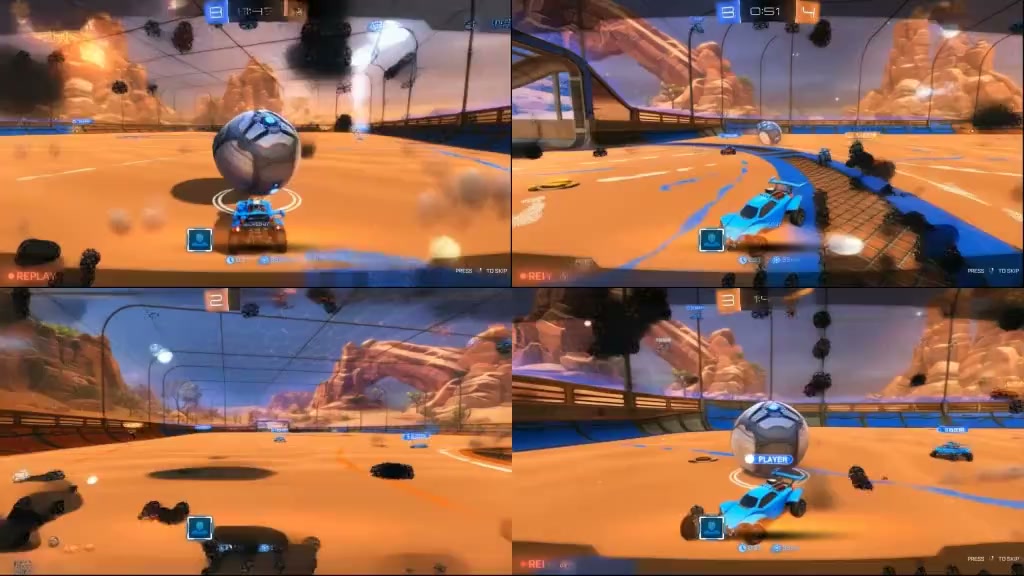}\\[2pt]{\scriptsize\textbf{1.}\ \itshape purposely driven out of distribution}
    \end{minipage}\hfill
    \begin{minipage}[t]{0.48\linewidth}\centering
      \includegraphics[width=\linewidth]{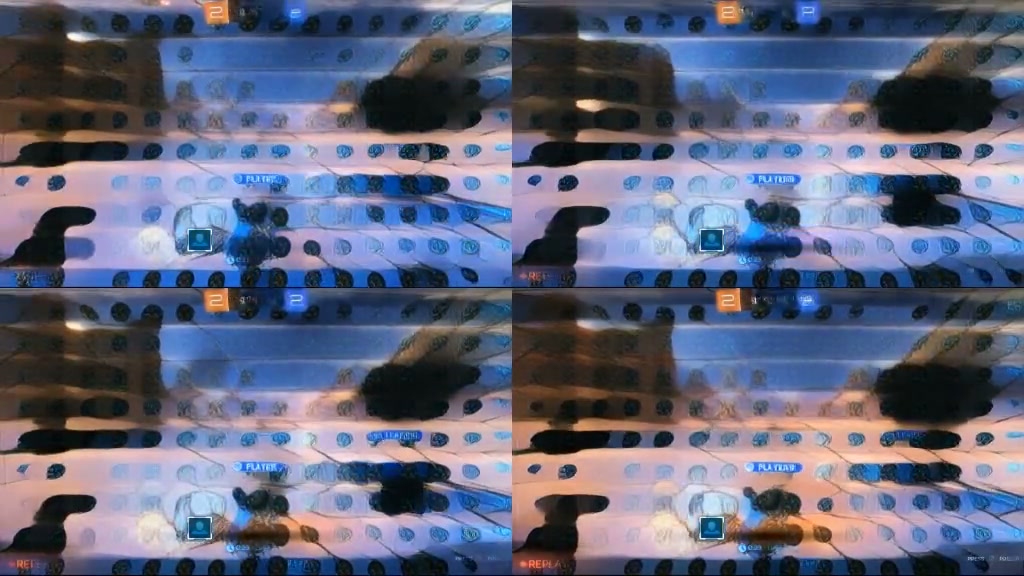}\\[2pt]{\scriptsize\textbf{2.}\ \itshape divergence grows}
    \end{minipage}\\[6pt]
    \begin{minipage}[t]{0.48\linewidth}\centering
      \includegraphics[width=\linewidth]{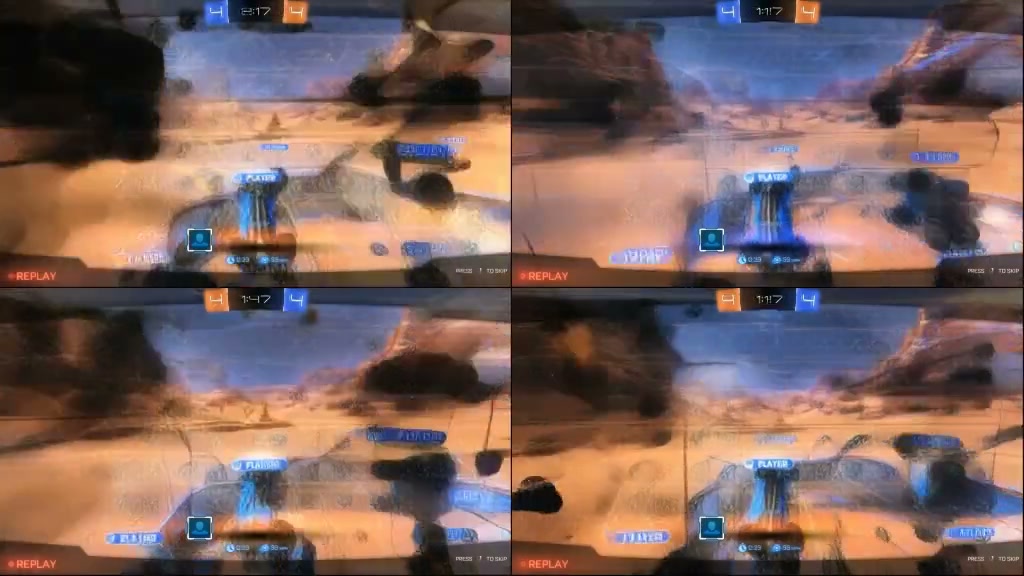}\\[2pt]{\scriptsize\textbf{3.}\ \itshape recovery begins}
    \end{minipage}\hfill
    \begin{minipage}[t]{0.48\linewidth}\centering
      \includegraphics[width=\linewidth]{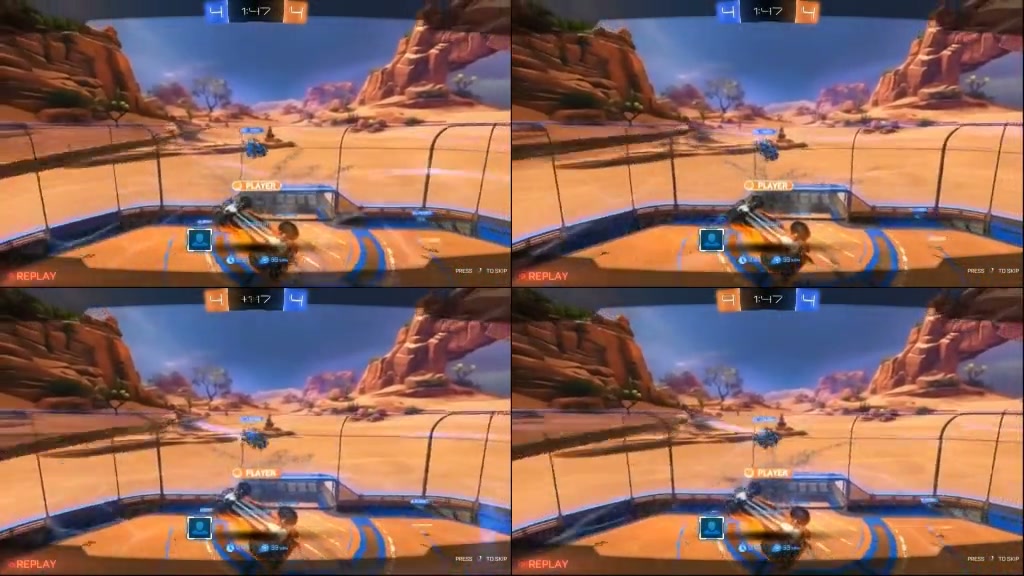}\\[2pt]{\scriptsize\textbf{4.}\ \itshape views resync on a goal replay}
    \end{minipage}\\[6pt]
    \begin{minipage}[t]{0.48\linewidth}\centering
      \includegraphics[width=\linewidth]{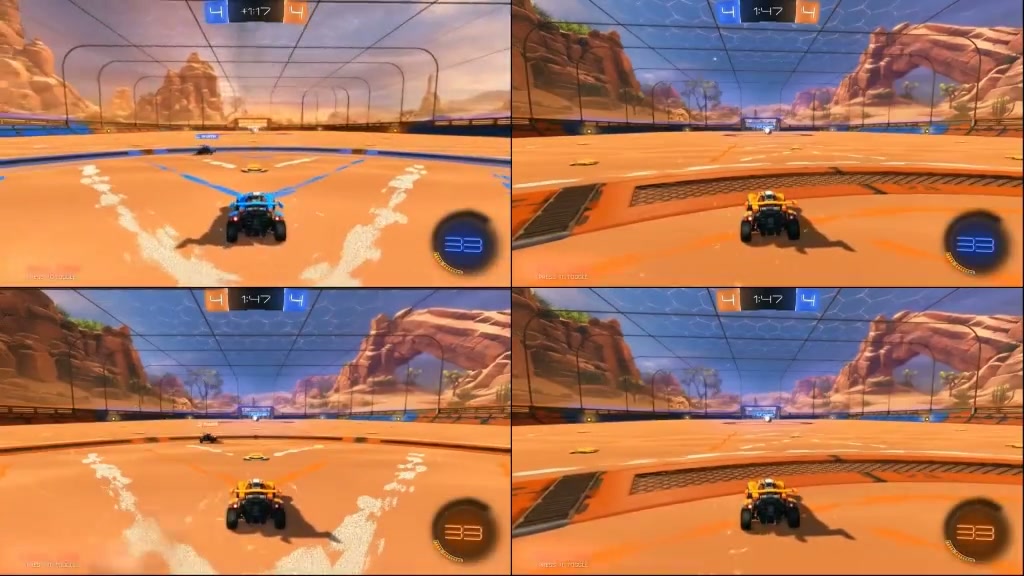}\\[2pt]{\scriptsize\textbf{5.}\ \itshape players respawn in each view}
    \end{minipage}\hfill
    \begin{minipage}[t]{0.48\linewidth}\centering
      \includegraphics[width=\linewidth]{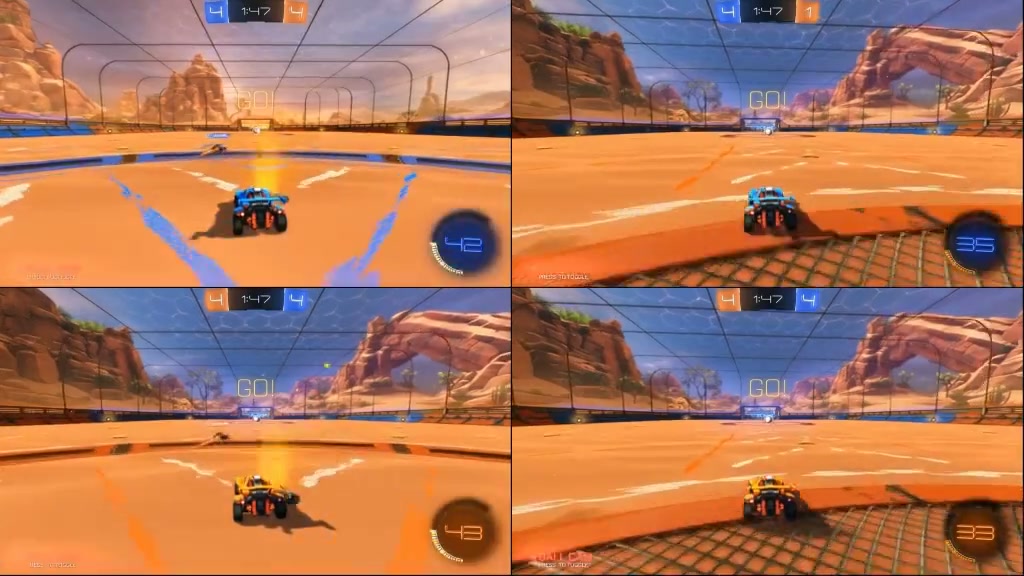}\\[2pt]{\scriptsize\textbf{6.}\ \itshape car color snaps back}
    \end{minipage}
    \caption{\textbf{Recovering from an out-of-distribution excursion.} A multiplayer rollout with
    the four tiled views, read left to right and top to bottom. We drive the model out of distribution
    on purpose~\textbf{(1)} and its four views diverge into noise~\textbf{(2)}; recovery then
    begins~\textbf{(3)} as the model resynchronizes the views on a scripted goal replay~\textbf{(4)},
    respawns the players in their own views at the post-goal kickoff~\textbf{(5)}, and snaps the last
    mis-colored car back to its correct color~\textbf{(6)}. We do not train for this recovery.}
    \label{fig:recovery}
\end{figure}

\Cref{fig:recovery} shows a multiplayer rollout recovering on its own after being driven far
out of distribution: the four views first diverge into noise, then resynchronize on a scripted
goal replay and resume coherent play. We do not train for this behavior.

\end{document}